\documentclass{article}
\usepackage{graphicx}
\usepackage{apalike}
\usepackage{setspace}
\usepackage{hyperref}
\usepackage{amsmath}
\usepackage{adjustbox}
\usepackage[left=1in,right=1in,top=1in,bottom=1in]{geometry}
\usepackage{subfig}

\title{Towards Autonomous Excavation Planning}
 
\author{
Lorenzo Terenzi, Marco Hutter \\
Robotics System Lab\\
ETH Zurich\\
8092 Zurich, Switzerland\\
}

\begin{document}

\maketitle

\begin{abstract}
  
Excavation plans are crucial in construction projects, dictating the dirt disposal strategy and excavation sequence based on the final geometry and machinery available. While most construction processes rely heavily on coarse sequence planning and local execution planning driven by human expertise and intuition, fully automated planning tools are notably absent from the industry. This paper introduces a fully autonomous excavation planning system. Initially, the site is mapped, followed by user selection of the desired excavation geometry. The system then invokes a global planner to determine the sequence of poses for the excavator, ensuring complete site coverage. For each pose, a local excavation planner decides how to move the soil around the machine, and a digging planner subsequently dictates the sequence of digging trajectories to complete a patch. We showcased our system by autonomously excavating the largest pit documented so far, achieving an average digging cycle time of roughly 30 seconds, comparable to the one of a human operator.

\end{abstract}

\section{Introduction}
The construction industry is essential for building infrastructure for core human needs like shelter and transport. Yet, it faces persistent issues: lagging productivity, rising labor costs \cite{BeatingLowproductivityTrap}, and exacerbated labor shortages \cite{AddressingUSConstruction}. Moreover, construction sites can be perilous, leading to high injury and fatality risks for workers \cite{LookWorkplaceDeaths}.

Advancements in automation, as seen in manufacturing, could be the answer. Especially, the automation of earthwork tasks using excavators offers promise due to their repetitive nature and the precision needed for modern computational landscape design \cite{hurkxkensRoboticLandscapesTopological2020}. While the complexity of these tasks has historically posed hurdles, recent research shows promising opportunities for creating fully autonomous excavation systems.

\begin{figure} [!hbt]
  \subfloat[]{
  \begin{minipage}{0.48\textwidth}
  \includegraphics[height=4.2cm,width=\linewidth]{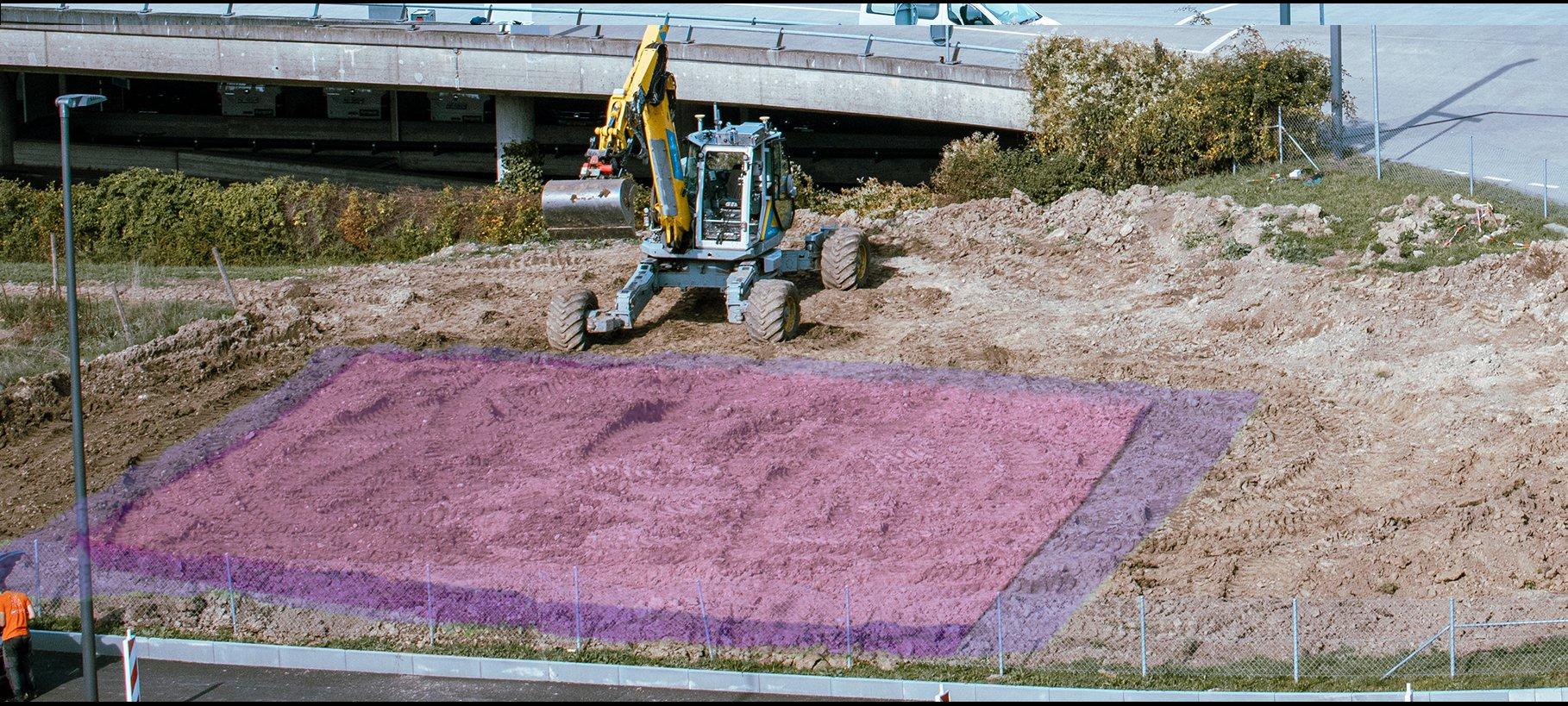}
  \end{minipage}
  }
  \subfloat[]{
  \begin{minipage}{0.48\textwidth}
  \includegraphics[height=4.2cm,width=\linewidth]{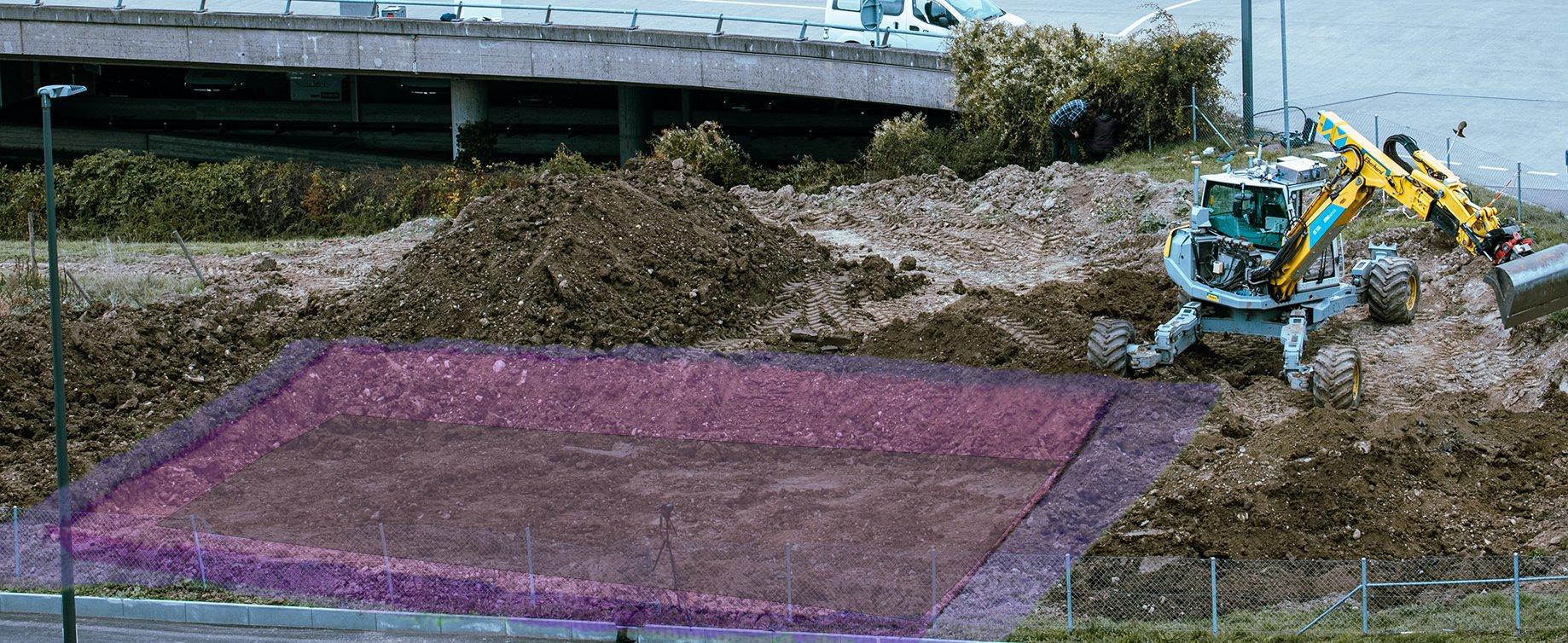}
  \end{minipage}
  }
  \caption{\textbf{(a)}: terrain before the excavation, the excavation area is indicated in violet. \textbf{(b)}: excavated pit, the sides and bottom are highlighted in violet and brown.}
  \label{fig:intro}
\end{figure}


A global plan outlines the sequence of excavator base poses to excavate the entire site while avoiding obstacles and preventing the excavator from becoming blocked or cornered by digging in the wrong sequence. The local-level planner then incorporates details on moving the earth around the excavator to follow and refine the global plan, including deciding where and how to dig to achieve the desired excavation geometry. The problem of global excavation planning is challenging as it involves both navigation and earthworks planning, which are closely interrelated. An incorrect digging sequence or improperly planned soil disposal can block the excavator and impede progress in other parts of the site.
Additionally, early decisions in the planning process can have a ripple effect on the entire plan, even hours later. The global planner must also consider obstacles for navigation and constraints on soil disposal, which the literature does not address adequately. Furthermore, in large sites with multiple digging areas, the problem of efficiently visiting all areas cannot be simplified by casting the problem into a graph traversal problem, such as the traveling salesman problem, as the connectivity of cells is dependent on the excavation subroutine used, cells may need to be traversed multiple times before being excavated, and there is no requirement to return to the starting point. Once a global plan is found, the local planner ensures successful execution by efficiently rearranging soil around the excavator's current position and transporting it to the designated disposal location without obstructing future navigation. Depending on the excavation geometry, this process may involve multiple rearrangements of excavated soil. Finally, the digging plan must consider the desired final excavation geometry and enable an excavator to execute it efficiently with a minimum number of scoops while also being able to re-plan in case of possible spillages or wall collapses. Additionally, it should avoid many small scoops to refine the geometry and achieve the desired geometry within the required tolerances in the least amount of time.
\\

A global plan outlines a sequence in which the excavator positions itself throughout the site, ensuring it avoids obstacles and doesn't get trapped or hindered by ill-advised excavation sequences. The local plan, on the other hand, delves into the specifics of the earth movement around the excavator, finetuning the overarching global strategy. It determines the precise excavation methods to achieve the desired shape.

The challenge in global excavation planning lies in its intertwined nature of navigation and earthworks planning. Mistakes in the excavation sequence or suboptimal soil disposal strategies can halt the excavator, impeding progress elsewhere on the site. Furthermore, initial planning decisions can profoundly impact the entire operation, with consequences manifesting even hours later. While the literature offers limited guidance, the global planner has to navigate obstacles and establish soil disposal constraints. Moreover, in expansive sites with numerous excavation zones, efficiently covering all areas is more complex than converting the task to a familiar graph traversal challenge like the traveling salesman problem. The interconnectedness of zones is influenced by the specific excavation methods employed. Zones might need multiple passes before complete excavation, and there's no mandate to return to the starting position.

Once the global strategy is in place, the local planner focuses on execution, managing the soil in the excavator's vicinity and ensuring transportation to the assigned disposal point without hampering subsequent navigation. Depending on the excavation design, this may necessitate several soil reconfigurations. The excavation strategy must also regard the intended final design, guiding the excavator to achieve this shape optimally using the fewest scoops. It must also be robust enough to adapt to unexpected issues like spillages or wall collapses and efficient enough to reach the target design swiftly, avoiding numerous minor adjustments.

In recent years, there have been advances in automating tasks like trenching \cite{cannonExtendedEarthmovingAutonomous1999,judRoboticEmbankment2021,kumparakBuiltRoboticsRaises2022a}, loading materials onto trucks \cite{stentzRoboticExcavatorAutonomous1998,zhangAutonomousExcavatorSystem2021a}, and embankments \cite{judRoboticEmbankment2021}. But these methods often focus on specific geometries or site sizes. While large-scale excavations have been attempted with coverage routines \cite{kimIntelligentNavigationStrategies2012,chosetCoverageKnownSpaces2000}, they seldom address local soil rearrangement and most rely heavily on heuristics \cite{judRoboticEmbankment2021,cannonExtendedEarthmovingAutonomous1999,judAutonomousFreeFormTrenching2019}.

The present work presents a holistic solution for autonomously excavating large structures, including rearranging soil and moving the machine. 
We treat the global excavation planning as a coverage problem, similar to\cite{kimIntelligentNavigationStrategies2012}. We develop a method to find the sequence of excavator base poses that allow the entire workspace to be excavated while satisfying constraints on soil disposal. The workspace is divided into cells, i.e., coverable areas. In coverage navigation, commonly employed in applications like floor cleaning robots, a 'coverable area' refers to a specific region within a larger workspace that can be navigated entirely, accessed, or serviced by a robot using its predefined navigation primitives or patterns. The order of cells is determined using graph navigation algorithms that preserve the connectivity of the non-excavated regions while minimizing path length. We then use dynamic programming to optimize the entry and exit points of coverable cells covered using simple zig-zag navigation primitives to simplify the dirt-handling problem as shown in \autoref{fig:paths}.
\newline
Earthmoving machine operators typically redistribute soil around the base of the excavator before moving to a new base pose. This method is more efficient regarding the volume of soil moved per hour instead of constantly alternating between digging and driving the machine. Thus, we define the problem of local earthwork planning as the task of proficiently redistributing soil around the base of the excavator to execute a global excavation plan. The area where this occurs, reachable by the excavator's arm from a fixed base pose, is termed the local workspace. The local planner selects the digging area based on the current excavated and desired geometry. The dump location is chosen primarily based on its proximity to the designated area for soil disposal. The digging planner uses Bayesian optimization to quickly determine the initial condition of a parametrized digging trajectory.
The navigation stack employs an RRT* planner and a pure pursuit controller to minimize travel distance and maximize clearance from obstacles and hazardous excavated areas. An overview of the system and its components is shown in \autoref{fig:system_architecture}.

\section{Related Work}
We build on prior work in research and field deployment of autonomous excavators, global earthworks planning, local earthworks planning, and trajectory planning for digging.

\paragraph*{Advancements in Excavator Automation}

There has been significant progress in the automation of excavators in recent years. Early efforts, such as LUCIE \cite{bradleyDevelopingRealtimeAutonomous1995}, focused on performing dig cycles without environmental perception. More recent work has incorporated range sensors \cite{cannonExtendedEarthmovingAutonomous1999,stentzRoboticExcavatorAutonomous1998} to enable autonomous digging in varied terrain profiles and the loading of soil onto trucks. The use of state machines and kinematic motion primitives has also been explored for autonomous trenching \cite{grollAutonomousTrenchingHierarchically2019}, with the addition of LIDAR-based elevation mapping and velocity and force control for the arm in \cite{judAutonomousFreeFormTrenching2019}. Built Robotics has recently developed an automation kit for trenching tasks \cite{kumparakBuiltRoboticsRaises2022a}.

Other recent research has focused on more complex excavation tasks, such as creating embankments \cite{judRoboticEmbankment2021}. Our work aims to extend this progress by addressing excavation tasks of different geometries and constraints on dirt disposal and navigation.

\paragraph*{Global Earthworks Planning}

Global earthworks planning problem involves devising a set of poses for the excavator during a task and determining a sequence of digging and dirt disposal areas. This problem has been addressed in a variety of ways in prior research.

In one approach, \cite{wooDeepReinforcementLearning2018a} employed a discretized workspace, using a reinforcement learning agent to generate a value function over the grid. The cell with the highest estimated value is selected for digging. However, this approach overlooks dirt handling and fails to ensure the reachability of the subsequent dig cell. On the other hand, \cite{judRoboticEmbankment2021,grollAutonomousTrenchingHierarchically2019} proposed task-specific navigation trajectories for excavating an embankment and a trench, respectively.

Global earthworks planning can also be viewed as a coverage problem, where the objective is to find a sequence of base poses that allow the arm to reach and excavate the entire area. Coverage path planning (CPP) algorithms have been widely used in diverse fields such as agriculture, house cleaning robots, underwater exploration, and mapping via unmanned aerial vehicles \cite{galceranSurveyCoveragePath2013}. Commonly, these algorithms presuppose the space to be covered is known beforehand.

Coverage algorithms such as Wavefront and Spanning Tree Coverage prove efficient when considering discretized space. These, however, can generate trajectories involving backtracking (impossible during digging) and possibly complex paths, thereby complicating dirt handling. If the space is continuous, the initial step of a CPP algorithm often involves dividing the space into coverable zones using simple motion subroutines like a standard zigzag trajectory. For this purpose, decomposition methods such as Morse \cite{acarMorseDecompositionsCoverage2002b} and Boustrophedon \cite{chosetCoverageKnownSpaces2000} can be employed.

\cite{kimIntelligentNavigationStrategies2012} proposed a coverage path planning algorithm for navigating multiple earthwork systems operating simultaneously in a known environment. This approach utilized a Morse function to partition the space and solved the Travelling Salesman Problem (TSP) to compute the sequence of cells to be processed. This method, however, has several limitations, including reliance on edge adjacency to establish cell connectivity, a large set of navigation patterns, and no consideration for dirt handling—additional research utilized dynamic programming techniques to determine both a global cell ordering and a viable path. Specifically, the work addressed the TSP-CPP problem for UAVs using dynamic programming and straightforward subroutines\cite{xieIntegratedTravelingSalesman2019}. Furthermore, a hierarchical TSP problem was investigated \cite{caoHierarchicalCoveragePath2020}.




 \paragraph*{Local Earthworks Planning}
The challenge of local earthworks planning has been largely overlooked in the literature. The solution is relatively straightforward for specific tasks like trenching, with the soil being dumped beside the trench \cite{grollAutonomousTrenchingHierarchically2019}. In other works, the problem is simplified by assuming that the dirt is immediately disposed of or dumped into a truck \cite{cannonExtendedEarthmovingAutonomous1999}.

Prior approaches to this problem have used a semi-circular digging workspace in front of the excavator. This concept has been reflected in various works, such as \cite{singhMultiresolutionPlanningEarthmoving1998a} and \cite{seoTaskPlannerDesign2011a}.

 \paragraph{Digging Planners and Controllers}
 A digging planner is responsible for planning the sequence of digging trajectories to complete a task within a given digging zone. The problem is typically divided into two subproblems: the design of a digging planner that selects the attack point (the point where the digging trajectory begins on the excavation surface) and the design of a 2D digging trajectory planner/controller that executes a digging trajectory given a predefined attack point and a digging plane.

\cite{singhMultiresolutionPlanningEarthmoving1998a} use expert operator heuristics to divide the digging zone into sections along the radial and tangential directions relative to the excavator base, which are dug sequentially until a certain precision is reached. However, these approaches can force the 2D digging planner to repeatedly dig low-volume scoops until the required precision in a subsection is reached. They may also make the plan brittle to wall collapses or soil spillage during the excavation of future adjacent sections or during dirt transport to the dump zone. In contrast, \cite{zhangAutonomousExcavatorSystem2021a} suggests using a data-driven approach to select the attack point of the digging trajectory by learning from human operator preferences.

\cite{sonExpertEmulatingExcavationTrajectory2020a} use dynamic motion primitives to efficiently learn from expert data, with a modulation module to adapt the trajectory to different soil types. \cite{leeRealTimeMotionPlanning2021b} use model predictive control (MPC) to plan digging trajectories in simulation. However, whether the system can bridge the sim-to-real gap without accurate soil modeling and a model of the machine's dynamics is uncertain. \cite{egliSoilAdaptiveExcavationUsing2022a} successfully demonstrate the deployment of a soil-adaptive digging controller trained with reinforcement learning (RL) in simulation only.

\subsection{Contributions}
This article presents a novel and comprehensive approach for autonomously excavating large sites using a single excavator. Demonstrated on a legged Menzi Muck M545 excavator, our framework offers a solution to specify the excavation area, constraints on navigation, and dirt disposal in georeferenced coordinates. It introduces a global workspace planner that calculates the required base poses, guaranteeing they are collision-free and achievable. In contrast to existing methods, we propose using cell corner adjacency, as these corners represent local subroutines' start or endpoints, to facilitate easier planning.

The local workspace planner in our approach is tasked with deciding soil redistribution around the excavator without base movement. Here, our system improves upon existing designs by creating five adaptable zones in the local workspace. These zones are dynamic in size, based on the required excavation geometry and serve as areas for digging and dumping soil.

The digging trajectory planner, another integral part of our system, determines the precise excavation point and trajectory. In a significant improvement over previous methods, we employ Bayesian optimization to identify the free parameters of a digging trajectory that greedily maximizes scooped soil volume.

Our system was put to the test by digging an approximately rectangular pit, measuring 11.6 x 15.6 x 1 m, which required planning intermediate dirt dumping sites. The successful completion of this complex task, involving hundreds of load cycles and moving about 300 tons of material within half a work day, demonstrates the system's capabilities and reliability in real-world excavation scenarios.

Notably, our system holds the distinction of being the first of its kind capable of autonomously performing diverse excavation tasks such as digging trenches, pits, and handling more complex projects with dirt disposal and navigation constraints. Its development marks a significant step forward in excavation automation, moving beyond the limitations of previous systems that were constrained to simple 2D tasks or required extensive human intervention.
It is also the first system to demonstrate reliable autonomous digging for hundreds of load cycles and the movement of approximately 300 tons of material in a half work day.

Highlighting our paper's key contributions:
\begin{itemize}
  \item A global workspace planner that decomposes the workspace using the Boustrophedon algorithm, finds the related quotient graph that encodes the connectivity of the workspace, computes the minimum branching tree of the graph, and finds the visiting sequence of the nodes with a postorder traversal to ensure the connectivity of yet-to-be-excavated areas. The planner also takes into account dumping constraints and the presence of obstacles when using dynamic programming to choose the start and end point of each cell.
  \item A local workspace planner that determines how to move dirt around the excavator without moving the base, considering user input and constraints on dirt disposal. The planner selects dig areas based on the discrepancy between the current soil geometry and the desired one and dump areas based on the distance to a user-designated dumping zone.
  \item A digging trajectory planner that aims at reaching the target geometry in a dig area by using Bayesian optimization to find the free parameters of a digging trajectory, which greedily maximizes scooped soil volume.
  \item A safe and robust navigation system based on a sampling planner and pure pursuit controller that allows the excavator to navigate to the next base pose while avoiding obstacles and maintaining a safe distance from the excavation site.
  \item An experimental validation and quantitative results that set a new benchmark for autonomous excavation performance
  \item The introduction of a novel dataset comprising authentic building silhouettes and urban crop layouts, the development of a software program for generating realistic excavation shapes through procedural methods and a benchmark to assess the effectiveness of the excavation planning system.  All associated code and resources are open-sourced at the digbench repository footnote{\url{git@github.com:leggedrobotics/digbench.git}}.
\end{itemize}

\section{Method}

\paragraph{HEAP}
HEAP is a M545 Menzi Muck 12-ton legged excavator that has been adapted for autonomous forestry operations \cite{jelavicRoboticPrecisionHarvesting2022}, rock wall construction \cite{johnsAutonomousDryStone2020a}, and digging tasks \cite{judRoboticEmbankment2021,egliSoilAdaptiveExcavationUsing2022a}. The chassis is equipped with servo valves and pressure sensors that enable the deployment of a force-based chassis balancing controller \cite{hutterForceControlActive2017}, which significantly aids navigation over challenging terrain. The system is equipped with an Ouster OS-0 128 LIDAR for mapping, localization, and online site mapping during operations. A GPS receiver with RTK correction and an IMU is mounted on the cabin. IMUs are also mounted on each arm link to estimate the kinematic position of the shovel. An encoder measures the cabin's orientation relative to the base. For more details on the hardware setup, see \cite{judHEAPAutonomousWalking2021}.

\begin{figure}[!hbt]
  \centering
  \rotatebox{-90}{\includegraphics[width=\linewidth]{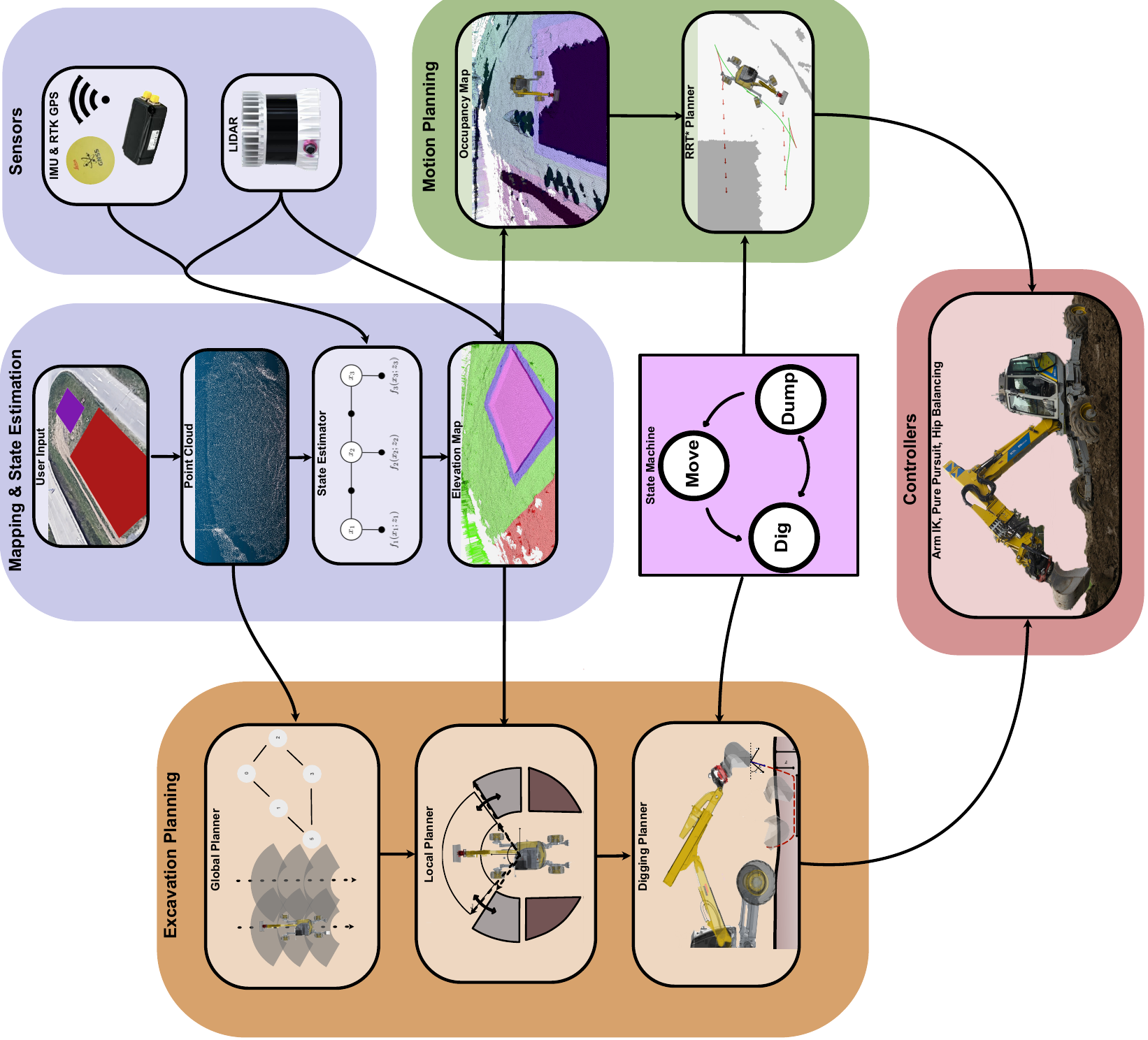}}
  \caption{System architecture diagram. Subsystems are grouped in modules indicated in different colors: sensors(blue), mapping and state estimation (violet), excavation planning (orange), motion planning (green), controllers (red). The state machine in the middle of the diagram triggers the execution of the different modules. The arrows represent the flow of information from one subsystem to another.}
  \label{fig:system_architecture}
\end{figure}

\paragraph{System Overview} The deployed system integrates mapping, localization, and various planning components to enable autonomous excavation. It incorporates a user interface for excavation planning, global earthwork planning, local excavation planning, a digging planner, and a navigation sampling-based planner, as depicted in \autoref{fig:system_architecture}. To assess the effectiveness of the excavation planning system. The planning modules can use different digging controllers, ranging from straightforward kinematic controllers to advanced reinforcement learning strategies.



\subsection{Mapping and State Estimation}

\paragraph{Mapping}
To plan excavation tasks effectively, the system requires a precise construction site map. This map, often represented as an elevation map, outlines the excavation area, target depth, dump zones, and any obstacles. Additionally, it can be used to differentiate between the original terrain and newly excavated soil.

To produce an initial map for global planning, we manually traversed the excavation site and collected point clouds. These were then registered in a georeferenced frame with a voxel size of 0.05 m using Open3D SLAM \cite{jelavicOpen3DSLAMPoint2022b}. This ICP-based SLAM system identifies loop closures through local map matching. We registered the map by integrating GPS readings with ICP registration. We used the Earth-centered, Earth-fixed (ECEF) coordinate system for mapping since excavation plans typically define dig points using ECEF or its derivatives. Alternatively, this map can also be acquired using traditional surveying methods \cite{LeicaGeosystems} or other robotic technologies like drones \cite{WingtraMappingDrone}.

\begin{figure}[!hbt]
\subfloat[Point cloud map]{
\begin{minipage}{0.48\textwidth}
\includegraphics[height=4cm, width=\linewidth]{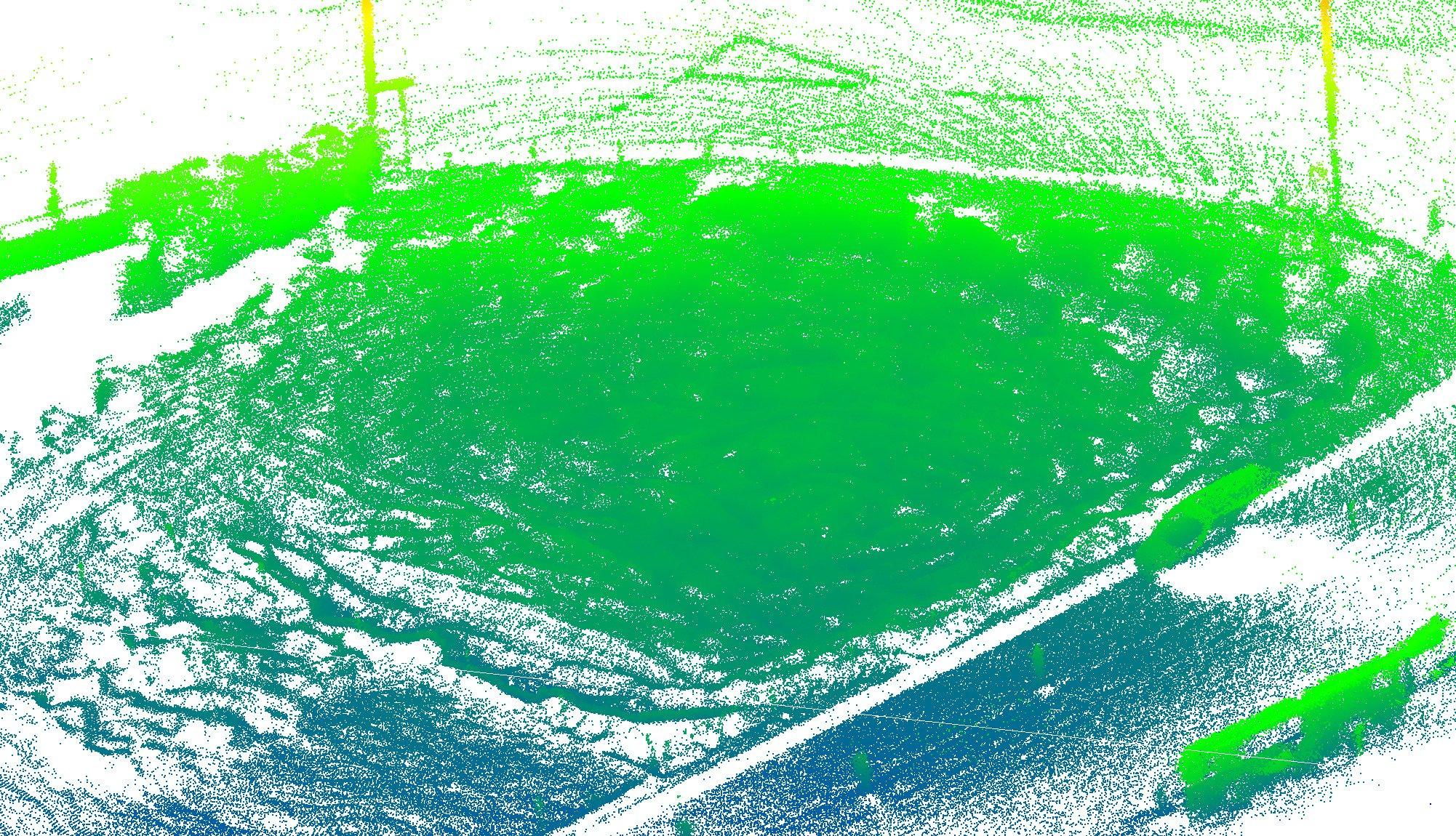}
\end{minipage}
}
\subfloat[Elevation and offline traversability map]{
\begin{minipage}{0.48\textwidth}
\includegraphics[height=4cm, width=\linewidth]{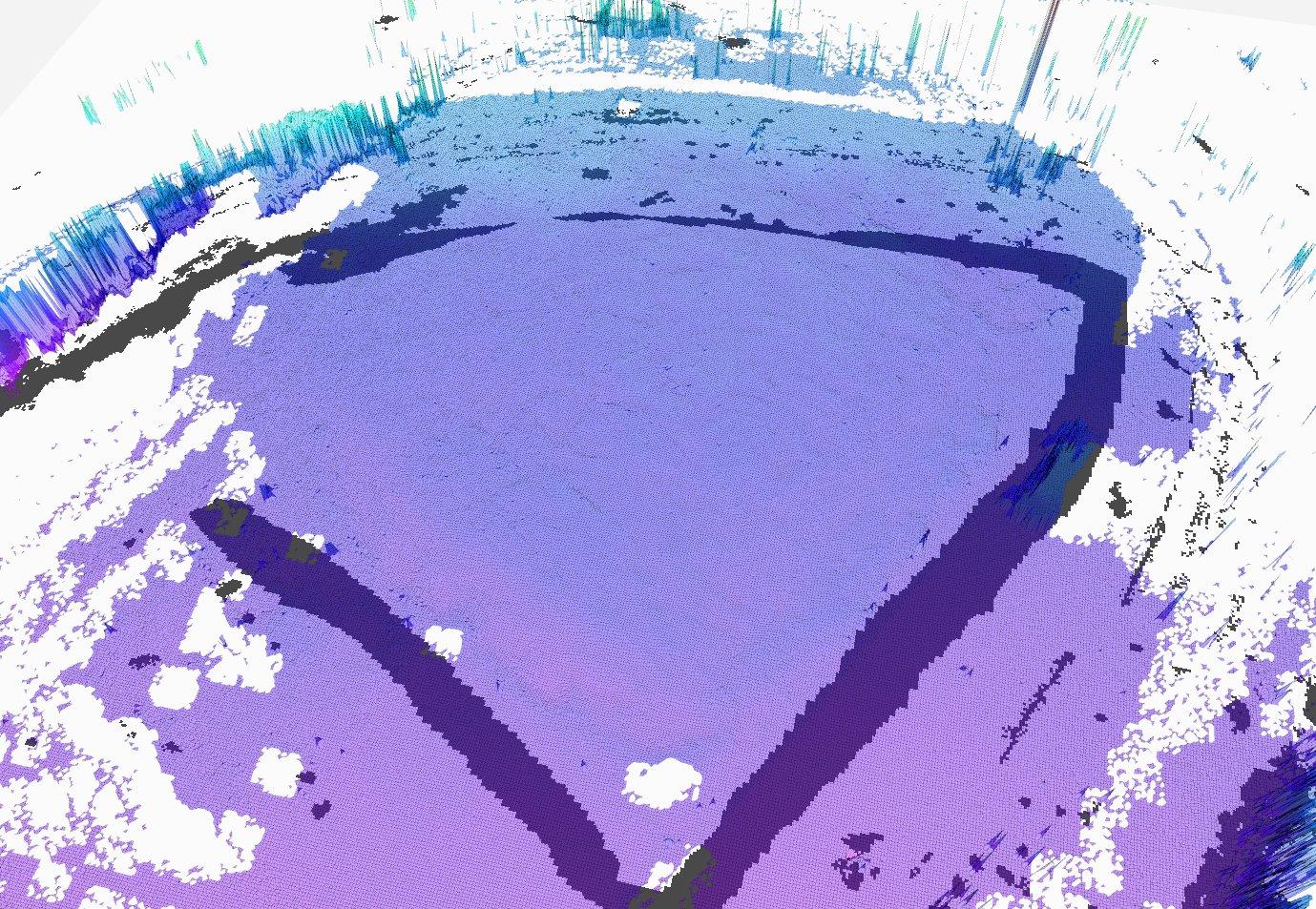}
\end{minipage}
}
\caption{\textbf{(a)}: Point cloud generated with Open3D SLAM of the excavation site. \textbf{(b))}: Elevation map (in blue) and traversability map (black non-traversable) generated from the point cloud map. The traversability has been manually modified to include the fence, which is not visible via the LIDAR.}
\label{fig:map}
\end{figure}

We then transformed the point cloud map into a 2.5D elevation map at a resolution of 0.1 m using the grid map library \cite{jelavicRoboticPrecisionHarvesting2022}, as depicted in \autoref{fig:map}. This map representation is consistently used throughout the paper. We applied a hole-filling filter to the map to counteract minor occlusions from the LIDAR's field of view due to soil irregularities.

\paragraph{User Input}
The system requires several input layers in the form of grid maps. These layers detail the excavation area, the desired depth, the locations of obstacles, and the designated dump sites. Each layer functions as a mapping $f(x, y) \rightarrow z$, where $(x, y)$ represents a position on the map and $z$ is the value of the layer at that position. In the context of an elevation map, $z$ corresponds to the terrain's elevation.

\begin{figure} [!hbt]
  \subfloat[Google Earth Pro input]{
  \begin{minipage}{0.48\textwidth}
  \includegraphics[height=4cm,width=\linewidth]{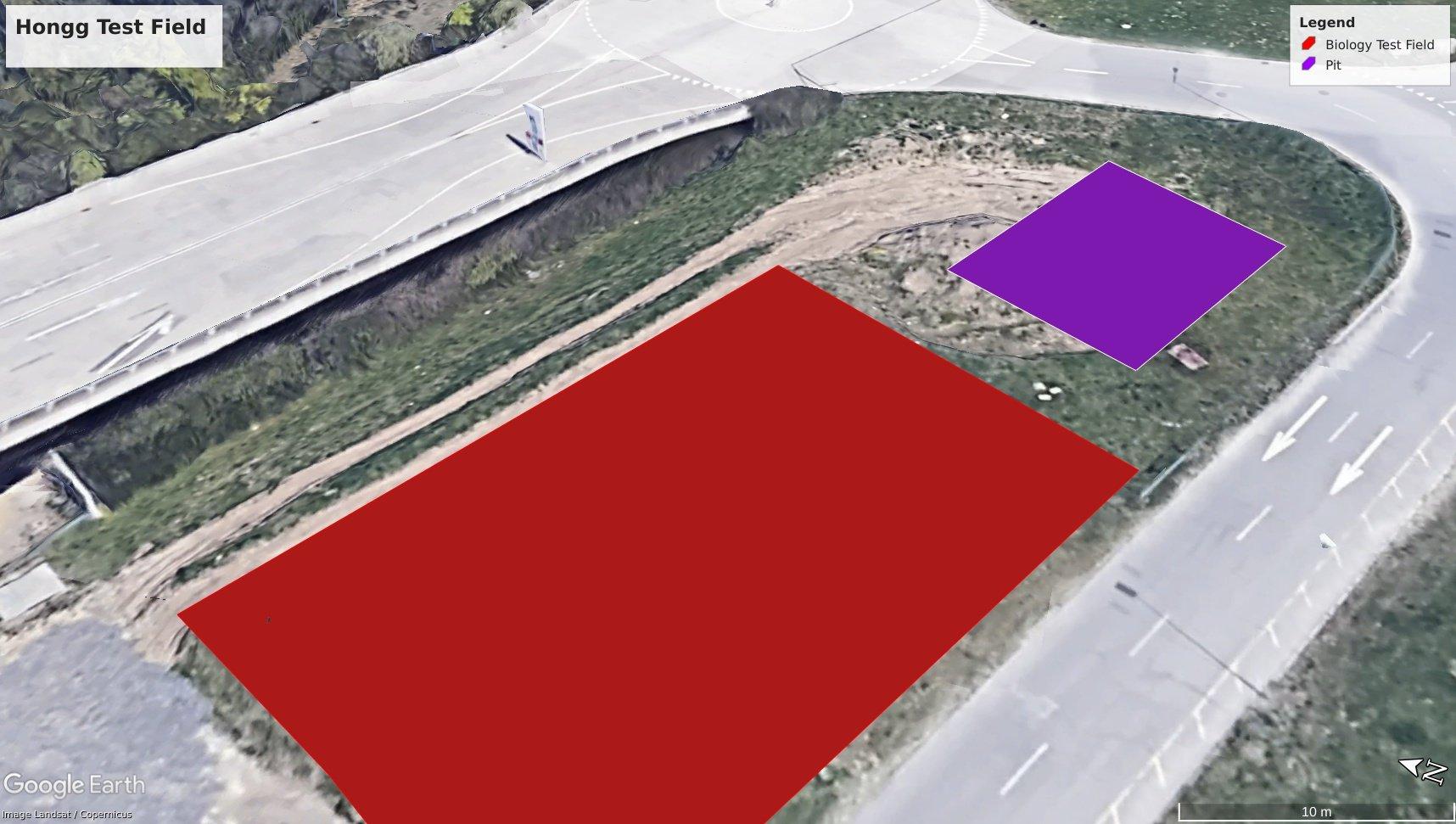}
  \end{minipage}
  }
  \subfloat[Target elevation]{
  \begin{minipage}{0.48\textwidth}
  \includegraphics[height=4cm,width=\linewidth]{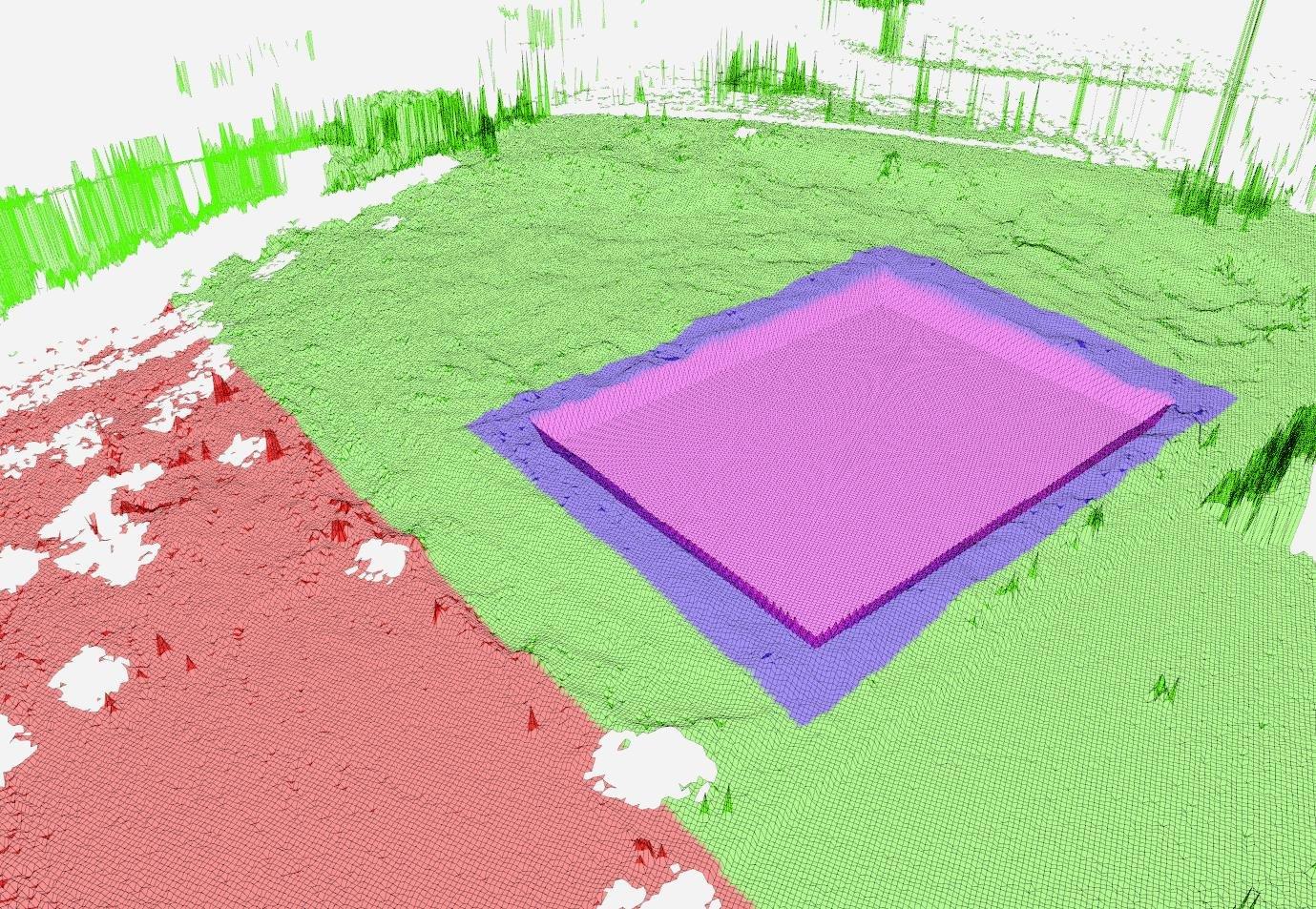}
  \end{minipage}
  }
  \caption{\textbf{(a)} Users use Google Earth Pro to delineate various layers. No-go zones appear in red, dirt dumping restrictions are marked in red, and excavation zones are indicated in violet. Excavation depths are assigned to these polygons. \textbf{(b)} The target elevation is color-coded according to the excavation mask layer. Red indicates no-dumping zones, green shows allowed dump sites, blue marks the excavation area boundary, and violet represents the excavation area.}
  \label{fig:gearth}
\end{figure}

Google Earth Pro serves as the tool for defining these layers. It allows users to create georeferenced shapes, allocate attributes, and specify elevation at each vertex. It also provides a 3D and top-down view of the excavation site. The user-defined layers are merged to produce a unified grid map, which the excavation planning system uses (see \autoref{fig:gearth}). The integrated map comprises the following layers:

\begin{itemize}
\item \textbf{Elevation}: This is sourced from a point cloud map, representing the terrain's height.
\item \textbf{Target elevation}: Defined in Google Earth Pro, this layer marks the intended excavation depth. Users can set this as absolute values or relative to the existing ground level.
\item \textbf{Excavation mask}: A layer with integer values showing both the excavation and dump sites.
\item \textbf{Occupancy}: A binary layer detailing terrain traversability, the presence of obstacles, and off-limits areas.
\end{itemize}

\paragraph{State Estimation}
To estimate the robot's state, we use a graph-based multi-sensor fusion approach \cite{nubertGraphbasedMultisensorFusion2022}. The state estimator fuses IMUs, encoders, RTK-GPS, and LIDAR measurements. The graph-based fusion of the LOAM-based LIDAR odometry, as described in CompSLAM \cite{khattakComplementaryMultiModal2020}, together with the RTK-GPS measurements allows the system to be robust to connectivity and GPS outages. A robust state estimator is crucial for the machine to operate reliably for several hours.

\subsection{Global Excavation Planner}
\begin{figure}
  \subfloat[Local Excavation Geometry]{
    \begin{minipage}{0.48\textwidth}
      \centering
      \includegraphics[width=0.9\linewidth]{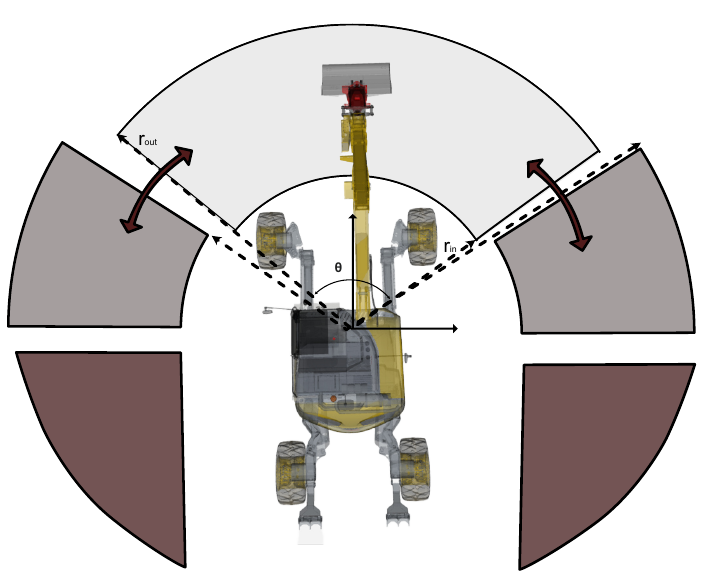}
      \label{fig:local_geometry}
    \end{minipage}
  }
  \subfloat[Coverage Paths]{
    \begin{minipage}{0.48\textwidth}
      \centering
      \includegraphics[width=\textwidth]{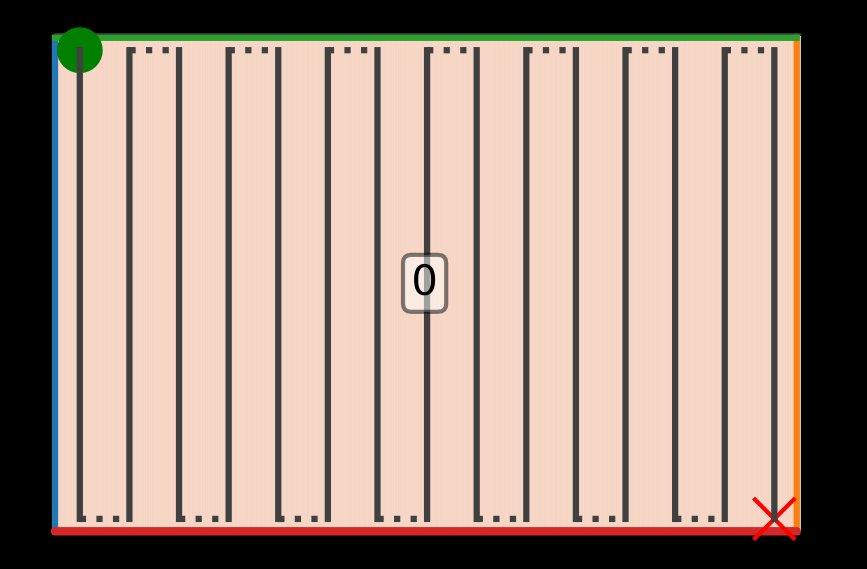}
      \label{fig:paths}
    \end{minipage}
  }
  \caption{\textbf{(a)}: Local excavation geometry, showing the five planning areas. $\theta$ is the workspace angle and has a value of 1.9 rad, $r_in$ and $r_{out}$ are the digging area inner and outer radius equal to 7.0 and 4.5 m. The inner and outer radii of the front-left and front-right areas are 7.5 and 3.5 m. \textbf{(b)}: coverage paths generate for a rectangular workspace. The green dot represents the starting point of the excavation, and the red cross is the endpoint. The dotted lines show the quickest way to join two excavation lines.}
  \label{fig:overview}
\end{figure}

The global excavation planner determines the sequence of poses for the excavator base to ensure the entire excavation area is within arm's reach. The planner design is grounded on the local excavation geometry shown in Figure \ref{fig:local_geometry}. The local digging geometry, illustrated in Figure \ref{fig:paths}, defines the excavator's accessible digging area based on its current base pose.

\paragraph{Overview}
The global planner accepts an excavation map from the user, encoding target dig and dump zones and obstacles. The system includes an outer and inner optimization loop. The inner optimization involves decomposing the dig space into coverable cells with Boustrophedon decomposition, generating a directed quotient graph over the cells, and finding a minimum branching spanning tree to minimize inter-cell movement. Post-order traversal of the tree ensures non-traversal of dug cells and maintains access to remaining cells. Dynamic programming is used to identify the cell corner sequence for minimum travel distance. The outer loop selects the coverage orientation to maximize the covered area while minimizing local workspaces and path length.

\begin{figure}
  \captionsetup[subfloat]{farskip=0pt,captionskip=0pt}
  
  \begin{minipage}[t]{0.45\textwidth}
    \centering
    \subfloat[Example with convex obstacle]{%
      \includegraphics[scale=0.6]{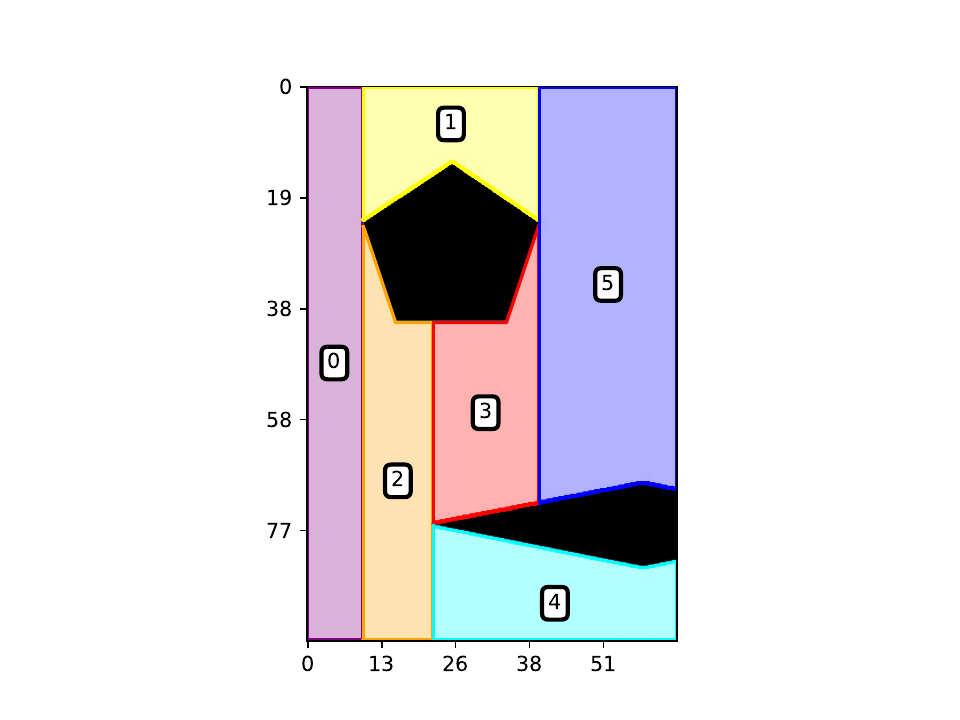}
      \label{fig:bou_decom_convex}
    }
    
    \vfill

    \subfloat[Quotient Graph (Convex)]{%
      \includegraphics[width=0.8\linewidth]{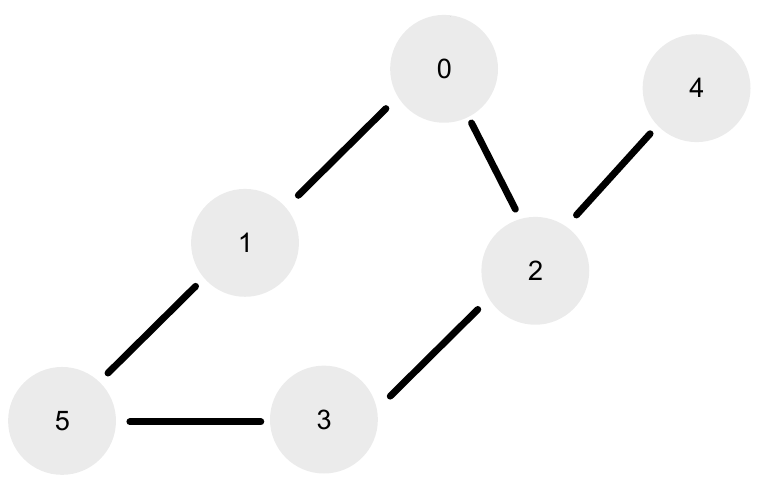}
      \label{fig:quotient_graph}
    }
    
    \vfill

    \subfloat[Minimum Branching Spanning Tree (Convex)]{%
      \includegraphics[width=0.7\linewidth]{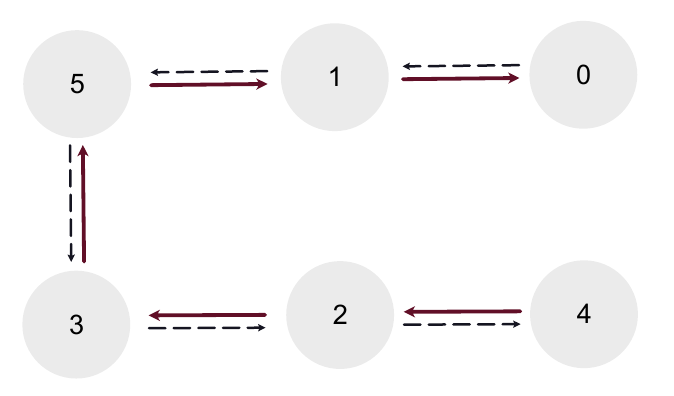}
      \label{fig:spanning_tree_convex}
    }
  \end{minipage}
  \hfill
  \vrule
  \hfill
  \begin{minipage}[t]{0.45\textwidth}
    \centering
    \subfloat[Example with concave obstacle]{%
      \includegraphics[width=\linewidth]{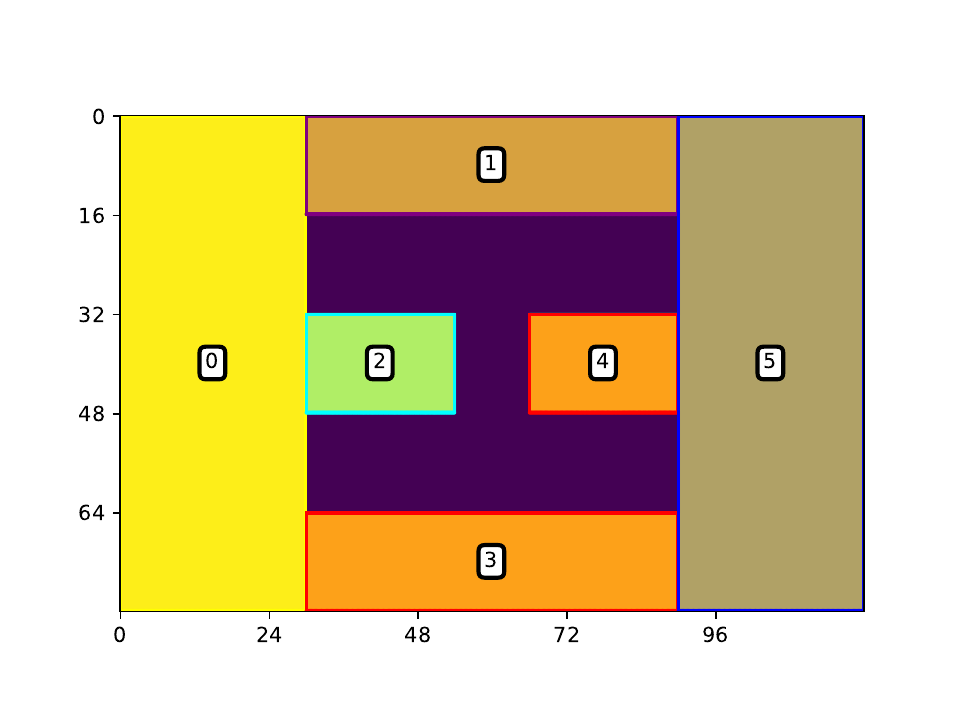}
      \label{fig:bou_decom_concave}
    }
    
    \vfill

    \subfloat[Quotient Graph (Concave)]{%
      \includegraphics[width=0.7\linewidth]{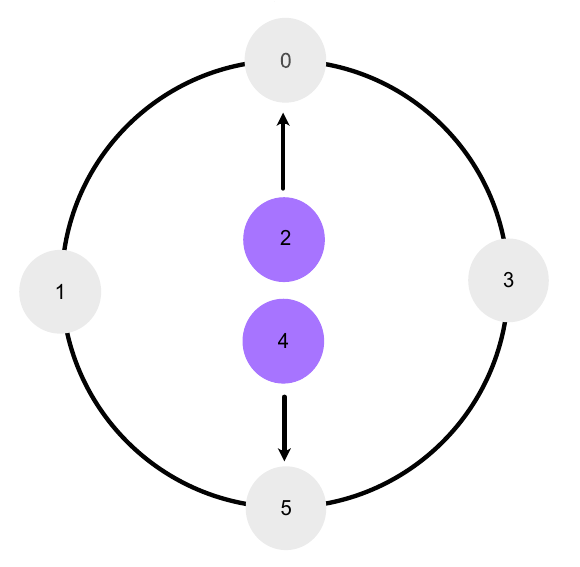}
      \label{fig:quotient_graph_concave}
    }
    
    \vfill

    \subfloat[Minimum Branching Spanning Tree (Concave)]{%
      \includegraphics[width=0.5\linewidth]{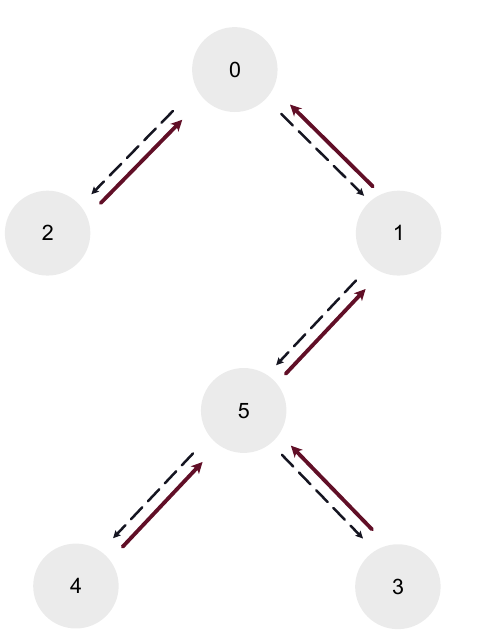}
      \label{fig:spanning_tree_concave}
    }
  \end{minipage}

  \caption{Workspace decomposition and cell visitation workflow, colors have been added just for easy visualization. \textbf{Top}: Two examples of Boustrophedon decomposition. The space has been subdivided into different cells, each marked with a different color. \textbf{Center}: Corresponding quotient graph to the two examples. For the left column case, the graph is undirected since all cells share one with at least another cell in one corner; for the right column, no cell shares a corner with cells 2 and 4. Therefore these two cells must be excavated before cells 0 and 5, respectively. The simply connected components of the graph are indicated with a different color. \textbf{Bottom}: Resulting in minimum branching spanning trees for the above graphs. A dashed edge indicates that the machine moved to the cell, while a solid edge indicates that the machine excavated the cells.}
  \label{fig:decomposition_workflow}
\end{figure}

\paragraph{Boustrophedon Decomposition and Quotient Graph}
The first step in the process is to decompose the target space using the Boustrophedon decomposition \cite{chosetCoverageKnownSpaces2000} based on a specific coverage orientation. During this decomposition, the space is segmented into different cells as a slice passes over the workspace, altering its connectivity. Figure \ref{fig:bou_decom_convex} and Figure \ref{fig:bou_decom_concave} provide examples of such decomposition, demonstrating a vertical coverage orientation.
Following the decomposition, the cells' connectivity is represented through a graph, where each node symbolizes one of a cell's four extreme points. The adjacency matrix is then constructed using an equivalence relationship between two vertices $v_1$ and $v_2$, as defined in Equation \ref{eq:equivalence}.
\begin{equation}\label{eq:equivalence}
    v_1 \sim v_2 \iff \exists c \in C, v_1 \in c, v_2 \in c
\end{equation}
where $C$ signifies the set of cells. Subsequently, the quotient graph is generated by collapsing all equivalent vertices, with the edges connecting cells sharing at least one common vertex (see Figure \ref{fig:quotient_graph}). The graph is bidirectional in scenarios involving only convex obstacles, except when the obstacles align with the sweeping slice's direction. Conversely, the graph is directed when one vertex shares corners with another, but not vice versa (e.g., when the obstacle is concave, and the sweeping slice passes through its concave part), as illustrated in Figure \ref{fig:quotient_graph_concave}.

\paragraph{Spanning Tree and Topological Sorting}
The excavation problem's structure presents practical challenges that prevent a direct formulation as a Travelling Salesman Problem (TSP) or its variants. In this context, multiple cell visits are allowed and might be necessary, but a dug cell becomes unavailable for navigation. Furthermore, starting and ending at the same cell is not a requirement.

We address this problem by constructing a spanning tree of the graph, using post-order traversal to ensure full tree traversal. This strategy enables the excavator to dig the entire area, revisiting undug cells and minimizing driving time by reducing the number of branches in the tree. A tree with no vertex with a branching degree greater than one allows continuous digging. The problem, known as the Minimum Branch Vertices Problem (MBVP), is efficiently solvable via dynamic programming for both directed and undirected graphs \cite{silvestriBranchandcutAlgorithmMinimum2017}. After generating the spanning tree, the sequence of cells to visit is determined through post-order traversal, with the results depicted in Figures \autoref{fig:spanning_tree_convex} and \autoref{fig:spanning_tree_concave}.

\begin{figure}[!hbt]
    \centering

    \begin{subfloat}[Alternating travel direction]{\includegraphics[width=0.3\linewidth]{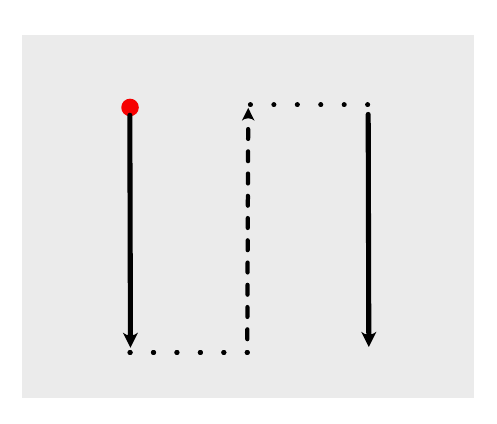}\label{subfig:lane1}}
    \end{subfloat}\hfill
    \begin{subfloat}[Same travel direction]{\includegraphics[width=0.31\linewidth]{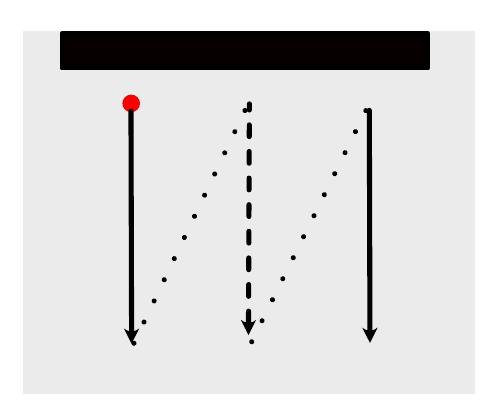}\label{subfig:lane2}}
    \end{subfloat}\hfill
    \begin{subfloat}[Coverage with obstacles]{\includegraphics[width=0.3\linewidth]{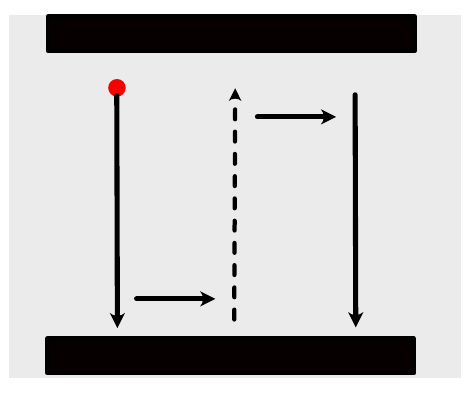}\label{subfig:lane3}}
    \end{subfloat}

    \caption{Three possible subroutines used to cover a cell. \textbf{(a)}: alternating lane directions with obstacles lying only one side. \textbf{(b)}: two consecutive lanes having the same direction. \textbf{(c)}: coverage with obstacles on both sides present.}
    \label{fig:subroutines}
\end{figure}

\begin{figure}[!hbt]
  \subfloat[Alternating travel direction]{
    \begin{minipage}{0.31\textwidth}
      \centering
      \includegraphics[width=0.7\linewidth]{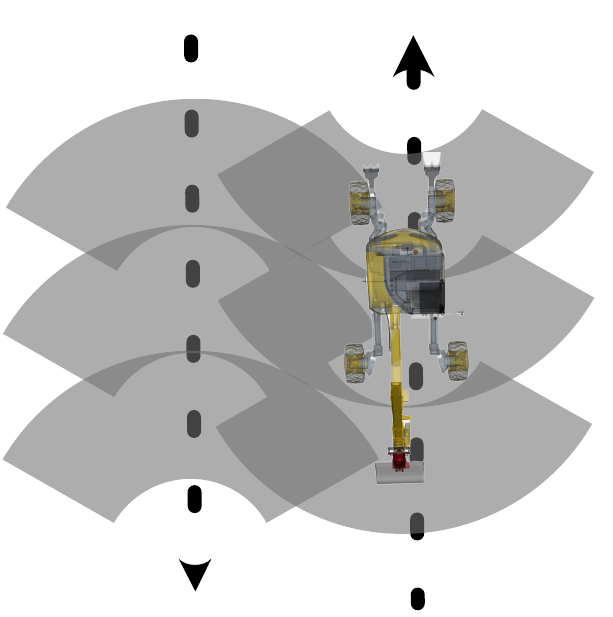}
      \label{fig:alternating_lanes}
    \end{minipage}
  }
  \subfloat[Same travel direction]{
    \begin{minipage}{0.31\textwidth}
      \centering
      \includegraphics[width=0.7\linewidth]{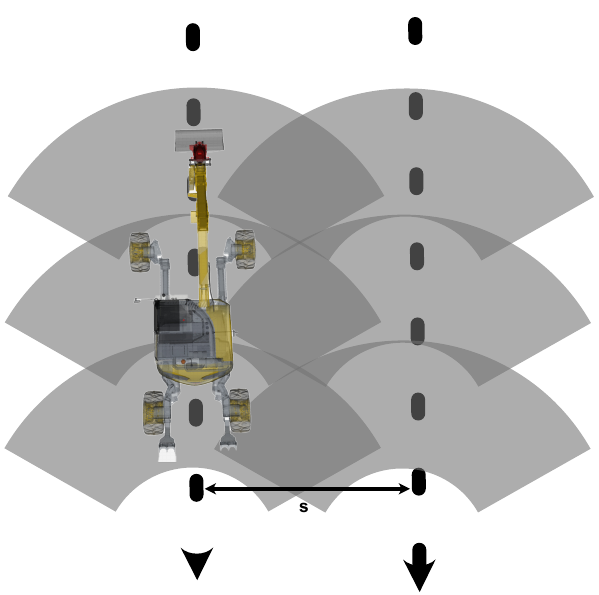}
      \label{fig:same_lanes}
    \end{minipage}
  }
  \subfloat[Coverage with obstacles]{
    \begin{minipage}{0.31\textwidth}
      \centering
      \includegraphics[width=0.7\linewidth]{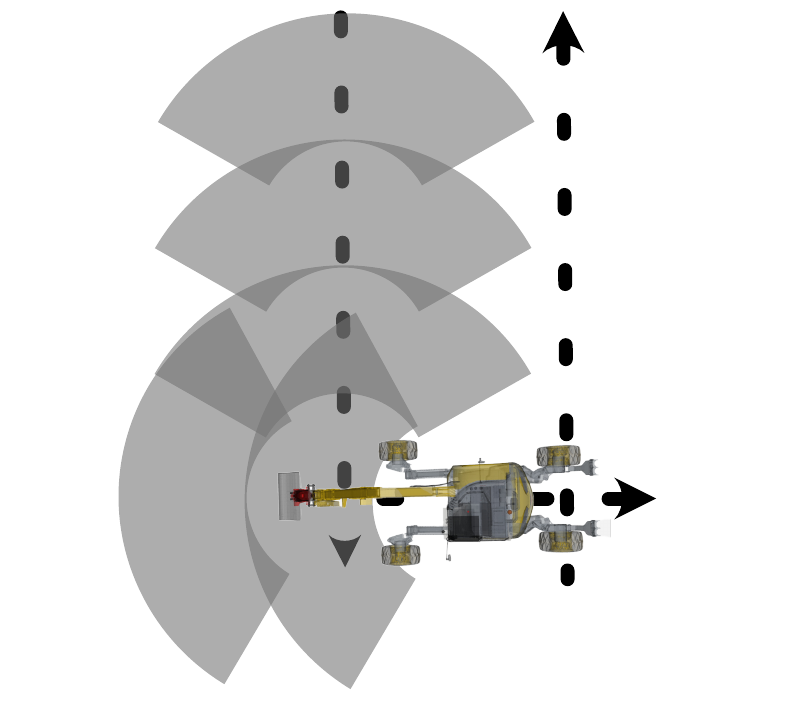}
      \label{fig:obstacle_coverage}
    \end{minipage}
  }
  \caption{Coverage geometry, where the excavator always moves backward, corresponding to the subroutines shown in Figure \ref{fig:subroutines}. The worst-case alignment of the local workspace is used to determine the lane spacing.}
  \label{fig:lanes}
\end{figure}

\paragraph{Dynamic Programming for Local Coverage}
With the sequence of cells determined, the problem now focuses on selecting optimal corners to visit and the associated coverage subroutine. Three potential subroutines connect the vertices of a cell (Figure \autoref{fig:subroutines}), creating a series of spaced base poses (Figure \autoref{fig:lanes}). The distance covered during each routine is the sum of distances between successive base poses, with an extra cost term accounting for the turns, approximating the additional distance needed for maneuvering. Should the robot's footprint collide with any obstacle, the path length is set to infinity; otherwise, distances in undug areas are approximated as the Euclidean distance between points.

The problem of finding the optimal sequence of corners to visit can also be efficiently solved using dynamic programming. The complexity of this approach is $O(C V^2)$, where $C$ represents the number of corners and $V$ represents the number of subroutines available.

We use the following recurrence relation shown in \autoref{eq:dynamic_programming} to solve the problem using dynamic programming.

\begin{equation}
    \label{eq:dynamic_programming}
    D_{i,c_k} = 
        \min_{c_n,c_j,l} \left( D_{i - 1,c_n} +  d_{i, l}(c_n,c_j) + d_{o}(c_j, c_k)  \right)
\end{equation}

In this equation, $D_{i,c_k}$ represents the minimum distance required to reach corner $c_k$ in cell $i$. The term $d_{i, l}(c_n,c_j)$ denotes the length of the coverage path between corners $c_n$ and $c_j$ in cell $i$. The coverage path can have a flipped lane in the last position, controlled by the binary variable $l$. Finally, $d_{o}(c_j, c_k)$ signifies the length of the path between corners $c_j$ of cell $i$ and $c_k$ in cell $i + 1$.

\begin{figure}[!hbt]
  \subfloat[]{
    \begin{minipage}{0.44\textwidth}
      \centering
      \includegraphics[scale=0.7]{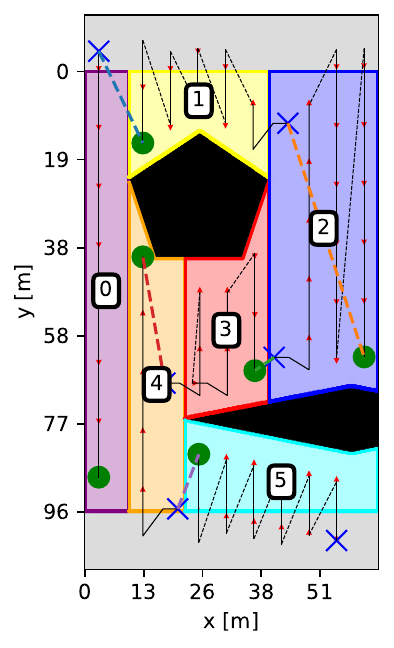}
      \label{fig:path_convex}
    \end{minipage}
  }
  \subfloat[]{
    \begin{minipage}{0.54\textwidth}
      \centering
      \includegraphics[width=\linewidth]{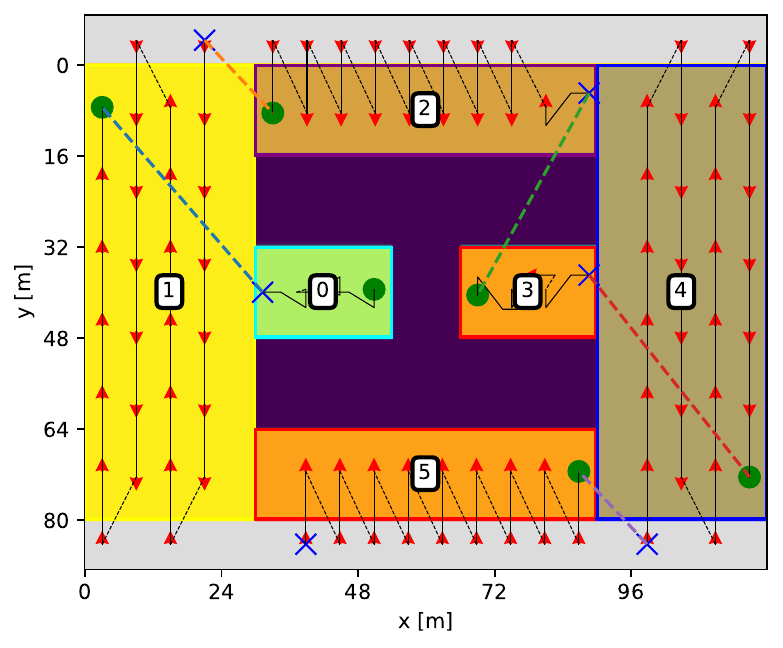}
      \label{fig:path_concave}
    \end{minipage}
  }
  \caption{Coverage paths for the cases considered in \autoref{fig:decomposition_workflow}. The excavation for each cell starts at the green dot and ends at the red cross. Solid lines indicate the excavation path, dotted lines indicate the forward orientation of the base along the path navigation to the next line. The cells have been re-indexed to reflect the coverage order.}
  \label{fig:solutions}
\end{figure}

In \autoref{fig:solutions} are shown the solutions to the global coverage problem for the cases we have considered.

\paragraph{Outer Optimization} 

The excavation layer is initially divided into locally connected areas. The global planning problem is solved for each area, and a global path is created by concatenating the paths of these areas based on the solution to a traveling salesman problem. The choice of orientation for each connected area significantly influences the excavation plans, including the number of cells, workspaces, and the path's feasibility. 

The best orientation for a given excavation geometry is found by solving the following nonlinear optimization problem:
\begin{equation}
    \theta = \arg \min J(\theta) = c_{\text{ax}} (\theta - \phi) + c_p L_p + c_n N_w + c_a A_c
\end{equation}

where $\phi$ is the main axis of the excavation object, $L_p$ is the length of the coverage path, $N_w$ is the number of workspaces, and $A_c$ is the fraction of covered area. Depending on the excavation type, the user can tune the coefficients accordingly. For example, in the case of trenching, the only coefficient that should be non-zero is the main axis coefficient $c_{\text{ax}}$ because this creates a coverage plan aligned with the main direction of the trench.



\subsection{Volumetric Simulation of Earth Movement} \label{subsec:kinematic_soil}
This section describes our method to simulate soil movement for system simulation. We represent the soil as a simplified geometric solid volume, excluding dynamics, as our primary focus is excavation planning.

To mimic digging, we track the shovel's path, removing soil without considering the forces on the shovel's edge. Mass conservation is maintained by calculating the volume of soil removed with each scoop and adding an equivalent volume to the dumping location.

Our simulation models the falling soil particles and calculates their ground distribution. This is achieved by dividing the space into "bucket slices," assuming that particles in each slice fall according to a normal distribution independent of one another. The mean of this distribution is the slice's center, and the standard deviation is half the bucket's width. Additionally, we assume that the dumping location is relatively flat, preventing lateral soil sliding post-dumping.

The resulting ground height distribution is described by \autoref{eq:height_soil} and \autoref{eq:covariance}, where variables are defined as follows: $\sigma_x$ and $\sigma_y$ are half the shovel dimensions in the body frame along the x and y axes, $\psi$ is the shovel's heading angle in the world frame, $N$ is the number of discretized elements in the bucket, and $V_b$ is the bucket's volume.

\begin{equation}
  \label{eq:height_soil}
h(x) =  \frac{V_b}{N} \sum_{i = 0}^N \frac{1}{(2\pi)^{n/2}|\Sigma|^{1/2}} \exp\left(-\frac{1}{2}(x-\mu_i)^T\Sigma^{-1}(x-\mu_i)\right)
\end{equation}
\begin{equation}
    \label{eq:covariance}
    \Sigma =   \begin{bmatrix}
      \cos(\psi) & -\sin(\psi) \\
      \sin(\psi) & \cos(\psi)
    \end{bmatrix} \begin{bmatrix}
        \sigma_x^2 & 0 \\
        0 & \sigma_y^2
    \end{bmatrix}
\end{equation}

\begin{figure}[!hbt]
  \subfloat[]{
    \begin{minipage}{0.48\textwidth}
      \includegraphics[height=3.5cm, width=\linewidth]{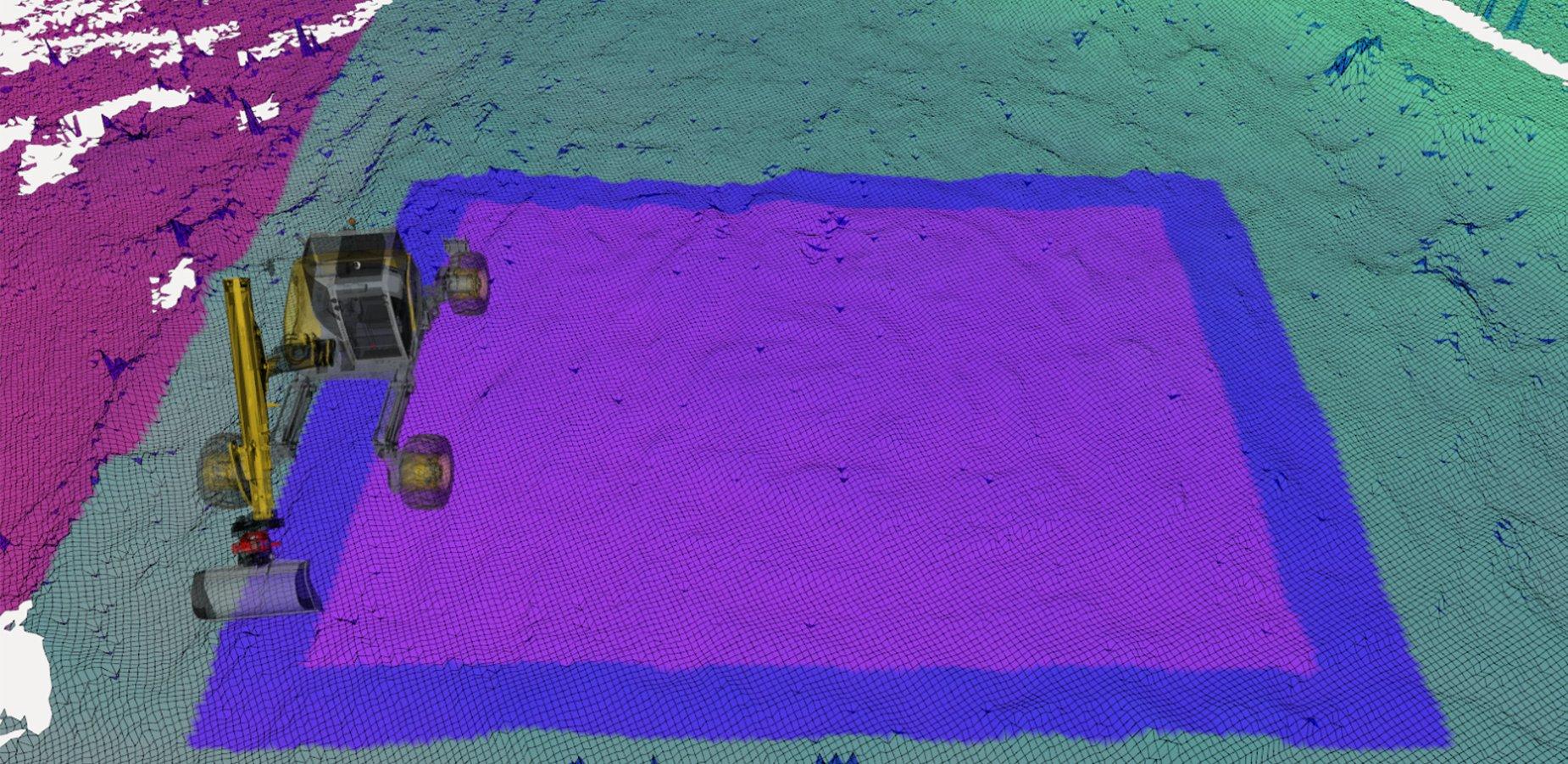}
      \label{fig:start_excavation}
    \end{minipage}
  }
  \subfloat[]{
    \begin{minipage}{0.48\textwidth}
      \includegraphics[height=3.5cm, width=\linewidth]{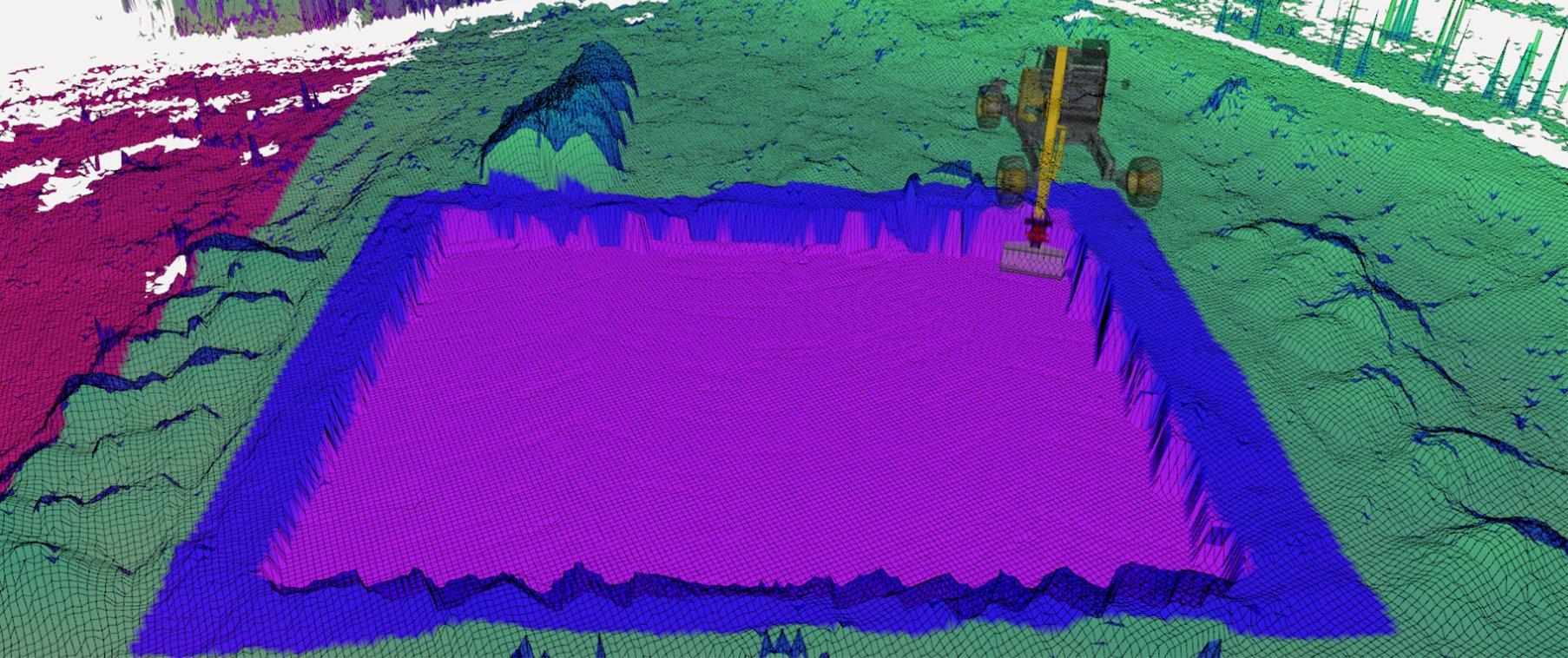}
      \label{fig:end_excavation}
    \end{minipage}
  }
  \caption{Elevation maps generated while excavating in simulation. \textbf{(a)}: Elevation map at the beginning of the excavation. \textbf{(b)}: Elevation map after the excavation is finished.}
  \label{fig:excavation_sim}
\end{figure}

In \autoref{fig:excavation_sim}, two snapshots of a pit's excavation in our simulated test field are shown. The model's main goal was to expedite the system's overall development through simulation, not to create a realistic representation of soil transport. Validating the model using real-world data is challenging since factors such as water content, soil type, and compaction are not considered.

  

\subsection{Local Excavation Planner}
The local excavation planner arranges soil around the excavator's base to fulfill the excavation task. Figure \ref{fig:local_geometry} illustrates the local excavation geometry, divided into five areas:
\begin{itemize}
\item \textit{Front}: an unexcavated digging area.
\item \textit{Front-left} and \textit{front-right}: areas for digging or dumping previously dug soil.
\item \textit{Back-left} and \textit{back-right}: areas for dumping soil excavated from other regions.
\end{itemize}

The excavator's rear is kept clear for navigation, enabling operation in constrained spaces. The front areas' sizes can vary with the task.

The planner uses the excavation mask layer in the elevation map, containing constraints and target geometry information.

\paragraph{Excavation Mask} The updated excavation mask layer guides the choice of dig and dump zones based on user input. It includes:
\begin{itemize}
    \item \textit{Dig area} (violet): Identified by a discrepancy between target and current elevations.
    \item \textit{Permanent dump area} (green): For permanent disposal of excavated material.
    \item \textit{Neutral area} (light blue): An area for digging and temporary material dumping, eventually moving to the permanent dump area.
    \item \textit{No-go area} (red): User-defined, no dig or dump area.
    \item \textit{Boundary area} (dark blue): Equivalent to a no-go area at the excavation zone's edge, preventing soil spillage.
\end{itemize}

The neutral area combines user-defined areas, the convex hull of potential future base footprints, and unreachable dump areas from the current machine configuration. It ensures a continuous path to future target poses by keeping material away from this region.

\paragraph{Dig and Dump Zone Selection}
The local planner selects the dig and dump zones by first checking the front zone and then lateral front zones based on the criteria:
\begin{enumerate}
    \item Less than 10\% of elevation map cells deviate more than 0.1 m from the target elevation.
    \item Remaining volume to excavate is less than 8\% of the total volume.
\end{enumerate}
The front zone's target elevation uses the "desired elevation" layer, while the front-left and front-right zones use the "original elevation" layer. Thresholds for these conditions were determined experimentally for a good balance between accuracy and speed.

The dumping zone is chosen from front-left, front-right, back-left, or back-right zones, considering the lowest dumping cost, computed using Equation \ref{eq:dumping_cost}.

\begin{equation}\label{eq:dumping_cost}
C_D = \frac{1}{N} \sum_{i=1}^{N} SDF_{dump}(x_i) + \alpha ||x_{dig} - x_{dump}||^2
\end{equation}

Here, $C_D$ is the dumping cost, calculated as an average signed distance over the number of cells of the zone ($N$) from the closest permanent dump area ($SDF_{dump}$) plus a term for the distance between digging ($x_{dig}$) and dumping locations ($x_{dump}$), weighted by $\alpha=4.0$. This formulation aims to minimize transport time and number of moves before final disposal.

\begin{figure}[!hbt]
  \subfloat[Excavation mask]{
    \begin{minipage}{0.48\textwidth}
      \includegraphics[height=4.2cm, width=\linewidth]{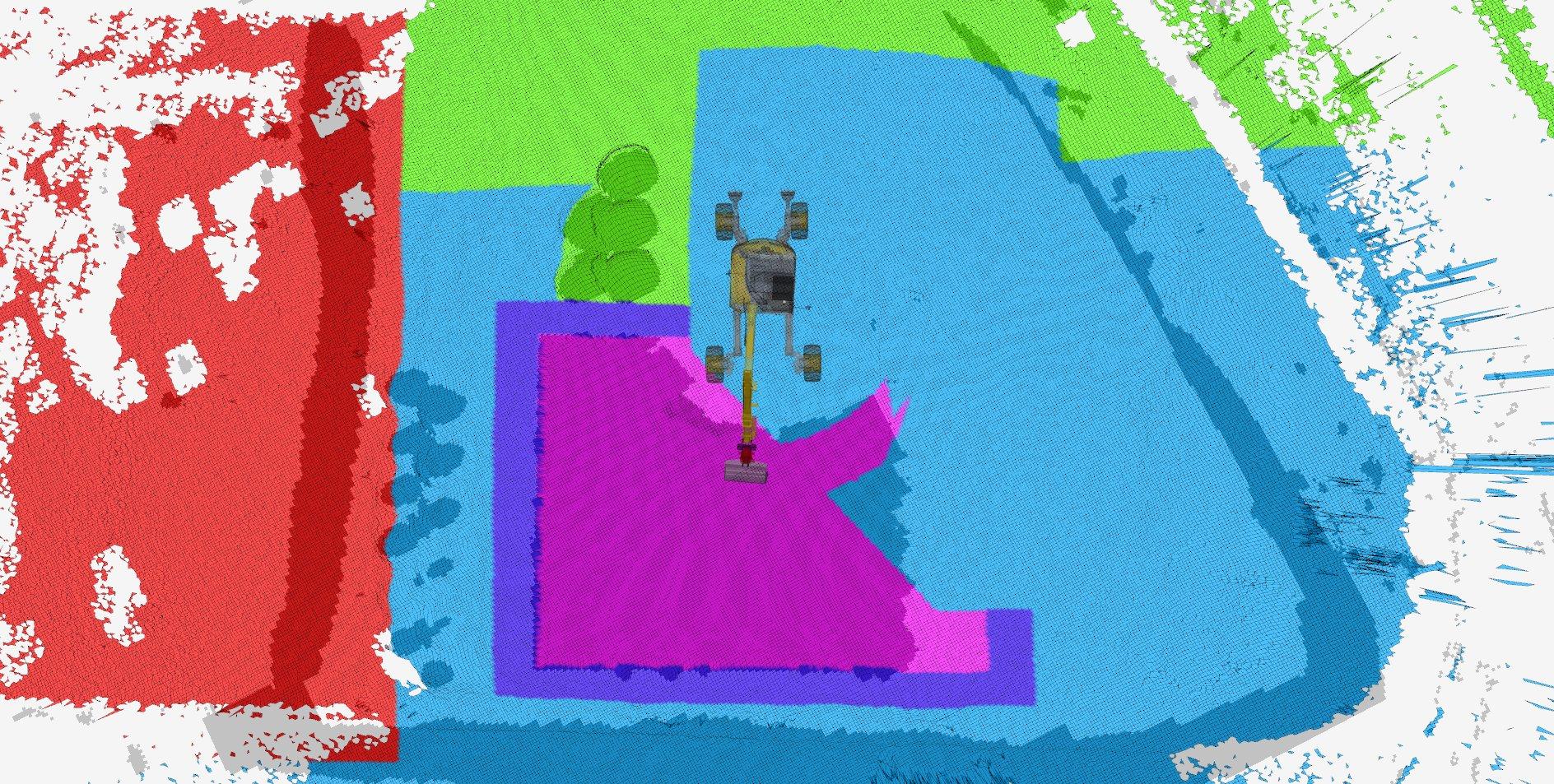}
      \label{fig:current_excavation_mask}
    \end{minipage}
  }
  \subfloat[Dumping cost]{
    \begin{minipage}{0.48\textwidth}
      \includegraphics[height=4.2cm, width=\linewidth]{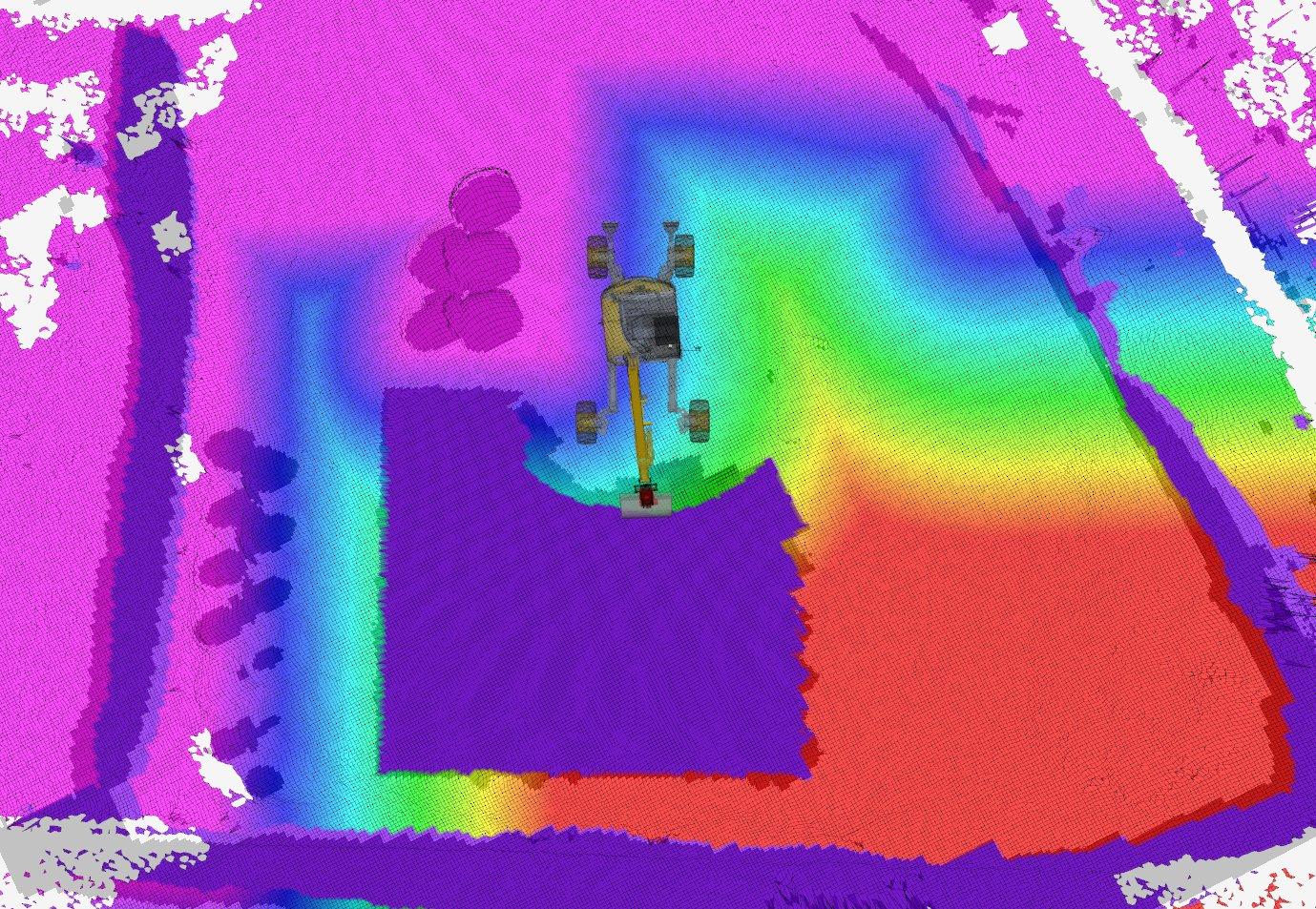}
      \label{fig:dumping_cost}
    \end{minipage}
  }
  \caption{Elevation maps of the terrain during the digging process in simulation. \textbf{(a)}: colored based on the excavation mask following the color scheme defined above. \textbf{(b)}: colored based on the dumping cost, with red indicating the highest cost and violet indicating the lowest cost.}
\end{figure}

\begin{figure}
  \captionsetup[subfloat]{farskip=0pt,captionskip=0pt}
  \centering
  \subfloat[]{
    \begin{minipage}{0.48\textwidth}
      \includegraphics[width=0.95\linewidth]{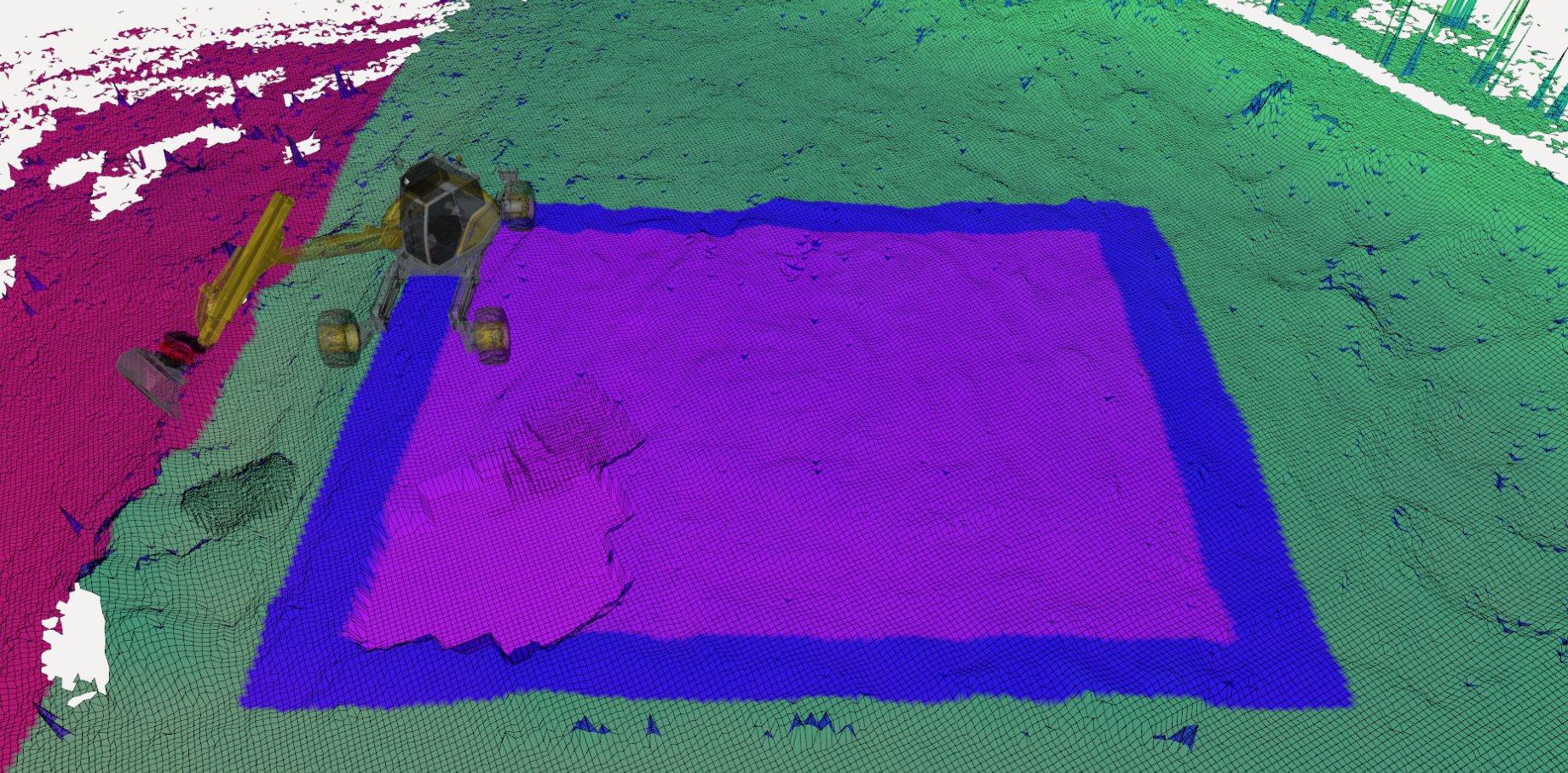}
      \label{fig:original_mask}
    \end{minipage}
  }
  \subfloat[]{
    \begin{minipage}{0.48\textwidth}
      \includegraphics[width=0.95\linewidth]{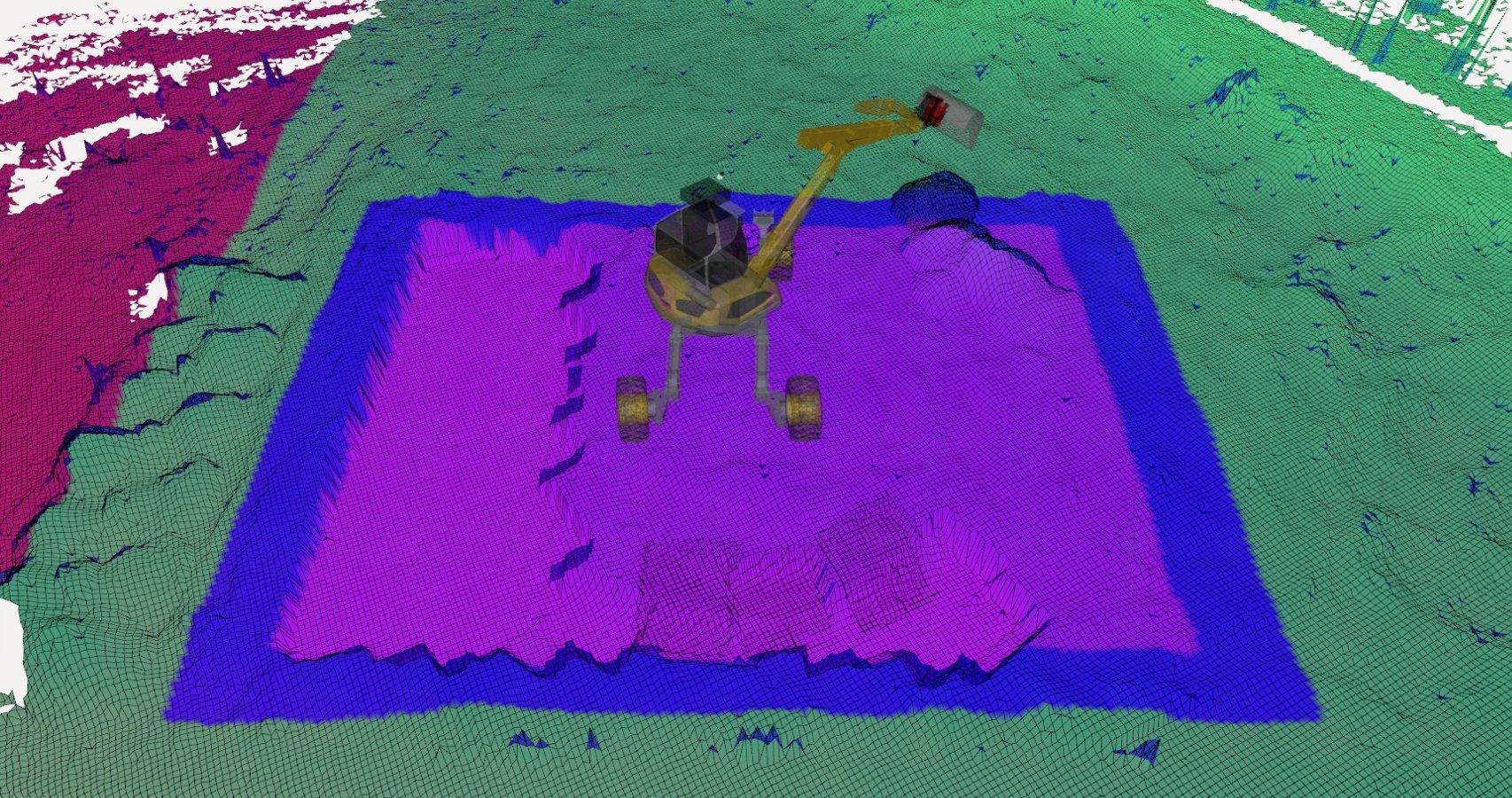}
      \label{fig:side_dump}
    \end{minipage}
  }
  \hfill
  \subfloat[ ]{
      \begin{minipage}{0.48\textwidth}
        \includegraphics[width=0.95\linewidth]{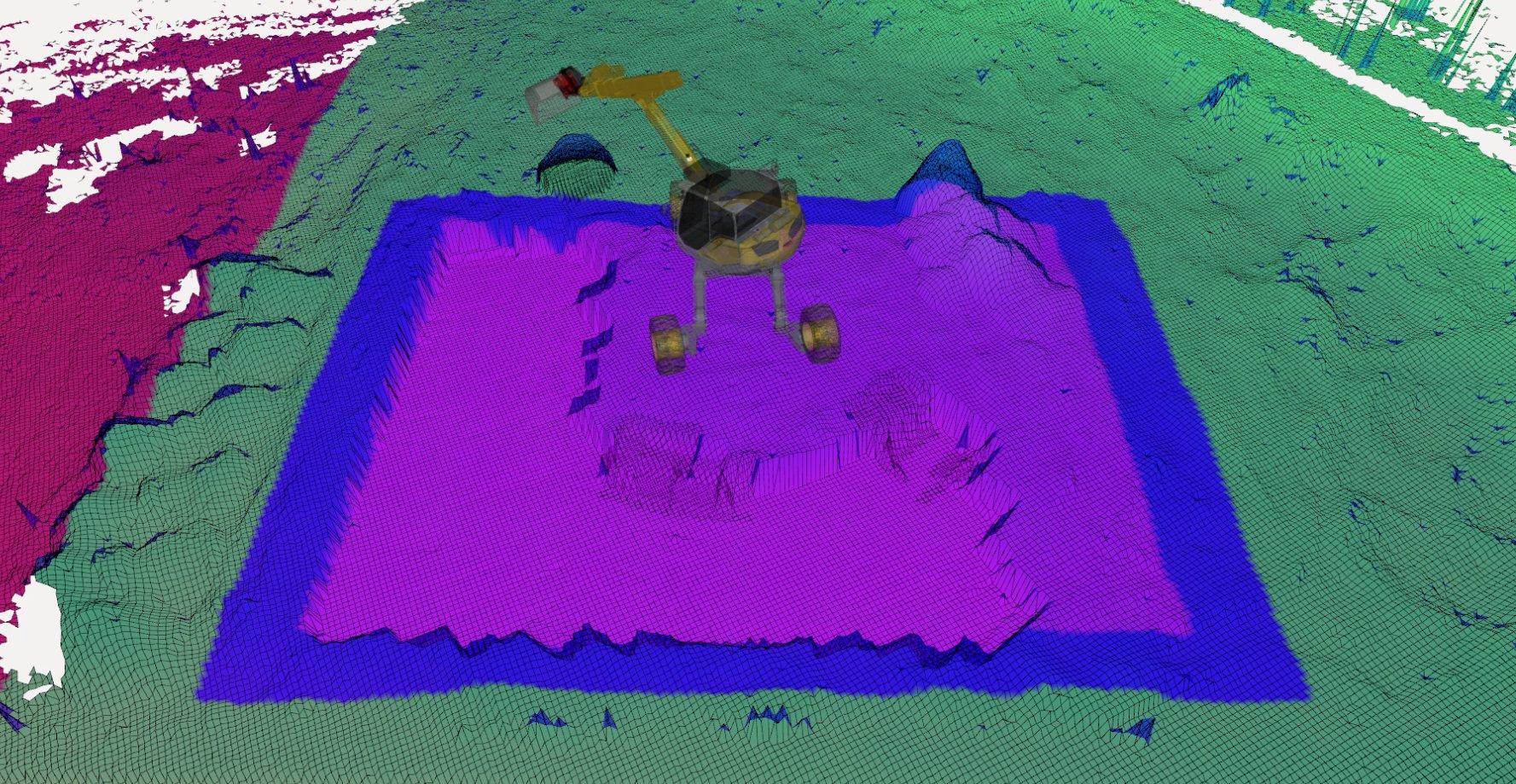}
        \label{fig:central}
      \end{minipage}
    }
    \subfloat[  ]{
      \begin{minipage}{0.48\textwidth}
        \includegraphics[width=0.95\linewidth]{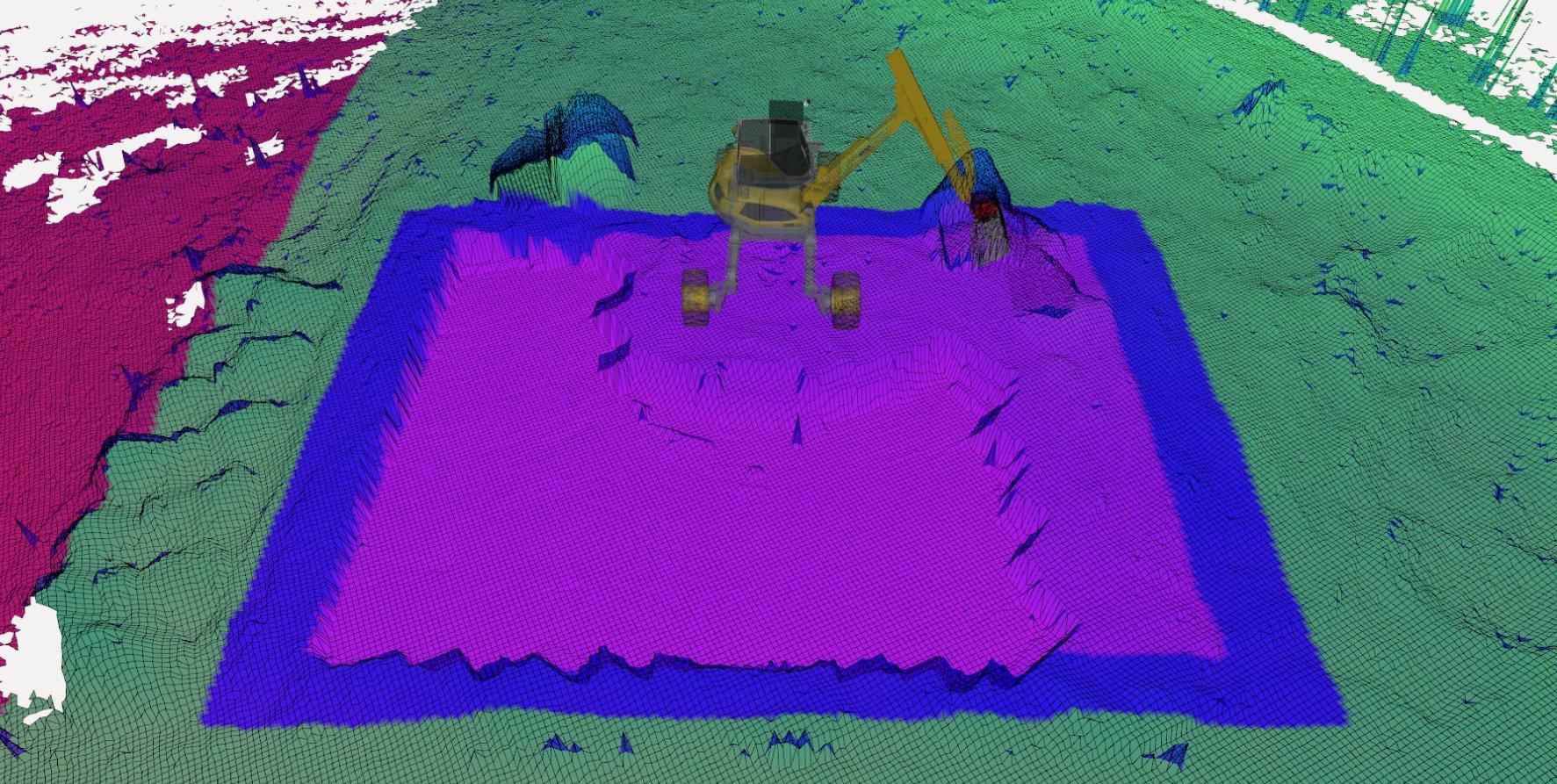}
        \label{fig:central_dirt}
      \end{minipage}
    }
    \hfill
    \subfloat[ ]{
      \begin{minipage}{0.48\textwidth}
        \includegraphics[width=0.95\linewidth]{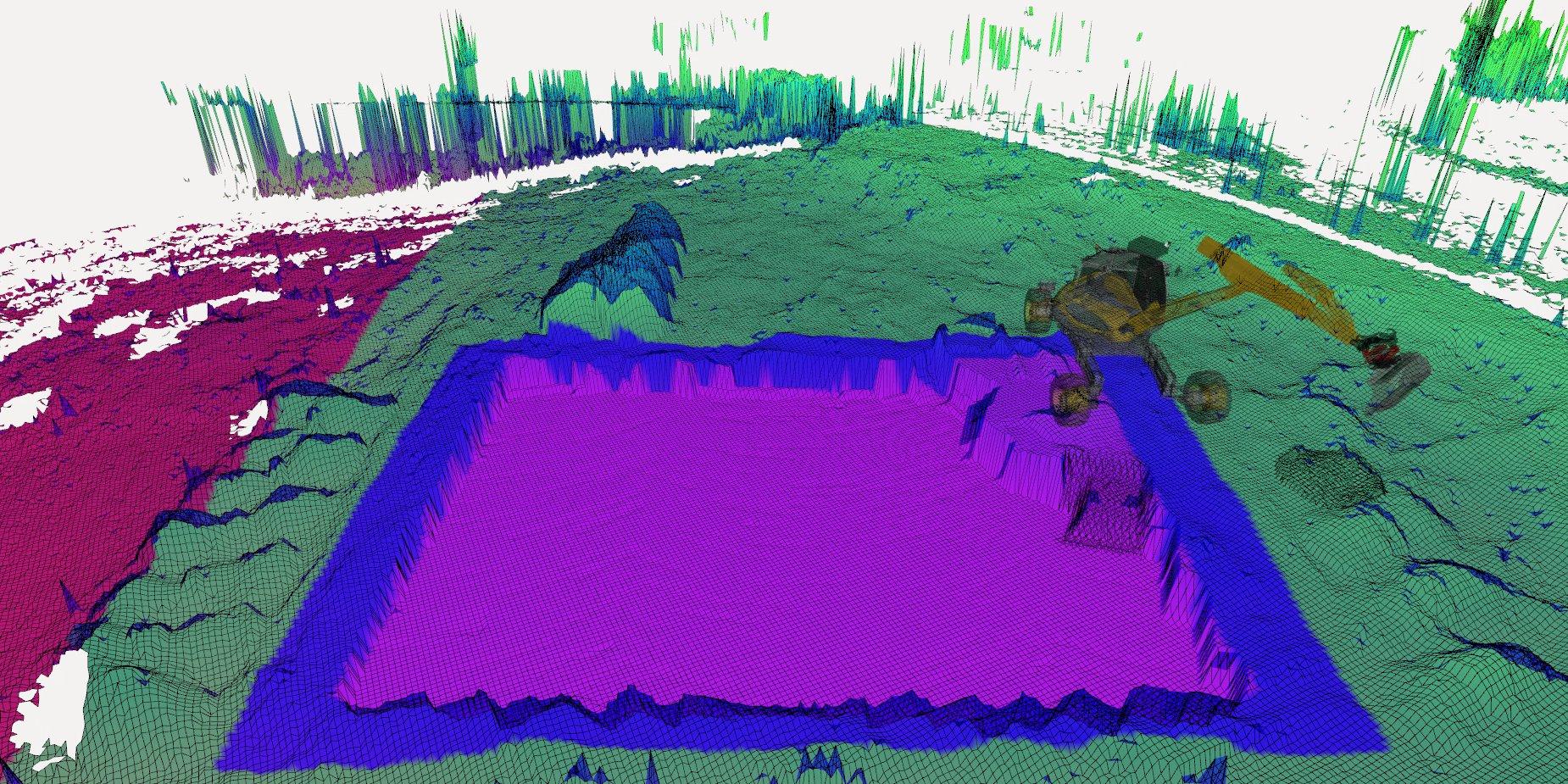}
        \label{fig:central_opposite}
      \end{minipage}
    }
    \subfloat[]{
      \begin{minipage}{0.48\textwidth}
        \includegraphics[width=0.95\linewidth]{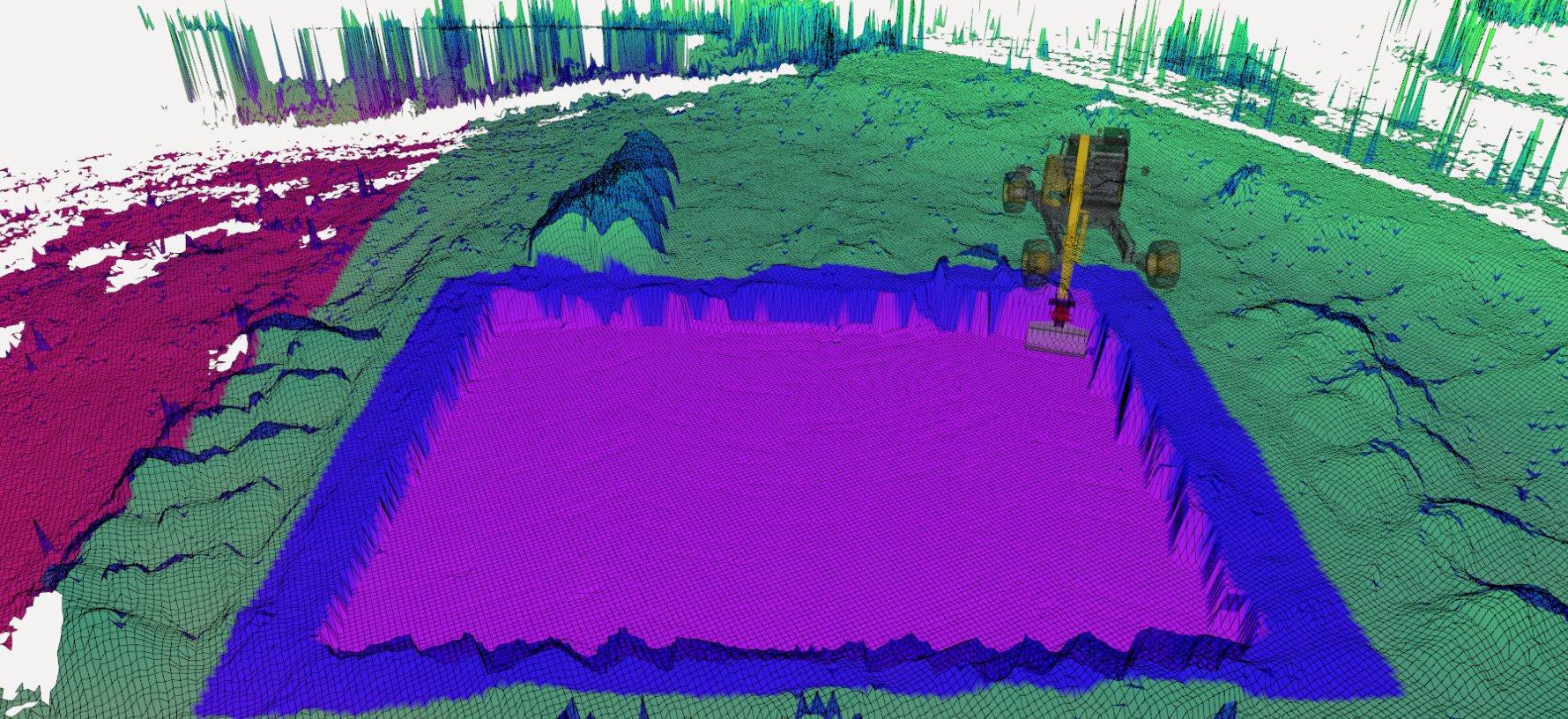}
        \label{fig:side_dump_left_back}
      \end{minipage}
    }
    \caption{Excavation mask layer and zone dumping costs for different cases throughout the excavation of a pit in simulation. \textbf{(a)}: excavation mask layer. \textbf{(b)}: the soil is moved from the front dig area to the front-left dump area. The back-left area has a higher cost because it is further away from the point where the excavator has dug. \textbf{(c)}: the first lane is completed, and soil is moved from the front to the back-right zone. The left zones are all inactive as they overlap with already dug areas. The front-right zone has a higher dumping cost since it is further away from the reachable dump zone in the back. \textbf{(d)}: dirt is moved from the front and front-right zone to the back-right zone. \textbf{(e)}: the permanent dump zone is unreachable, and the dirt is moved from the front, and front left zone to the back-right zone. \textbf{(f)}: first two lanes are completed, and multiple permanent dump zones are reachable. The dirt is moved from the front to the front-right or back-left zone, depending on which zone is closer to the digging point.}
    \label{fig:single_cell}
\end{figure}

A permanent dump area is accessible if within the excavator's reach or further along the path and does not obstruct future targets. The local excavation geometry, shown in Figure \autoref{fig:local_geometry}, and the global planner's strategy contribute to this. Several dirt handling scenarios in the single-cell excavation are demonstrated in \autoref{fig:single_cell}.

The dig and dump selection continues until all zones are excavated, and then the machine transitions to the refinement phase, involving soil grading.

During refinement, the excavator sweeps from left to right in a grading motion to smooth the terrain. The arm's back-and-forth movement levels any unevenness to achieve the desired slope and surface smoothness.

\subsection{Arm Trajectory Planner}
The arm trajectory planner governs all arm trajectories, such as digging, dumping, and grading motions. It remains agnostic to the specific trajectories and focuses on optimizing the trajectories' parameters or initial conditions. The digging system optimizes trajectories to scoop the maximum volume in the target zone. This work employs parameterized trajectories with inverse kinematics, but the planner is extendable to other low-level controllers.

\subsection{Parametrized Digging Trajectories}
\begin{figure}[!hbt]
  \subfloat[Digging on flat terrain]{
    \begin{minipage}{0.48\textwidth}
      \includegraphics[width=\linewidth]{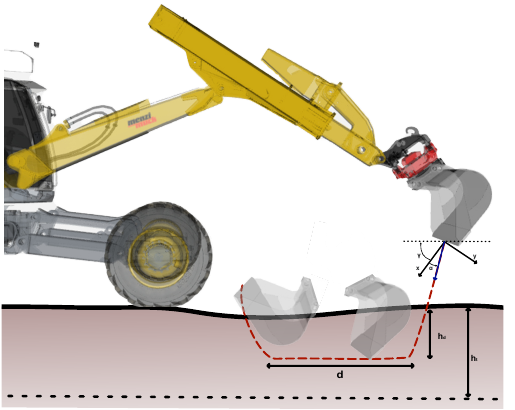}
      \label{fig:bucket_frame}
    \end{minipage}
  }
  \subfloat[Digging on receding terrain]{
    \begin{minipage}{0.48\textwidth}
      \includegraphics[width=\linewidth]{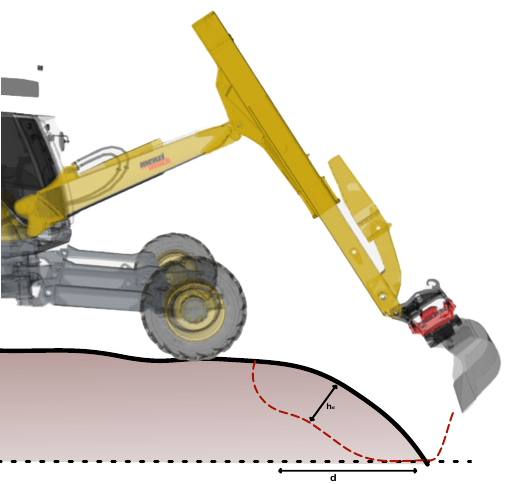}
      \label{fig:trench_dig}
    \end{minipage}
  }
  \caption{Two examples of digging trajectories on different terrain profiles. \textbf{(a)}: typical on undug flat soil. \textbf{(b)}: common while digging a pit or trench.}
\end{figure}

The digging trajectory is divided into penetration, dragging, and closing. Penetration starts at ground level at an initial position, $x_0$, with the shovel edge moving along the shovel's x-axis, as shown in Figure \ref{fig:bucket_frame} until the target depth is reached. This occurs when the target elevation or maximum depth to the surface is attained. The attitude angle, $\gamma$, is independent of the soil profile and varies linearly across the workspace, with values summarized in \autoref{tab:trajectory_params}.

In the dragging phase, the bucket moves radially along the boom, with its angle varying linearly between $\gamma_{\text{min}}$ and $\gamma_{\text{max}}$. The dragging motion stops when the bucket is full, a self-collision risk is detected, or it exits the diggable zone. To increase the average volume per scoop, the scooped volume is calculated by integrating the shovel depth, including areas outside the target workspace but still part of the next workspace. As excavation progresses, scoops diminish in size, ensuring maximum soil removal.

During the closing phase, the bucket begins to close, moving radially and vertically until fully closed and at a target height above the soil.

\paragraph*{Loose Soil Trajectories} Scooping loose soil is challenging as it lacks support from undisturbed soil and can easily displace and accumulate near the machine. The same parametrization is used to scoop loose soil, with different initial attitude angles and radial distance during the closing phase (as shown in Table \ref{tab:trajectory_params}). The initial attitude angle is reduced, and the bucket only moves vertically up during closing, thus reducing the radial forces on the soil, allowing scooping without dragging and spillage.



\begin{table}[!hbt]
  \centering
  \begin{tabular}{lll}
    \hline
    \textbf{Parameter} & \textbf{Value} & \textbf{Description} \\
    \hline
    $\gamma_{\text{min}}$ & 0.5 rad & Minimum attitude angle \\
    $\gamma_{\text{max}}$ & 1.5 rad & Maximum attitude angle \\
    $\gamma_{\text{max-dirt}}$ & 0.9 rad & Maximum attitude angle for loose soil \\
    $d_{\text{max}}$ & 1.5 m & Maximum dragging distance \\
    $h_{\text{max}}$ & 0.3 m & Maximum depth \\
    $V_{b}$ & 0.6 m$^3$ & Bucket volume \\
    $V_{\text{max}}$ & 0.8 m$^3$ & Maximum volume \\
    $h_{c}$ & 0.5 m & Height change in the closing motion \\
    $v_{d}$ & 0.5 m/s & Digging velocity \\
    \hline
  \end{tabular}
  \caption{Trajectory parameters}
  \label{tab:trajectory_params}
\end{table}

\subsubsection{Digging Planner}
The digging planner selects the parameters for the digging trajectory to maximize the expected scooped soil volume in the target digging zone. We use Bayesian optimization as the trajectory parameter space is low-dimensional, we cannot estimate gradients of the objective function, and the objective function is costly to evaluate since the whole trajectory must be simulated. This is particularly important if using a different subsystem generates digging trajectories requiring more computation time.

In our case, the optimization problem (shown in Equation \ref{eq:opt_problem}) is reduced to finding the 2D coordinates of the initial position for the digging trajectory.

\begin{equation}\label{eq:opt_problem}
  \begin{split}
    \max_{\mathbf{(r, \theta)}} \quad & V_w(\tau(r, \theta)) \\
    \text{subject to} \quad & r_{in} < r < r_{out} \\
    & \quad \theta_{min} < \theta < \theta_{max} 
  \end{split}
\end{equation}

The optimization objective is to maximize the scooped soil volume $V_w(\tau(r, \theta))$ in the workspace over a circular sector defined by radii $r_{in}$, $r_{out}$ and angular limits $\theta_{min}$, $\theta_{max}$.

\begin{figure}[!hbt]  
  \captionsetup[subfloat]{farskip=0pt,captionskip=0pt}
  \centering
  \subfloat[Workspace]{
    \begin{minipage}{0.33\textwidth}
\adjustbox{trim=0.7cm 0.cm 0.0cm 0cm,clip,valign=t}{\includegraphics[height=4.2cm, width=\linewidth]{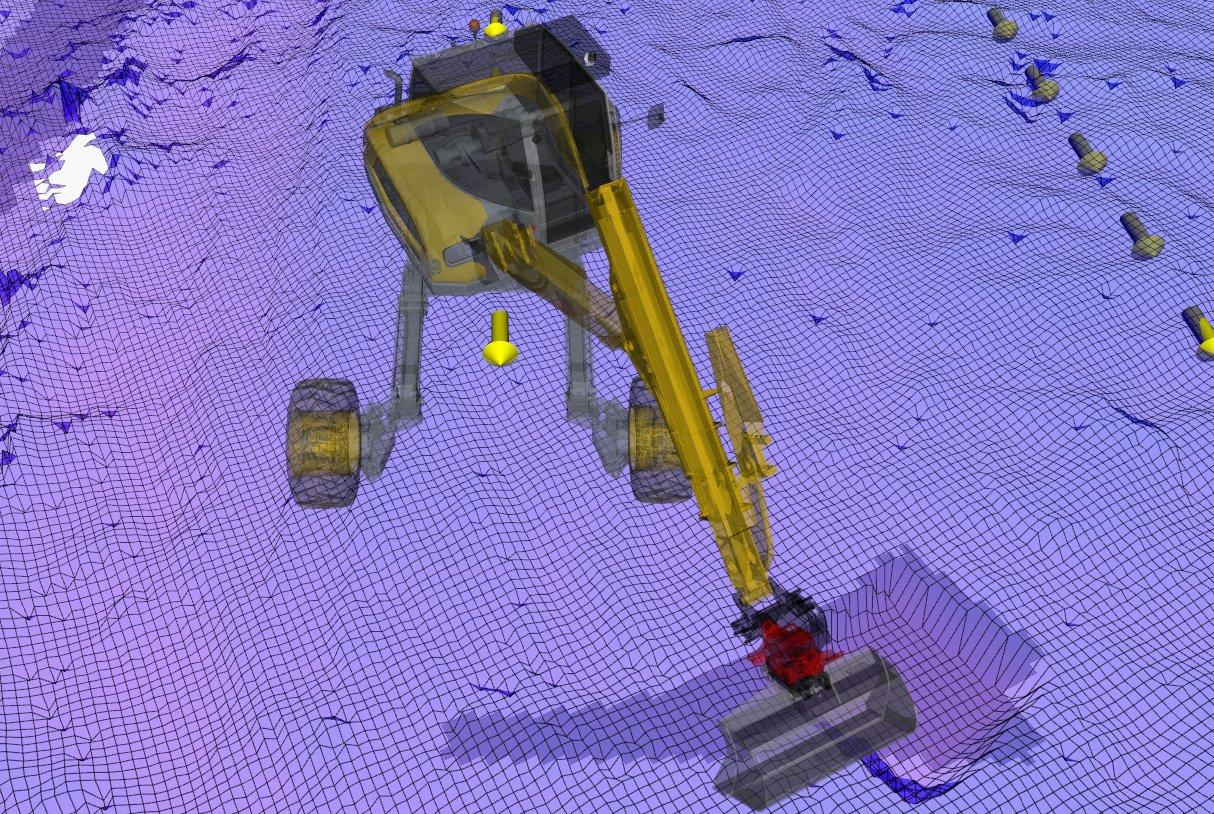}}
      \vspace{4pt}
    \end{minipage}
  }
  \subfloat[Estimated scoop volume]{
    \begin{minipage}{0.33\textwidth}
      \includegraphics[height=4.5cm,width=\linewidth]{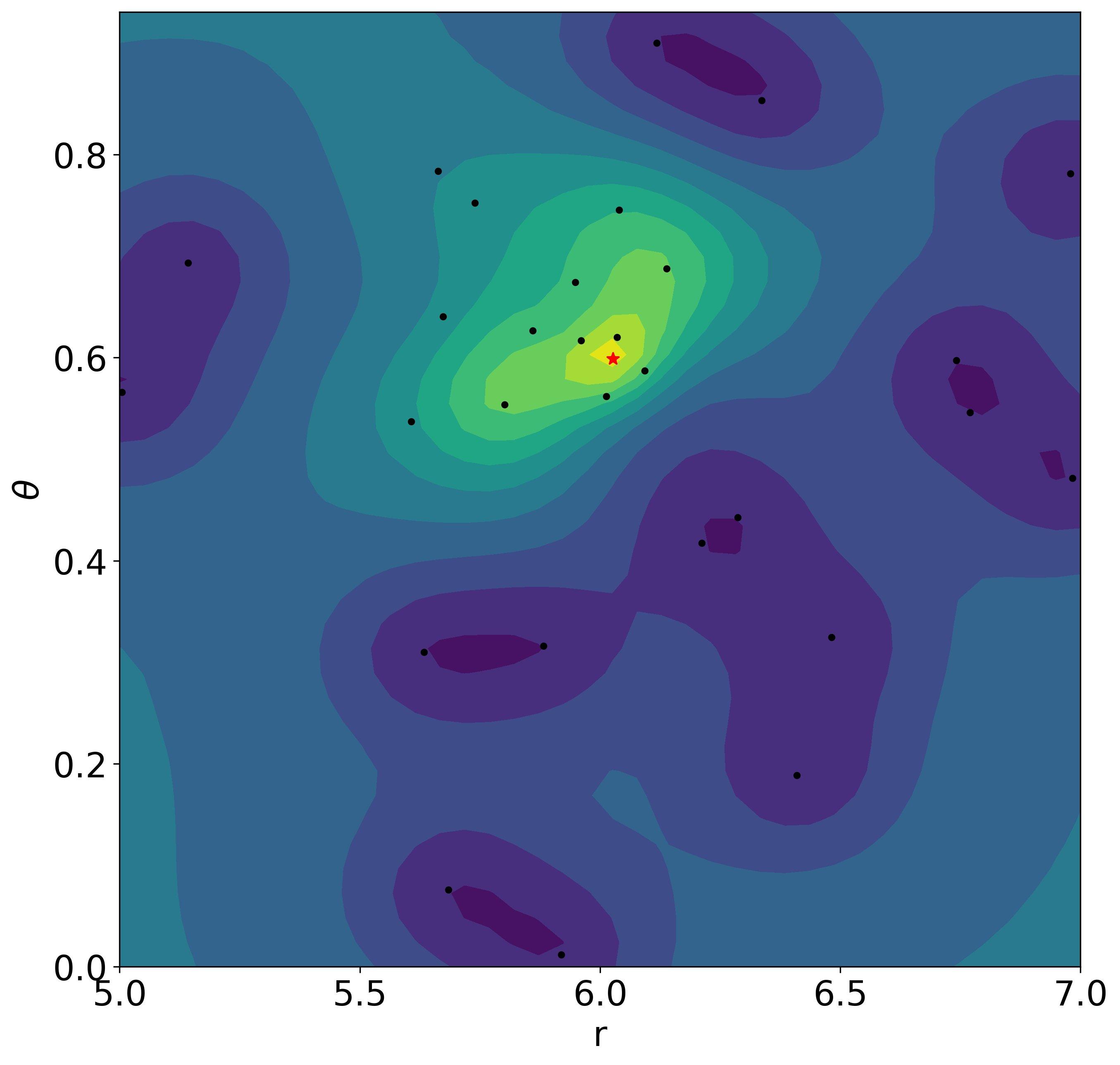}
    \end{minipage}
  }
  \subfloat[Expected scoop volume]{
    \begin{minipage}{0.33\textwidth}
      \includegraphics[height=4.5cm,width=\linewidth]{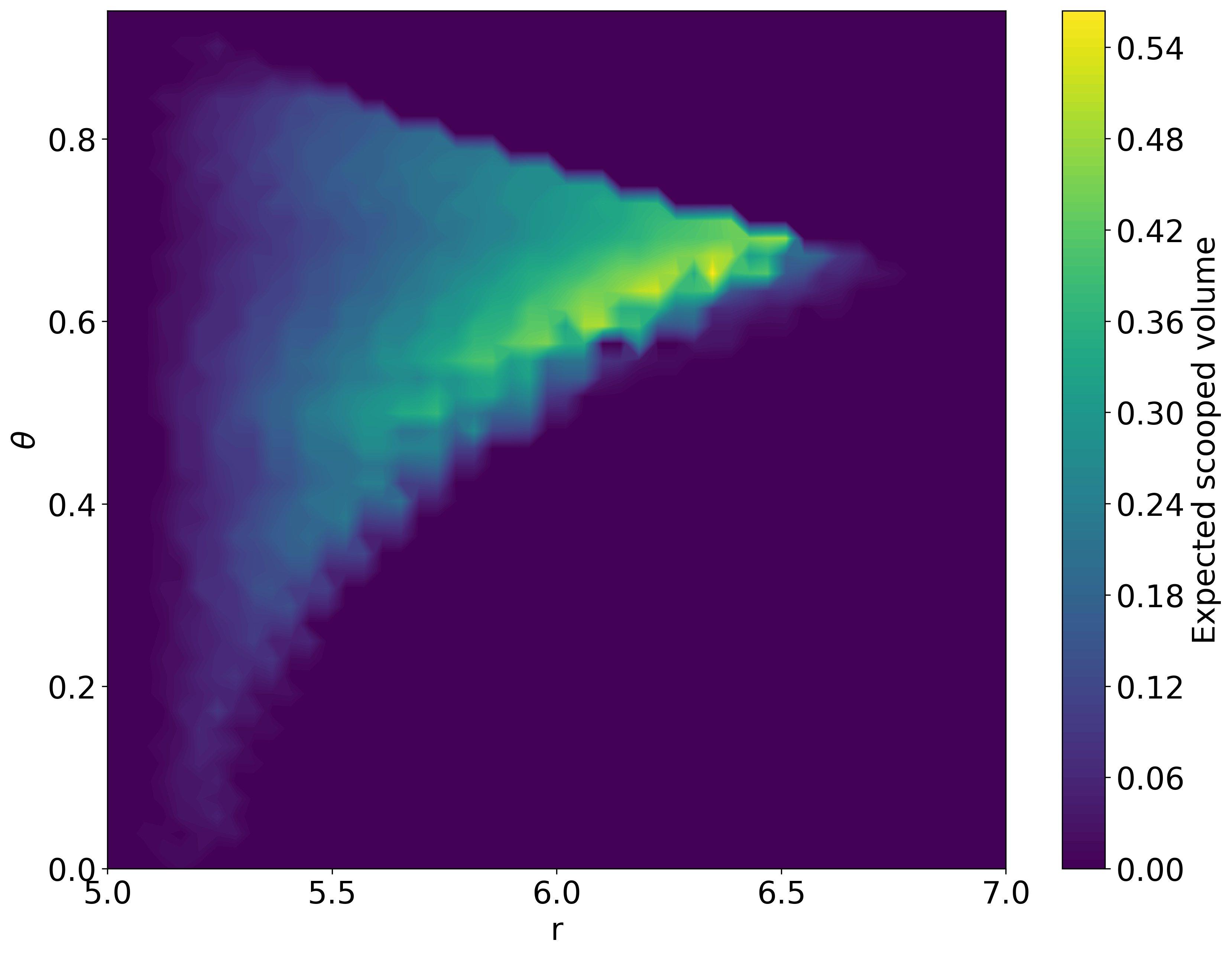}
    \end{minipage}
  }
  \caption{\textbf{(a)}: excavator digging at the first workspace of the pit, indicated in a darker color. The workspace is both constrained radially and tangentially by the edges of the pit. \textbf{(b)}:  estimated optimization landscape of the expected scooped volume function via a Gaussian process. The dots in the left plot are the points sampled by the optimizer with our custom sampler. \textbf{(c)}:  optimization landscape of the expected scooped volume function created with full search over the elevation map.}
  \label{fig:bayes}
\end{figure}

We solve this optimization problem using Bayesian optimization with Gaussian processes, expected improvement as the acquisition function, and a custom sampler. The initial points are sampled uniformly in Cartesian space, some Gaussian noise is added, and then the points are transformed to polar coordinates. Since we sample multiple times the same space, the added noise allows the system to explore the dig area better. The Gaussian noise has a standard deviation of one-fourth of the spacing between points. We query the optimizer with 20 randomly selected values and allow ten iterations of refinement for a total of 30 calls to the objective function. The optimizer has been tuned to yield results with an average error of 0.06 m$^3$, approximately 10\% of the scooped volume, from the actual maximum. The grid search optimizer, by comparison, needs 150 samples to achieve the same precision, scaling exponentially if the number of tunable parameters increases. The real and estimated optimization landscapes are compared in \autoref{fig:bayes}.

\subsubsection{Refinement Trajectories}
Inspired by expert operators aiming to achieve a final grade of the soil surface, we design refinement trajectories to remove minor imperfections and spillage left from digging. The front digging zone is expanded by 10\% in the radial and tangential directions, and any accumulated soil closer to the machine base than the inner radius of the current digging zone is removed in the next workspace. We use only radially inward motions to minimize the risk of pushing soil out of reach.

\begin{figure}[!hbt]
  \centering
  \includegraphics[width=0.45\textwidth]{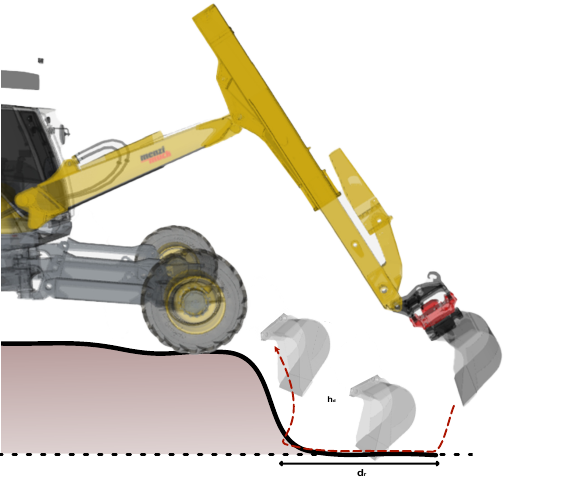}
  \caption{Refinement Trajectory}
  \label{fig:refinement_trajectory}
\end{figure}

During arm movement, the shovel edge height and attitude angle are maintained as specified, and the cabin is rotated by an angle equal to the angular dimension of the shovel edge, as shown in \autoref{fig:refinement_trajectory}.

\subsubsection{In-Air Motion Trajectories}
The arm trajectories for reaching the digging location and dumping soil are created using Hermite splines with a step velocity profile. This approach enables circular arm movement around the base, with the plan computed in cylindrical coordinates by interpolating between the bucket's initial and final positions. The orientation is fixed for trajectories leading the bucket to the digging trajectory's starting point. In contrast, the bucket is kept closed during the dumping trajectory until the final six seconds, when it begins to open. The ground is kept at a minimum distance of 0.3 meters to prevent collisions, as determined by querying the elevation map at various points along the shovel edge. The hierarchical inverse kinematic controller presented in \cite{judPlanningControlAutonomous2017a} tracks the plans.

The dump point is selected by convolving the shovel filter (a 2D projection of the bucket with unit weight) with the dumping zone, followed by a full grid search over the allowed locations to choose the cell with the minimum dumping cost, as defined in \autoref{eq:dumping_cost_convolved}.

\begin{equation}
  \begin{split}
    C_{dump}(x, y) = \alpha_d \sum_{i=1}^{n} \sum_{j=1}^{n} S_{ij} h_{dirt}(x + \Delta s, y + \Delta s) - \beta_b {}_B x_{bd} - \gamma_b {}_B y_{bd}
  \end{split}
  \label{eq:dumping_cost_convolved}
\end{equation}

Here, $h_{dirt}(x, y)$ is the height of the dirt at the dumping location, $S$ is the shovel filter, and $\Delta s$ is the grid map's resolution. The coordinates of the dump point in the base frame are given by ${}_B x_{bd}$ and ${}_B y_{bd}$. The dumping cost, a weighted sum of the dirt height and distance from the base is governed by weights $\alpha_d = 1.0$, $\beta_d = 0.1$, and $\gamma_d = 0.05$. This cost structure promotes uniform dirt spreading and prevents accumulation near the base.

\section{Navigation Planning}
The motion planning module produces safe and efficient paths for an excavator during excavation tasks, utilizing information from an occupancy map and motion constraints.

\subsection{Occupancy Map}
The motion planning module employs an occupancy map, a 2D representation of the terrain's traversability, to validate sampled states and compute path costs. The map is formed by merging information from:
\begin{itemize}
  \item An offline traversability map considering slopes and steps.
  \item User-defined non-accessible zones and manually identified obstacles invisible to LIDAR, like wire fences.
  \item Online-generated maps marking dug areas and soil piles as non-traversable.
  \item Dug areas marked as non-traversable to avoid soil deformation and surface quality degradation.
\end{itemize}
Figure \ref{fig:occupancy} depicts an operational example in our test area.

\subsection{Planner}
We utilize RRT* \cite{sucanOpenMotionPlanning2012} implemented in the OMPL with a state space defined by Reeds-Shepp curves in SE2 for local motion planning. This choice accommodates the changing nature of dug areas and soil piles.

The planner produces a sequence of poses, accounting for terrain traversability through an occupancy map comprising offline maps, inaccessible areas, and real-time updates. Excavator clearance is also considered, as loading soil too close to the edge may cause collapses, damaging the machine or affecting the grade. Falling risks are evaluated as well.

The validity of states is assessed through the occupancy map, and costs are calculated using the following:
\begin{equation}
\label{eq:cost}
C_n(s) = \alpha_n \cdot d(s1, s2) + \frac{1}{M} \sum_{j = 0}^M \beta_n \cdot \exp^{\left(-\gamma_n \cdot sdf_{dug}(f_{s2j})\right)}
\end{equation}
where $d(s1, s2)$ is the distance between the current and previous base position and $sdf_{dug}(f_s2j)$ is the SDF value for the footprint $f_{s2j}$ and is the number of elevation map cells inside the footprint of the robot. The parameters $\alpha_n$, $\beta_n$, and $\gamma_n$ are set to 10.0, 5.0, and 0.7, respectively.

\begin{figure}[!hbt]
  \subfloat[Occupancy map]{
    \begin{minipage}{0.48\textwidth}
      \includegraphics[height=5cm, width=\linewidth]{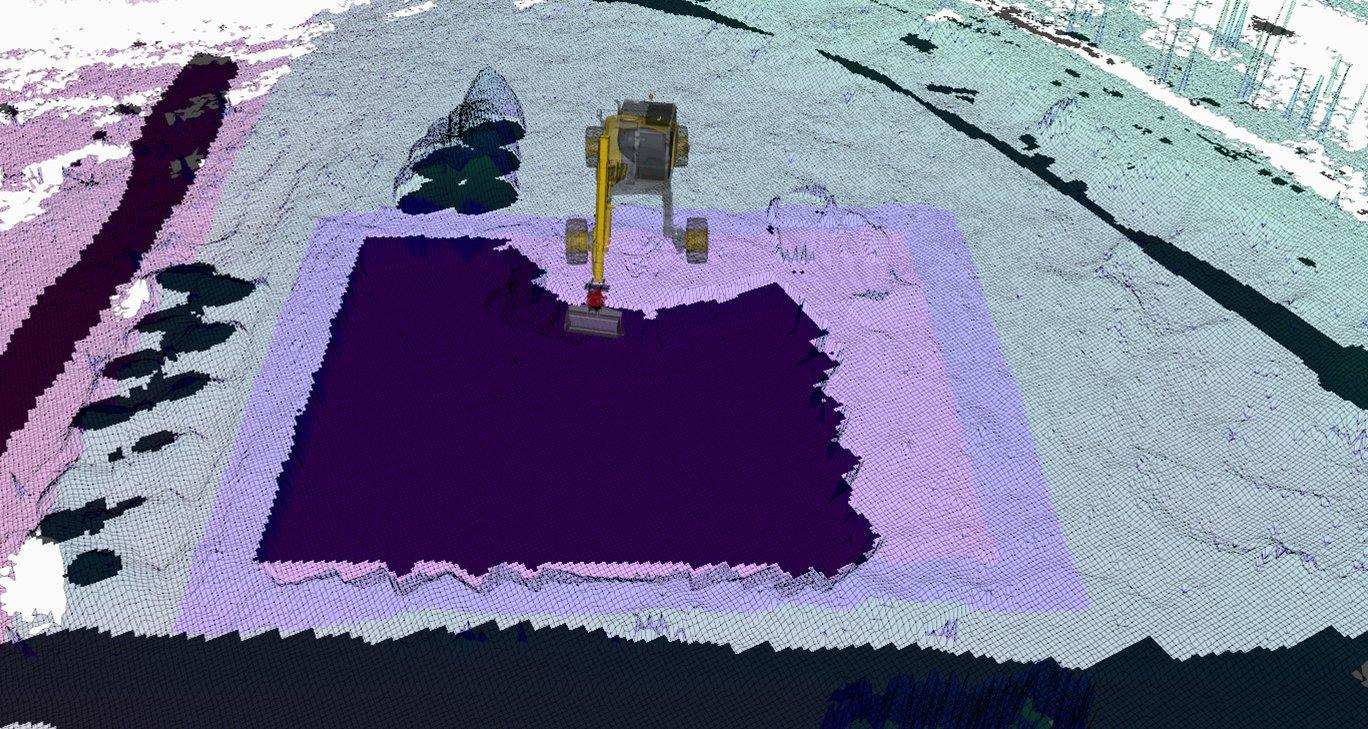}
      \label{fig:occupancy}
    \end{minipage}
  }
  \subfloat[Generated Plan]{
    \begin{minipage}{0.48\textwidth}
      \includegraphics[height=5cm, width=\linewidth]{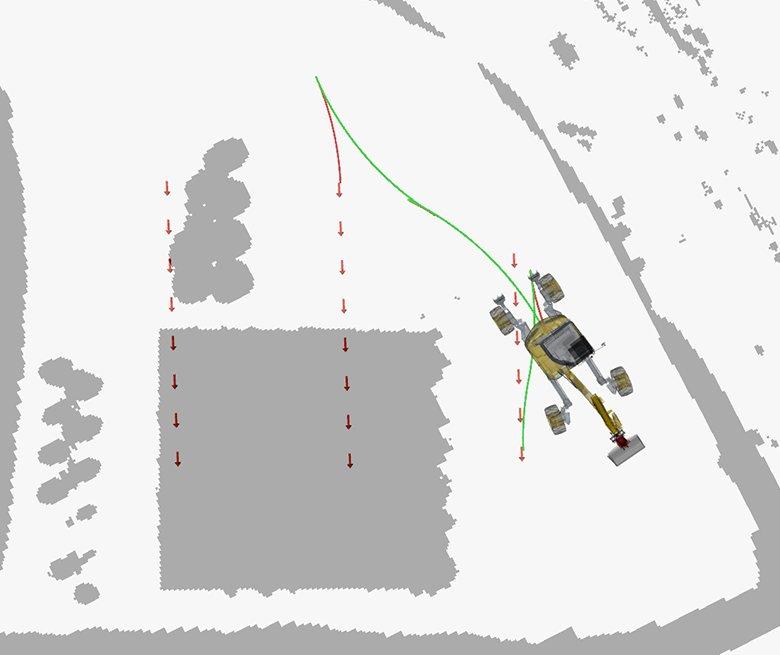}
      \label{fig:plan}
    \end{minipage}
  }
  \caption{The RRT* uses an occupancy map to check the feasibility of the path. Both images are generated in simulation. \textbf{(a)}: the occupancy map is generated by integrating the latest elevation map information. Note how the pit and dumping piles are marked as untraversable. \textbf{(b)}: the generated plan is shown in green (forward moving) and red (backward moving). The excavator is able to navigate around the pit and reach the last lane.}
\end{figure}

The full pose is obtained by constraining height relative to soil and maintaining gravity alignment \cite{hutterForceControlActive2017}. Plans are tracked using a pure pursuit controller \cite{jelavicRoboticPrecisionHarvesting2022}, with safety measures for deviation from target paths due to slippery or muddy terrain. Figure \ref{fig:plan} provides an example of a generated plan.

The planner's variable time ranges from 0.5 to 5 seconds for 0.5 to 60-meter plans, utilizing the RRT* algorithm with up to 10 trials. Success was achieved 78.7\% of the time in a simulated scenario, and the failure probability is less than $3 e^{-7}$.

\subsection{State Machine}
The control state machine for the excavator robot is organized as follows:

\begin{description}
  \item[\textit{Initialize Workspace}] Initializes the workspace by defining the local digging geometry.

  \item[\textit{Main Digging Loop}] While the workspace is not complete, the robot repeatedly goes through the following states:
  \begin{itemize}
    \item \textit{Check Workspace}: Determines if the workspace is complete and finds the next dig and dump point using the local planner.
    \item \textit{Find Dig Point}: Uses the digging planner to locate the starting point of the scoop and moves the arm to that position.
    \item \textit{Dig}: Executes the digging motion.
    \item \textit{Dump}: Identifies the dump point, moves to it, and discards the soil.
  \end{itemize}

  \item[\textit{Find Path Plan}] Locates a path plan to the subsequent base pose using the RRT* planner.

  \item[\textit{Driving}] Propels the excavator to the designated base pose utilizing the pure pursuit controller.
\end{description}

Note: The control state machine also handles various corner cases separately to ensure robust operation.

\section{Experiments}

In this section, we provide an overview of the experimental evaluation conducted to assess the performance of our excavation planner. We employ two distinct sets of experiments: simulation-based tests focusing on the global planner's performance in a realistic excavation scenario, and real-world deployment to evaluate the full system's capabilities during the excavation of a prototypical building foundation.

This section discusses the tests carried out to evaluate the performance of our excavation planner. We conducted two types of tests: 
1. Simulations that assess the global planner's efficiency in virtual digging environments.
2. Real-world tests that gauge the system's overall ability during the excavation of a typical building foundation.

\subsection{Global Planner Experiments}

This section discusses the tests carried out to evaluate the performance of our excavation planner. We conducted two types of tests: simulations that assess the global planner's efficiency in virtual digging environments and real-world tests that gauge the system's overall ability during the excavation of a typical building foundation.


These building maps vary in size, from 20 m to 100 m on each side. The random crops are even bigger, from 100 m to 1000 m per side. From this data, we generated five types of excavation tasks:
\begin{itemize}
    \item Foundations: digging the shape of a building with no obstacles.
    \item Exterior Foundations: digging the inverse shape of a building's foundation, treating the buildings as obstacles.
    \item Exterior Foundations Traversable: same as above, but here, the building shapes can be crossed or passed through.
    \item Crops: digging that involves multiple building shapes on one map.
    \item Exterior Crops: digging the insides of streets and parks, with buildings acting as obstacles.
\end{itemize}

For the first three tasks, we assumed dirt could be placed anywhere outside the dig area. For the last two tasks, due to their complex nature, we didn't consider how to manage the excavated dirt. Instead, we presumed dirt could be moved off-site from any location using other methods or machines.


We assessed the planner's performance based on four criteria: successful plan rate, path efficiency, workspaces efficiency, and the digging coverage.A plan is "unsuccessful" if the planner can't process the digging area's shape or can't find a digging solution at all.





The \emph{path efficiency} indicates how straight and short the system's movement paths are. It's found by adding up the straight-line distances between all digging positions, then dividing by the digging area's size:
\begin{equation}
S_p = \sum_{i = 0}^{N - 1} \frac{{(x_{B_{i + 1}} - x_{B_i}})}{\sqrt{A_d}}
\end{equation}
The \emph{workspace efficiency} measures how many local working areas are necessary for the planning problem. It's determined by:
\begin{equation}
S_w = \frac{N_w \cdot A_w}{A_d}
\end{equation}
Here, \(A_w = \frac{1}{2} \pi R_{\text{max}}^2\) represents the reference workspace area.
The \emph{digging coverage} reveals what percentage of the digging area the plan covers.

\begin{figure}[!hbt]
  \begin{minipage}{0.32\textwidth}
    \centering
    \includegraphics[width=\textwidth]{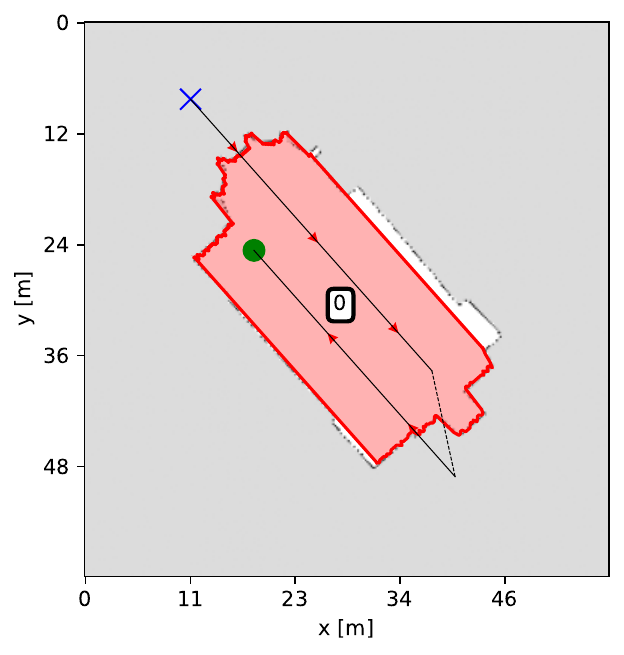}
  \end{minipage}
  \hfill
  \begin{minipage}{0.32\textwidth}
    \centering
    \includegraphics[width=\textwidth]{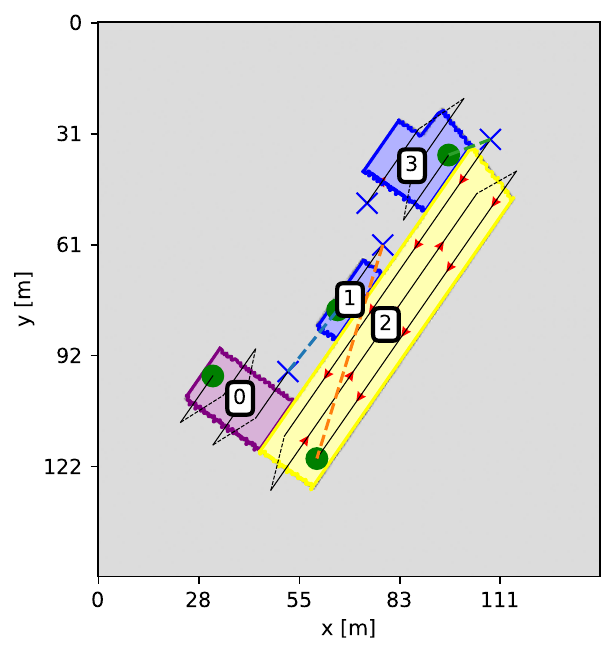}
  \end{minipage}
  \hfill
  \begin{minipage}{0.32\textwidth}
    \centering
    \includegraphics[width=\textwidth]{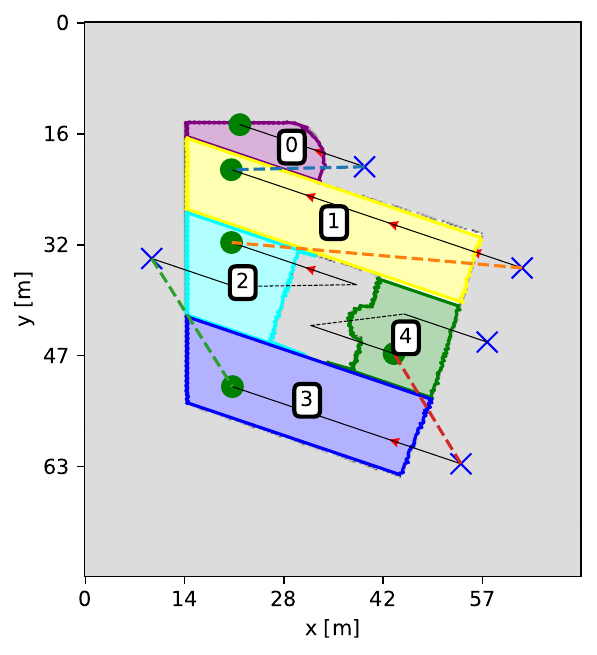}
  \end{minipage}
  \caption{Three solved samples from the "Foundations" dataset.}
  \label{fig:foundations}
\end{figure}

\begin{figure}[!hbt]
  \begin{minipage}{0.32\textwidth}
    \centering
    \includegraphics[width=0.8\textwidth]{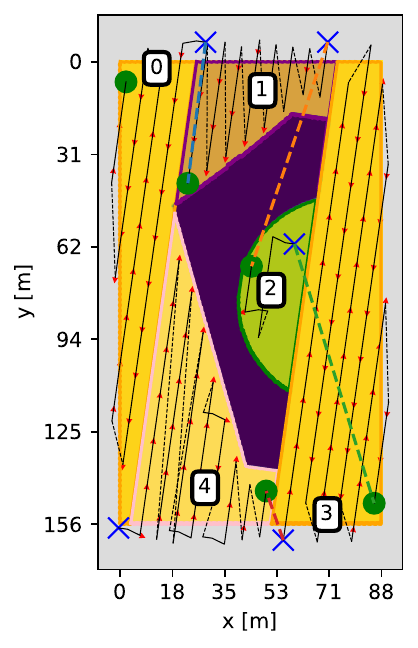}
  \end{minipage}
  \hfill
  \begin{minipage}{0.32\textwidth}
    \centering
    \includegraphics[width=\textwidth]{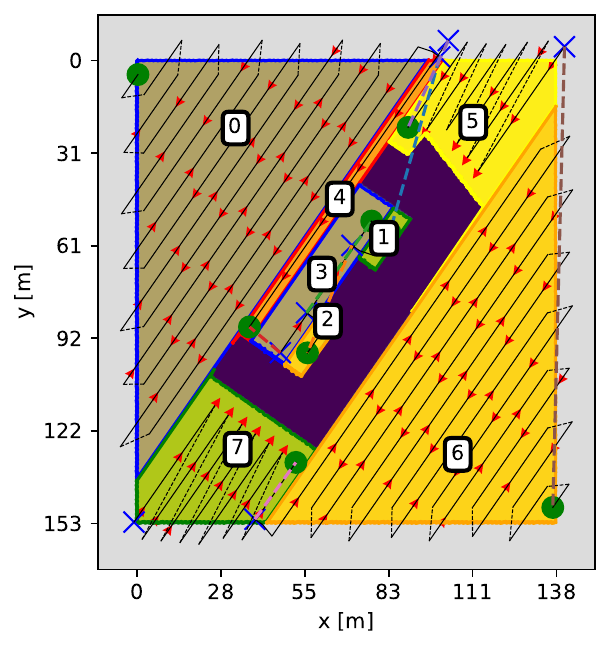}
  \end{minipage}
  \hfill
  \begin{minipage}{0.32\textwidth}
    \centering
    \includegraphics[width=\textwidth]{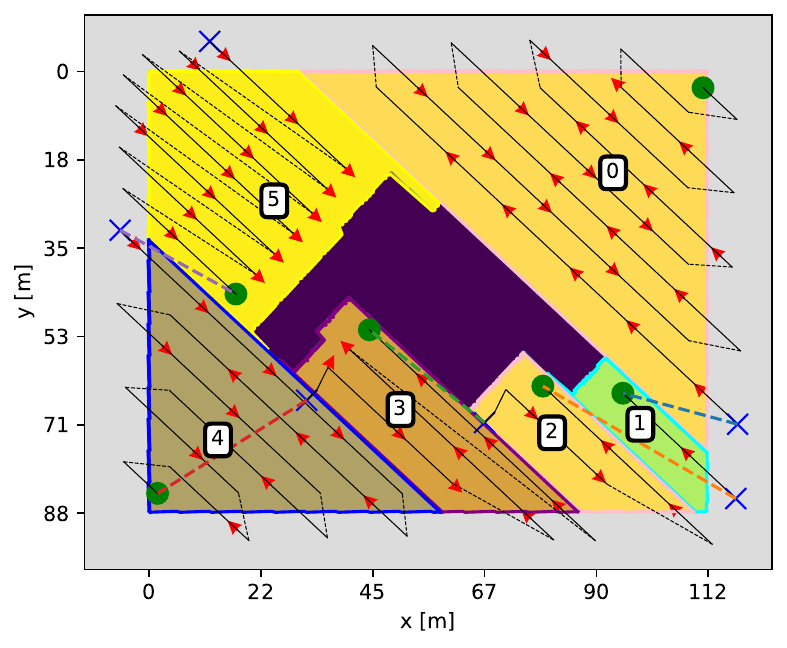}
  \end{minipage}
  \caption{Three solved samples from the "Exterior Foundations" dataset.}
  \label{fig:exterior_foundations}
\end{figure}

\begin{figure}[!hbt]
  \begin{minipage}{0.32\textwidth}
    \centering
    \includegraphics[width=\textwidth]{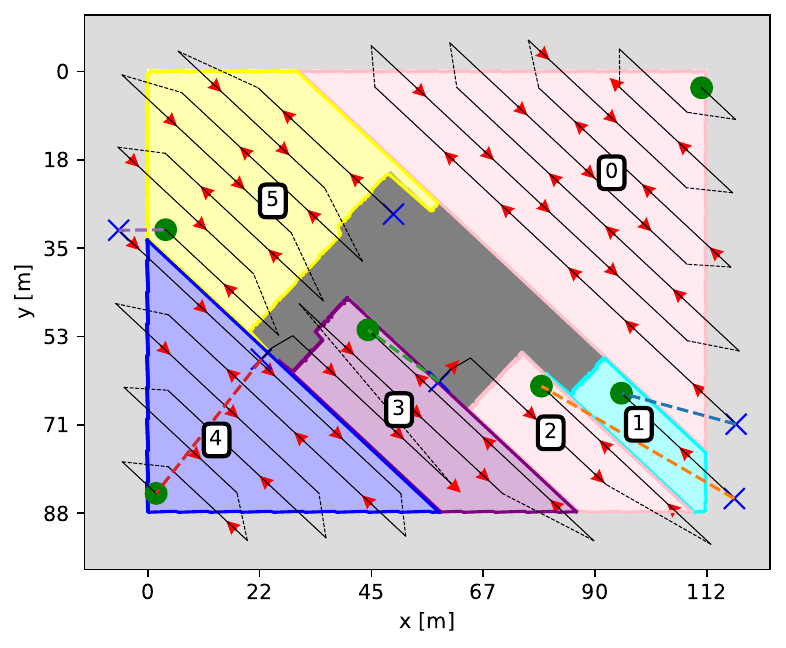}
  \end{minipage}
  \begin{minipage}{0.32\textwidth}
    \centering
    \includegraphics[width=\textwidth]{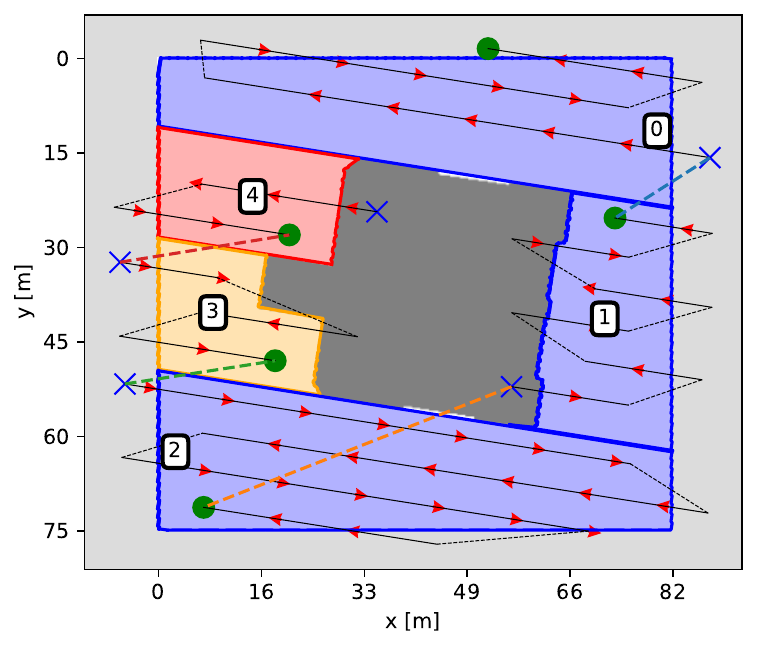}
  \end{minipage}
  \begin{minipage}{0.32\textwidth}
    \centering
    \includegraphics[width=\textwidth]{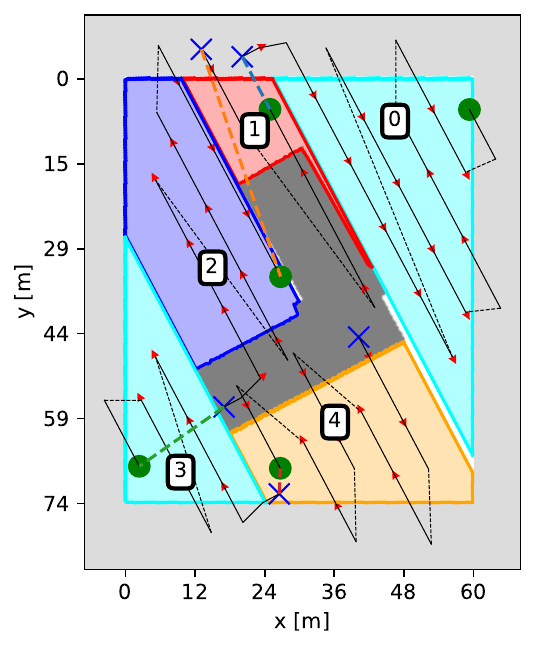}
  \end{minipage}
  \caption{Three solved samples from the "Exterior Foundations Traversable" dataset.}
  \label{fig:exterior_foundations_traversable}
\end{figure}

\begin{figure}[!hbt]
  \begin{minipage}{0.32\textwidth}
    \centering
    \includegraphics[width=\textwidth]{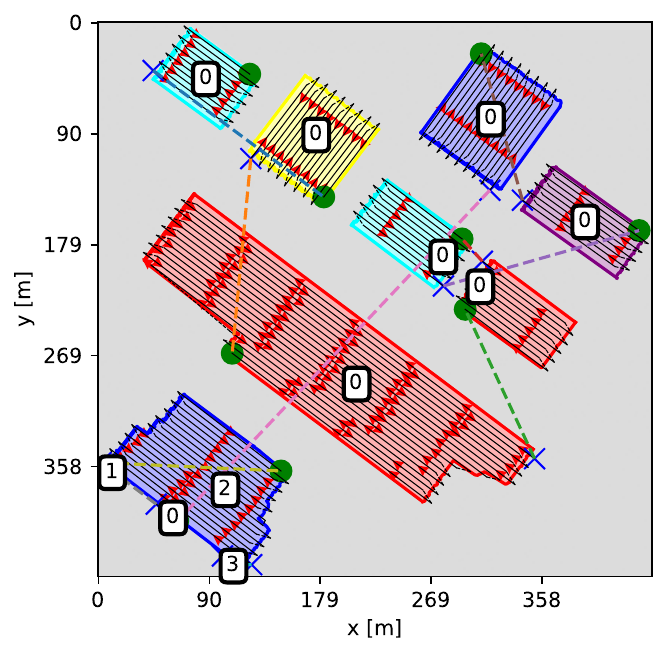}
  \end{minipage}
  \begin{minipage}{0.32\textwidth}
    \centering
    \includegraphics[width=\textwidth]{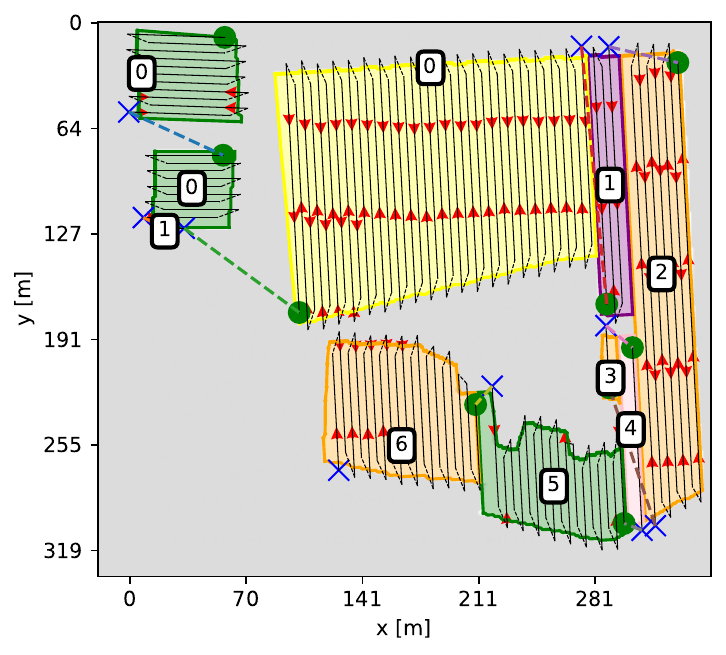}
  \end{minipage}
  \begin{minipage}{0.32\textwidth}
    \centering
    \includegraphics[width=\textwidth]{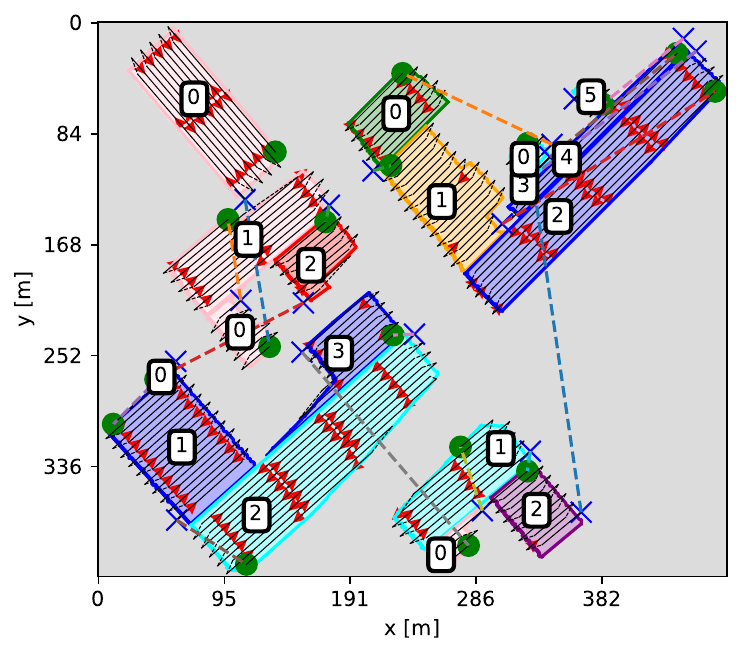}
  \end{minipage}
  \caption{Three solved samples from the "Crops" dataset.}
  \label{fig:crops}
\end{figure}

\begin{figure}[!hbt]
  \begin{minipage}{0.32\textwidth}
    \centering
    \includegraphics[width=\textwidth]{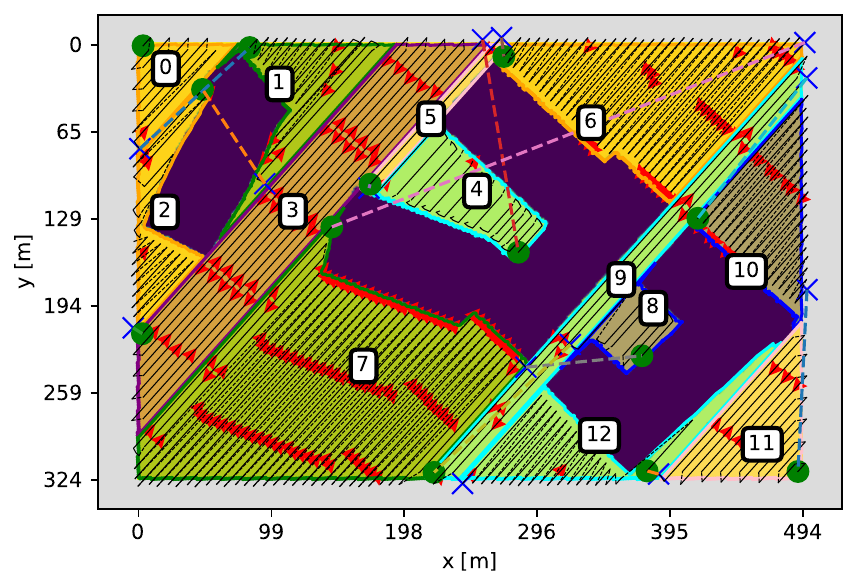}
  \end{minipage}
  \begin{minipage}{0.32\textwidth}
    \centering
    \includegraphics[width=\textwidth]{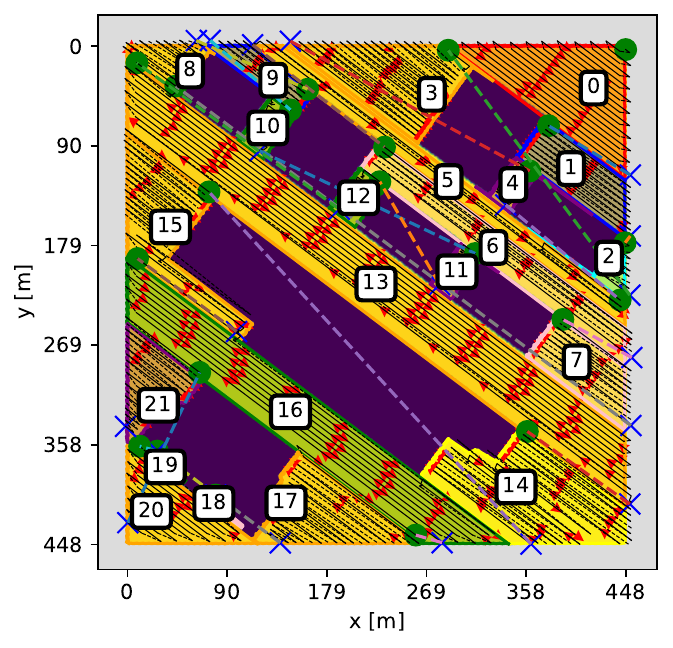}
  \end{minipage}
  \begin{minipage}{0.32\textwidth}
    \centering
    \includegraphics[width=\textwidth]{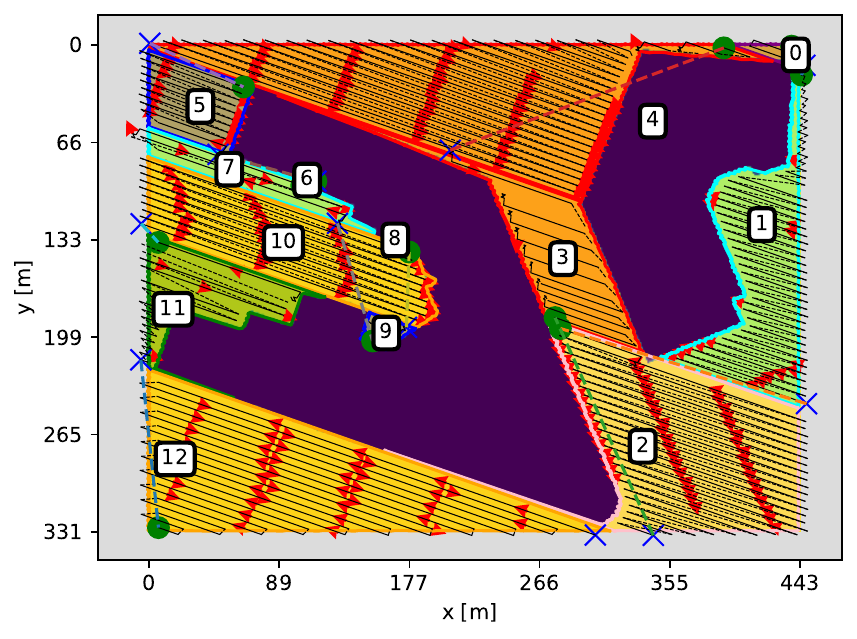}
  \end{minipage}
  \caption{Three solved samples from the "Exterior Crops" dataset.}
  \label{fig:exterior_crops}
\end{figure}

The solved excavation plans for the five different datasets can be seen in Figures \ref{fig:foundations}, \ref{fig:exterior_foundations}, \ref{fig:exterior_foundations_traversable}, \ref{fig:crops}, and \ref{fig:exterior_crops}. 

Tables \ref{tab:scores_wo_opt} and \ref{tab:scores_w_opt} show the benchmark scores for the global planner without and with coverage orientation optimization respectively. When not optimizing the coverage angle, it was aligned with the main axis of the excavation area, which already provided a stronger baseline compared to taking the original orientation of the excavation map into account.

\begin{table}[!htbp]
\centering
\caption{Benchmark scores for the global planner \textbf{without} coverage angle optimization.}
\label{tab:scores_wo_opt}
\begin{tabular}{lcccc}
\hline
\textbf{Dataset (size)} & \textbf{$S_p$} & \textbf{$S_w$} & \textbf{Coverage Fraction} \\ \hline
Foundations (838) & 3.23 (2.52) & 14.20 (4.09) & 0.944 (0.087) \\
Exterior Foundations (838) & 9.96 (4.60) & 15.03 (1.78) & 0.952 (0.115) \\
Exterior Foundations Traversable (838) & 9.78 (4.24) & 15.35 (1.77) & 0.960 (0.104) \\
Crops (100) & 68.12 (22.57) & 27.95 (3.29) & 0.912 (0.163) \\
Exterior Crops (100) & 105.23 (45.16) & 26.66 (3.51) & 0.782 (0.293) \\ \hline
\end{tabular}
\end{table}

\begin{table}[!htbp]
\centering
\caption{Benchmark scores for the global planner \textbf{with} coverage angle optimization. The coverage direction is aligned with the major axis of the excavation area.}
\label{tab:scores_w_opt}
\begin{tabular}{lcccc}
\hline
\textbf{Dataset (size)} & \textbf{$S_p$} & \textbf{$S_w$} & \textbf{Coverage Fraction} \\ \hline
Foundations (838) & 5.08 (2.51) & 19.89 (4.38) & 0.982 (0.052) \\
Exterior Foundations (838) & 9.87 (5.40) & 13.66 (2.11) & 0.972 (0.072) \\
Exterior Foundations Traversable (838) & 9.64 (4.42) & 13.91 (2.14) & 0.978 (0.041) \\
Crops (100) & 73.79 (19.75) & 29.09 (3.20) & 0.991 (0.028) \\
Exterior Crops (100) & 130.08 (41.67) & 26.45 (3.41) & 0.982 (0.039) \\ \hline
\end{tabular}
\end{table}



\subsection{Deployment Experiments}
The experiment assessed the performance of our autonomous excavator in digging a pit of dimensions 15.6 meters by 11.5 meters and a target depth of approximately 1 meter.

The same target height was set for the entire excavation area to maintain a uniform pit floor. This task was chosen as it is typical for excavator operators and requires multiple passes to relocate the dug soil without using a dump truck or wheel loader.

\begin{figure}[!htb]
  \centering
  \includegraphics[height=5.5cm, width=0.7\linewidth]{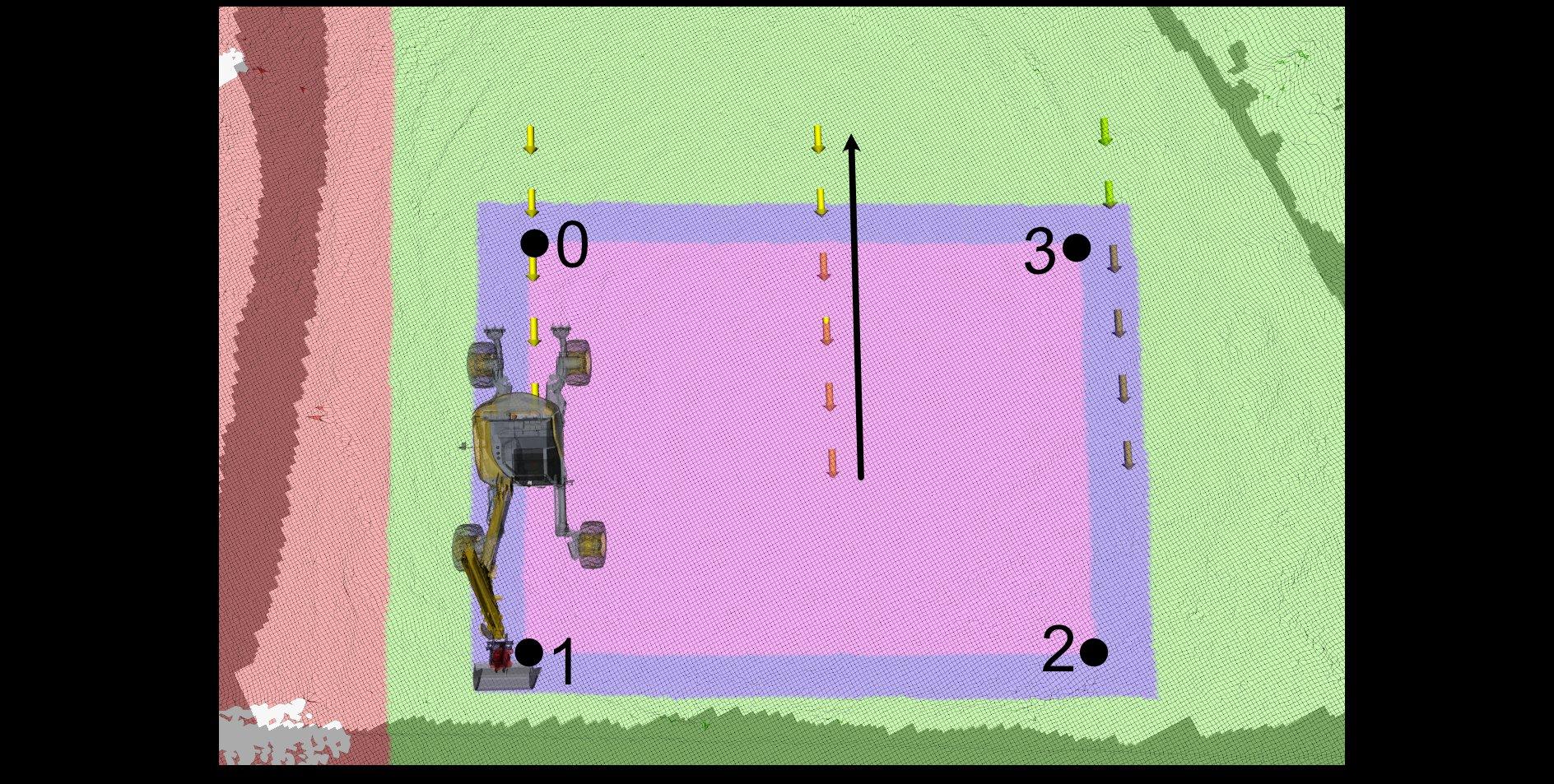}
  \caption{The solution to the excavation task. The yellow arrows correspond to the base poses, and the color scheme for the excavation mask is the same as used in Figure \ref{fig:single_cell}.}
  \label{fig:solution_pit}
\end{figure}

The solution is depicted in \autoref{fig:solution_pit}. Yellow arrows represent base poses, and the excavation mask color scheme is consistent with \autoref{fig:single_cell}. Only corners 1 and 2 of the four considered starting points yield feasible solutions due to spatial constraints and a manually marked fence, which prevent full pit coverage beginning from the other two corners.

\begin{figure}[!hbt]
  \captionsetup[subfloat]{farskip=0pt,captionskip=0pt}
  \centering
  \subfloat[]{
    \begin{minipage}{0.48\textwidth}
      \includegraphics[height=3.5cm, width=0.95\linewidth]{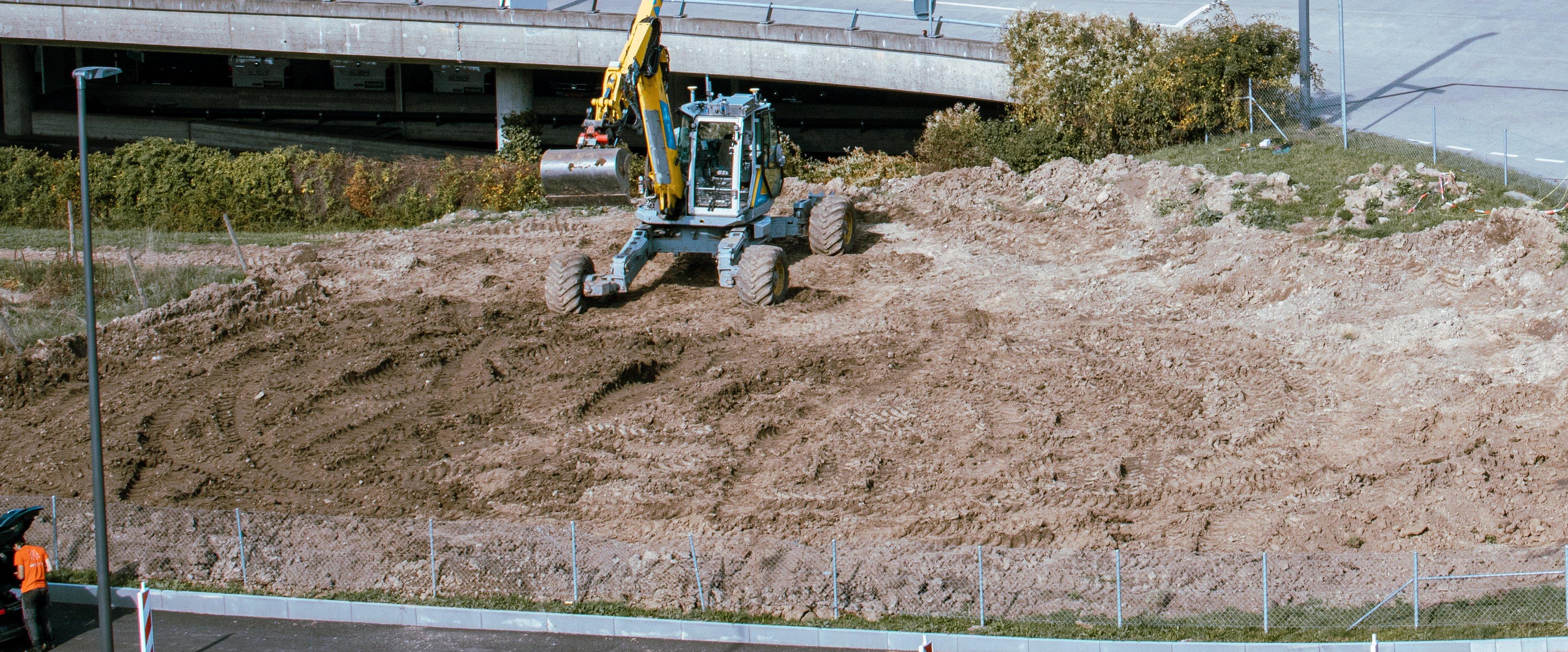}
      \label{fig:real_0}
    \end{minipage}
  }
  \subfloat[]{
    \begin{minipage}{0.48\textwidth}
      \includegraphics[height=3.5cm, width=0.95\linewidth]{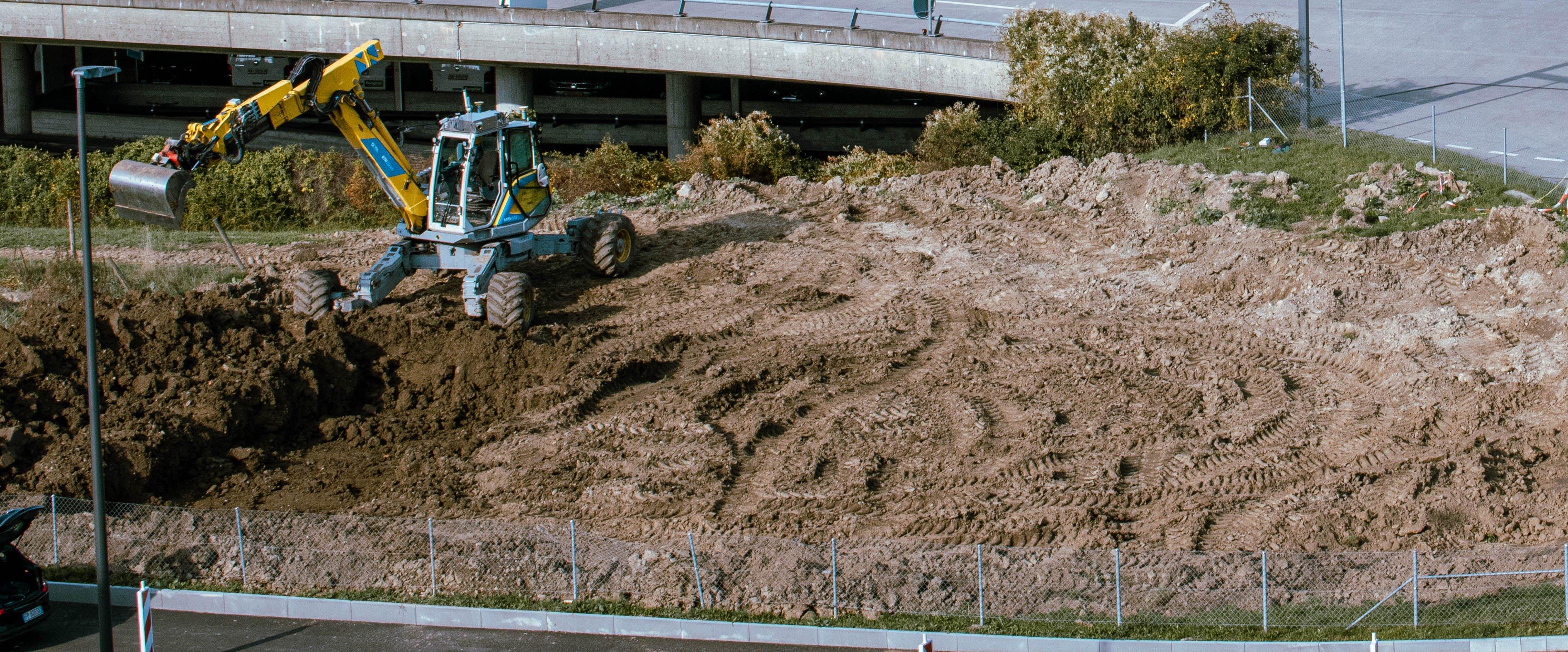}
      \label{fig:real_1}
    \end{minipage}
  }
  \hfill
  \subfloat[]{
    \begin{minipage}{0.48\textwidth}
      \includegraphics[height=3.5cm, width=0.95\linewidth]{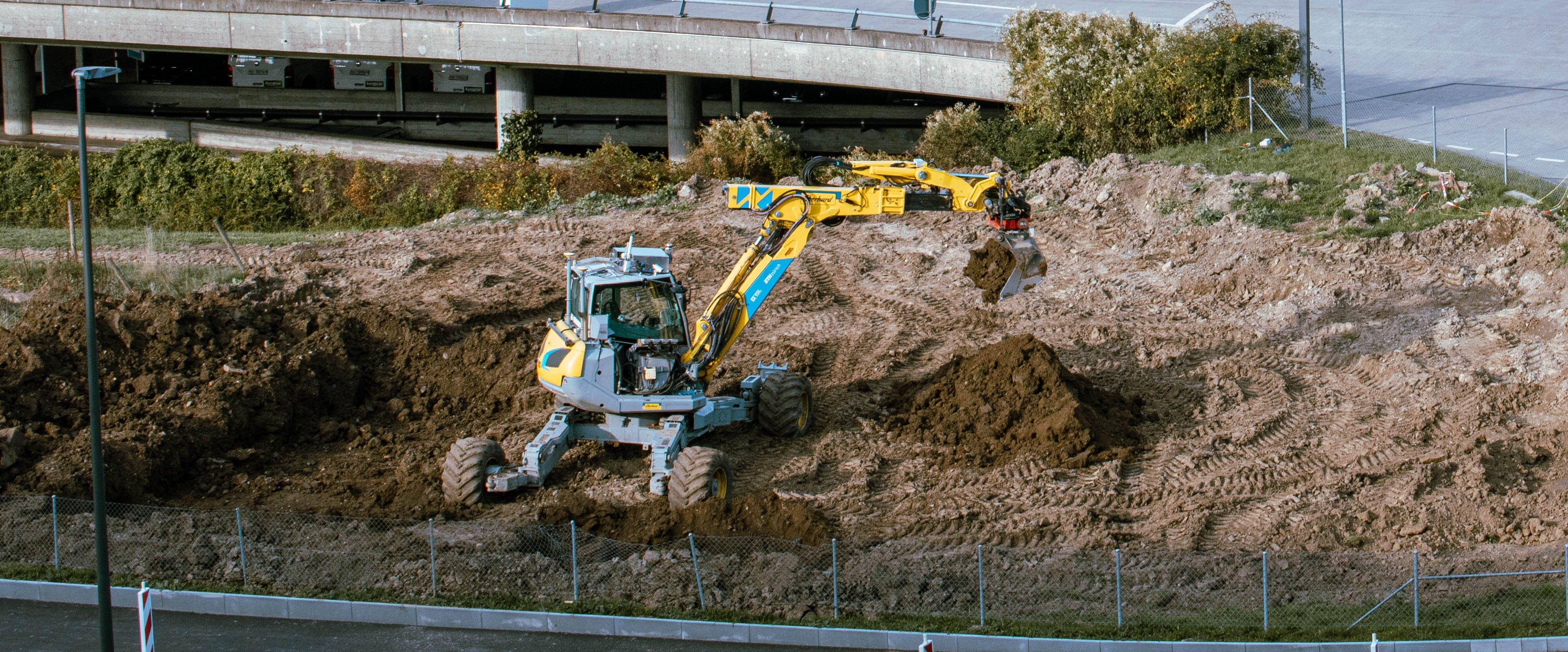}
      \label{fig:real_2}
    \end{minipage}
  }
  \subfloat[]{
    \begin{minipage}{0.48\textwidth}
      \includegraphics[height=3.5cm, width=0.95\linewidth]{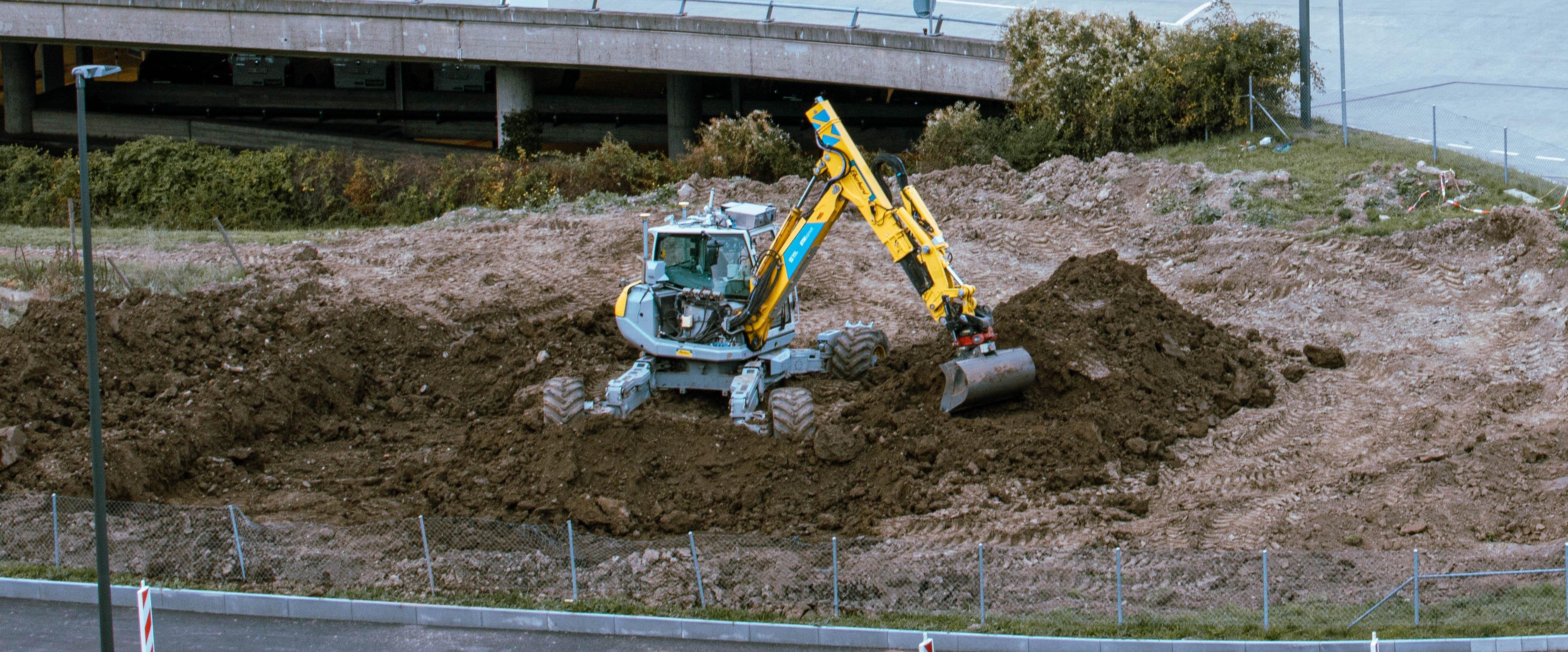}
      \label{fig:real_3}
    \end{minipage}
  }
  \hfill
  \subfloat[]{
      \begin{minipage}{0.48\textwidth}
        \includegraphics[height=3.5cm, width=0.95\linewidth]{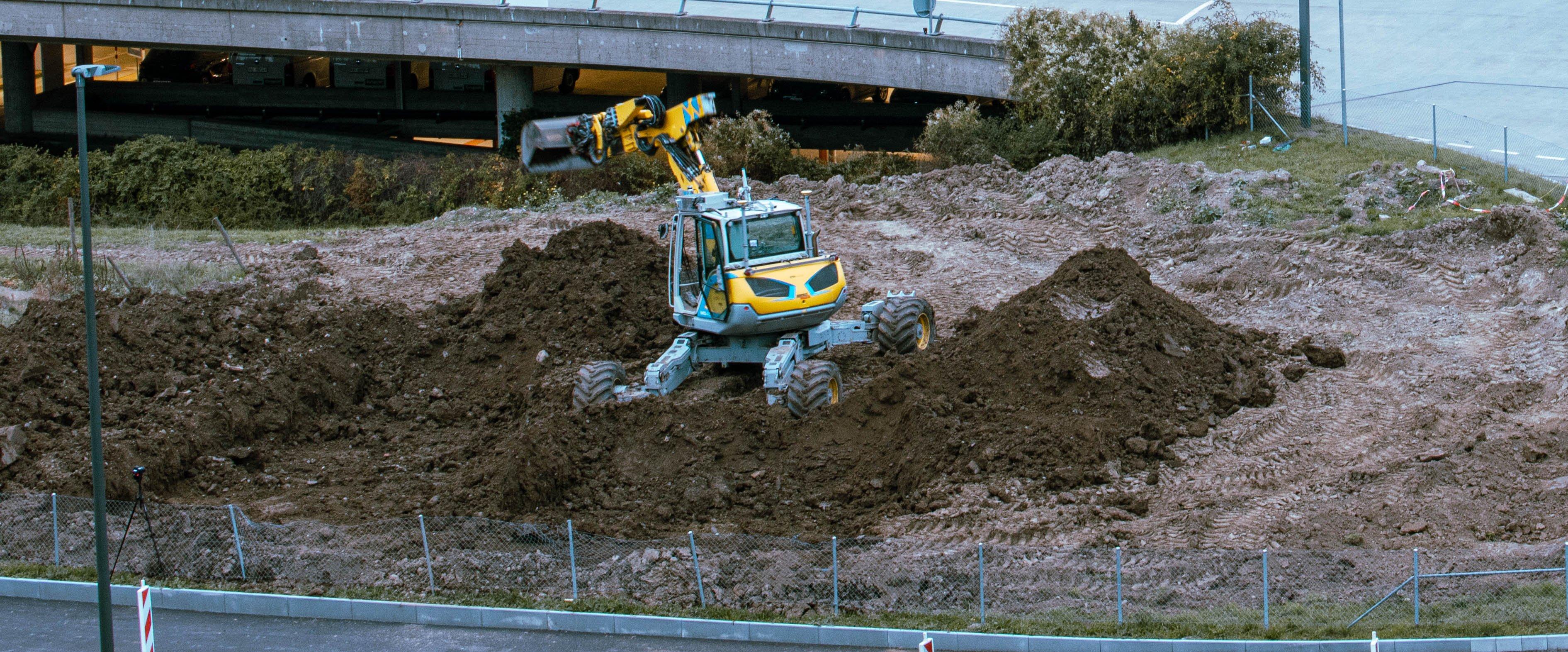}
        \label{fig:real_4}
      \end{minipage}
    }
    \subfloat[]{
      \begin{minipage}{0.48\textwidth}
        \includegraphics[height=3.5cm, width=0.95\linewidth]{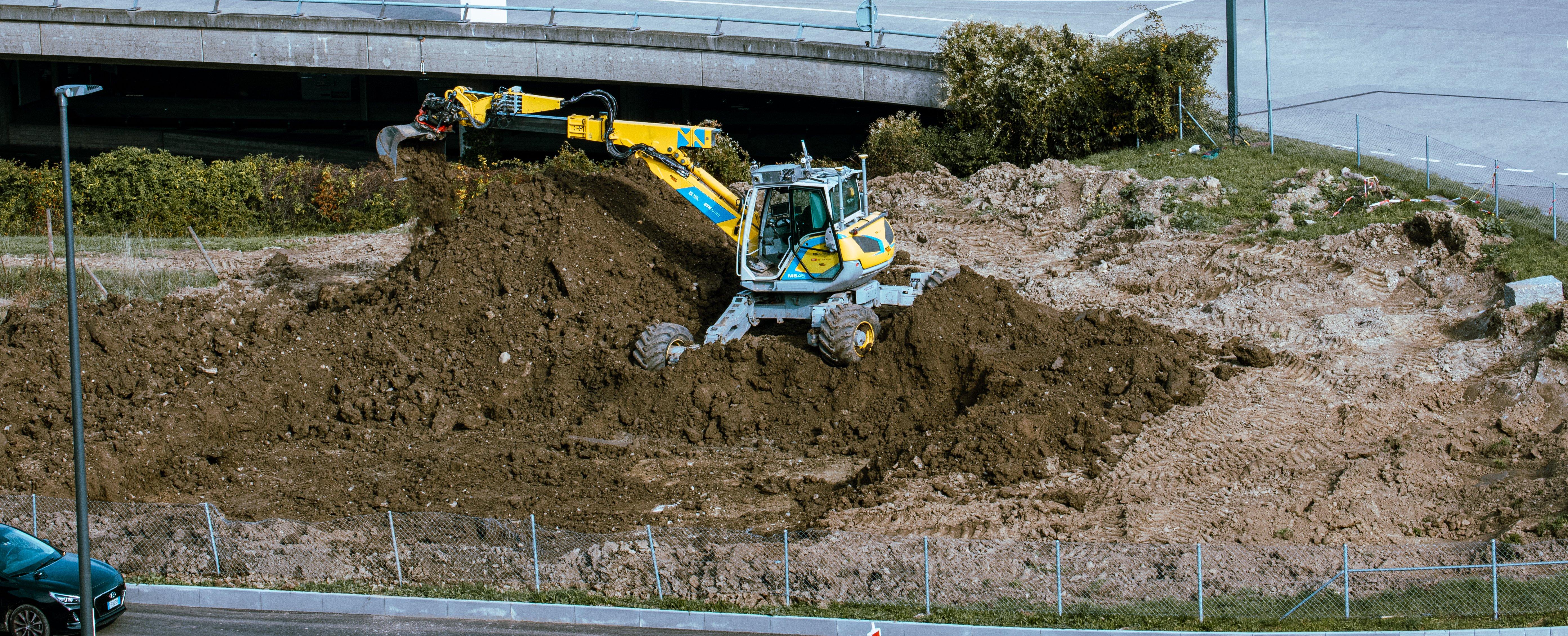}
        \label{fig:real_5}
      \end{minipage}
    }
    \hfill
    \subfloat[]{
      \begin{minipage}{0.48\textwidth}
        \includegraphics[height=3.5cm, width=0.95\linewidth]{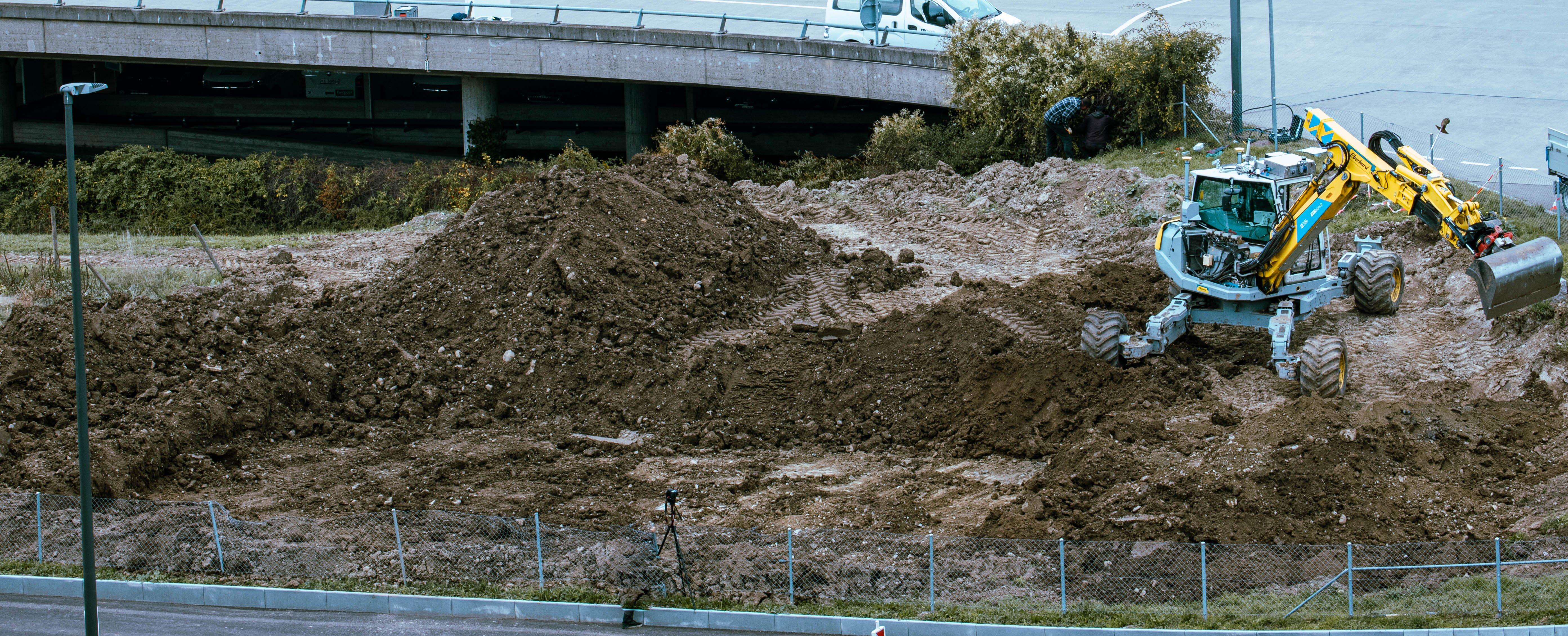}
        \label{fig:real_6}
      \end{minipage}
    }
    \subfloat[]{
      \begin{minipage}{0.48\textwidth}
        \includegraphics[height=3.5cm, width=0.95\linewidth]{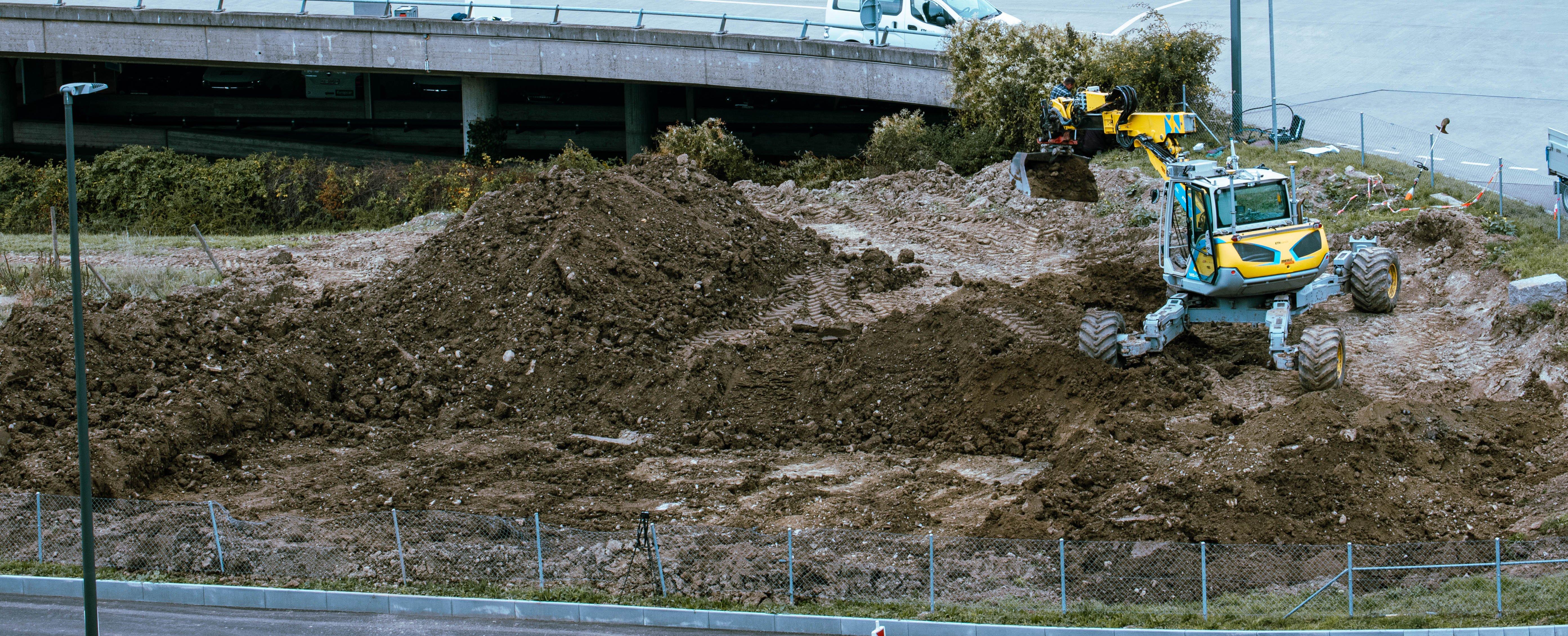}
        \label{fig:real_7}
      \end{minipage}
    }
    \caption{Snapshops of the pit excavation, each image shows a different earthworks planning strategy. \textbf{(a)}: terrain before the excavation. \textbf{(b)}: Dumping on the left side. \textbf{c}: the soil is moved from the front of the dig area to the back-left area, which is still inside the excavation area but is closer to its boundary. \textbf{(d)}: Moving dirt from front-left to back-left zone. \textbf{(e)}: end of the second lane, soil is accumulated behind the first excavated lane. \textbf{(f)} dirt is moved from the front and front-right zone to the back-right zone, which is now reachable. \textbf{(g)}: dumping on the right side. \textbf{(h)}: final result of the excavation.}
  \label{fig:experiments}
\end{figure}

Figures \ref{fig:gearth} display the test field's map, created using Open3DSlam, and the user input layers. The machine was operational for 4 hours and 25 minutes to complete the pit, with a time breakdown in \autoref{tab:time_breakdown}. Figure \ref{fig:experiments} illustrates the digging stages in the test field, previously shown in simulation in Figure \ref{fig:single_cell}. Videos of the excavation stages are available in the supplementary materials\footnote{\url{https://youtu.be/bWw4RRqz_dM}} \footnote{\url{https://youtube.com/playlist?list=PLpIdy9k5iMC9hMnRszYzi8OUj-WYN6UQ-}}.

\begin{table}[!htb]
  \caption{Breakdown of operations' time by state of the excavator}
  \label{tab:time_breakdown}
  \centering
  \begin{tabular}{c c c}
   \hline
  \textbf{State} & \textbf{Mean Duration (std)} & \textbf{Total Duration} \\
  \hline
  Initialize Workspace & 0.45 (0.18) s & 0h 0m \\
  \hline
  Check Workspace & 0.29 (0.13) s & 0h 2m \\
  \hline
  Dig & 14.13 (3.38) s & 1h 51m \\
  \hline
  Arm to Dig Point & 8.04 (3.64) s & 1h 8m \\
  \hline
  Arm to Dump Point and Dump Soil & 9.91 (2.30) s & 1h 1m \\
  \hline
  Find Path Plan & 9.38 (38.20) s & 0h 7m \\
  \hline
  Driving & 34.07 (37.07) s & 0h 12m \\
  \hline
  Retract Arm & 3.96 (1.07) s & 0h 1m \\
  \hline
  \textbf{Total} & - & 4h 25m \\
  \hline
  \end{tabular}
  \end{table}
  
Evaluating the efficiency involves two metrics: cycle time and volume of soil excavated per time unit. The recorded cycle time - the time needed to dig, dispose of soil, and reposition the arm for subsequent digging - is 32.08 (5.47) seconds, similar to the value of 30 seconds reported for human operators \cite{cycle_time}. However, directly comparing the system's efficiency to that of a proficient human operator can be challenging. This complexity arises from the fact that the cycle time is influenced by a multitude of variables, including but not limited to the size and type of excavator used, the properties of the soil, and the employed method of unloading \cite{litvinEvaluationEffectExcavator2020}. Hence, while drawing a direct comparison between the system and the human operator is tempting, the multitude of influencing factors suggests a nuanced evaluation is necessary.

Table \ref{tab:workspace_breakdown} details efficiency metrics based on excavation area and modes. The overall times include driving, digging, and refinement. The total volume is determined by comparing the initial map with the map after excavation in each workspace. The volume excavated in the front area represents the amount of undug dirt in the pit, whereas the volume moved in front-side regions corresponds to piles of dirt that need further relocation.

\begin{table}[!htb]
  \caption{Breakdown of efficiency metrics}
  \label{tab:workspace_breakdown}
  \centering
  \begin{tabular}{c c c c c }
   \hline
  \textbf{Dig Area} & \textbf{Duration [h]} & \textbf{Mean Scoop Vol. [m$^3$]} & \textbf{Vol. [m$^3$]} & \textbf{Dig Eff. [m$^3$/h]} \\
  \hline
  Front & 3h 20m & 0.43 & 142.47 & 42.70 \\
  \hline
  Front-Left & 0h 17m & 0.55 & 16.18 & 54.72 \\
  \hline
  Front-Right & 0h 0m & - & - & - \\
  \hline
  Refinement & 0h 24m & - & - & - \\
  \hline
  Overall including Loose Dirt & 4h 25m & - & 158.77 & 35.95 \\
  \hline
  Overall & 4h 25m & - & 142.47 & 32.35 \\ 
  \end{tabular}
\end{table}

\begin{figure}[!htb]
    \centering
    \begin{minipage}[c]{0.45\textwidth}
        \centering
        \includegraphics[width=0.8\linewidth]{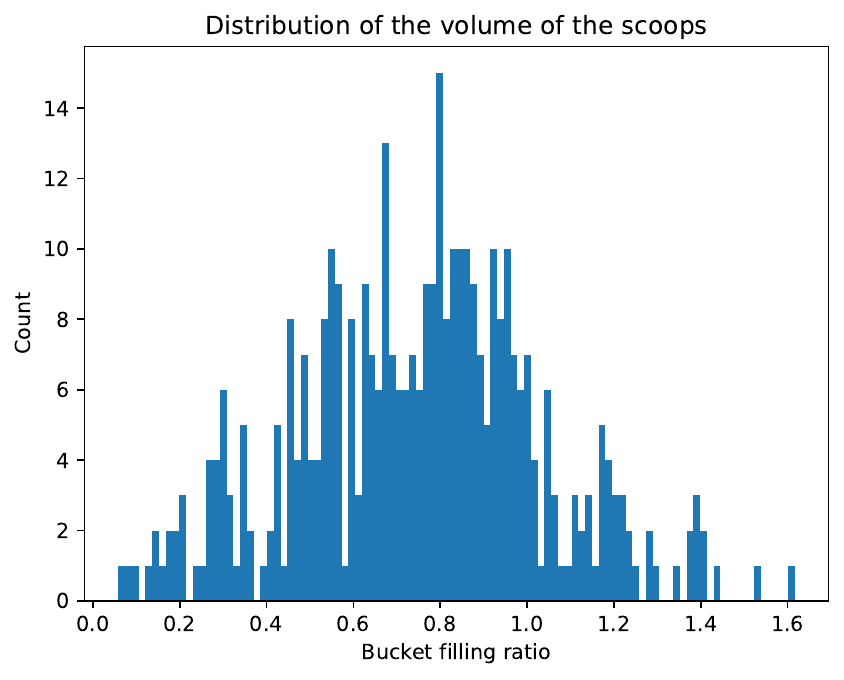}
        \captionof{figure}{Distribution of the scoops' volume expressed as a ratio of the bucket volume (0.6 $m^3$).}
        \label{fig:scoops}
    \end{minipage}%
    \hfill 
    \begin{minipage}[c]{0.45\textwidth}
        \centering
        \begin{tabular}{c c c}
            \hline
            \textbf{Condition} & \textbf{Mean Level} & \textbf{Abs. Err.} \\
            \hline
            Before ref. & -3.5 & 9.7 \\
            \hline
            After ref. & -1.3 & 7.2 \\
            \hline
        \end{tabular}
        \captionof{table}{Digging precision statistics in centimeters.}
        \label{tab:precision}
    \end{minipage}
\end{figure}

To better understand scooping efficiency, we analyzed the distribution of soil volume dug for each scoop. A total of 380 scoops were executed during the operation, excluding bucket motions used for refining the dig area. The distribution of scoop volume, shown in Figure \ref{fig:scoops}, appears to follow a Gaussian distribution with a mean of 0.45 $m^3$ and a standard deviation of 0.16. The bucket volume is approximately 0.6 $m^3$, but it can be overfilled for more efficient digging. Lower volume scoops are often used to remove the last remaining soil from the dig area or result from constraints on digging geometry, such as being close to the edges of the pit with the bucket.


The summary of the excavation precision is outlined in Table 1 (formerly referred to as \autoref{tab:precision}). The mean level metric represents the average height difference to the desired elevation. Significant inconsistencies in the flatness of the pit's bottom are evident from the values presented. Before refinement, the average absolute error measures around 10 centimeters. However, through refinement, this error was successfully reduced by 3 centimeters.

Several factors hinder finer precision, such as unidirectional grading, sensor noise from the lidar system, and the global positioning system's height covariance. Moreover, imprecise tracking from using a singular set of low-level PID gains to manage arm velocity causes potential errors in varied arm configurations during digging and grading.






\subsection{Different Excavation Geometries}
\begin{figure}[!htbp]
  \captionsetup[subfloat]{farskip=0pt,captionskip=0pt}
  \centering
  \subfloat[]{
    \begin{minipage}{0.33\textwidth}
      \centering
      \includegraphics[height=3.3cm,width=\linewidth]{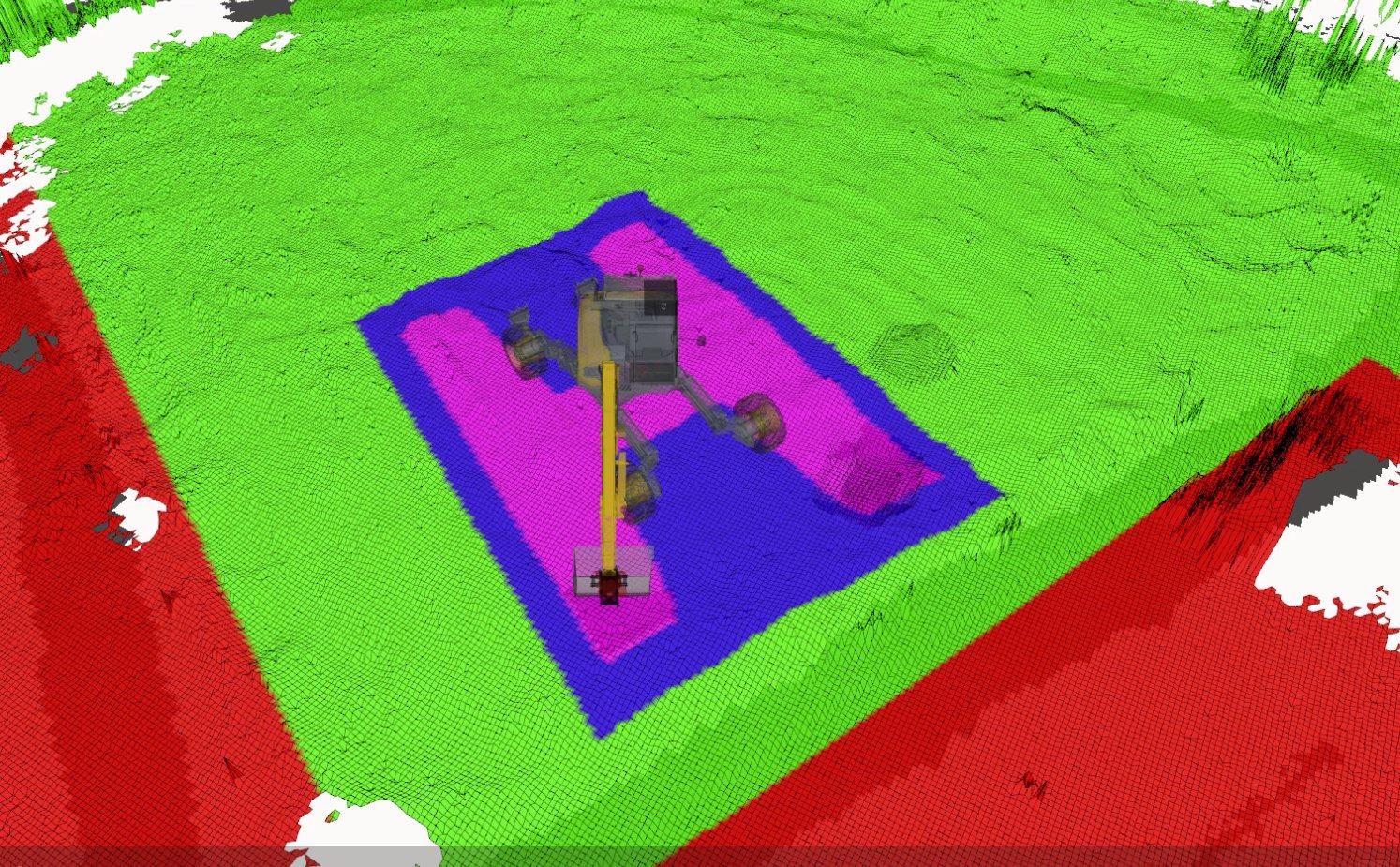}
      \label{fig:h1}
    \end{minipage}
  }
  \subfloat[]{
    \begin{minipage}{0.33\textwidth}
      \centering
      \includegraphics[height=3.3cm,width=\linewidth]{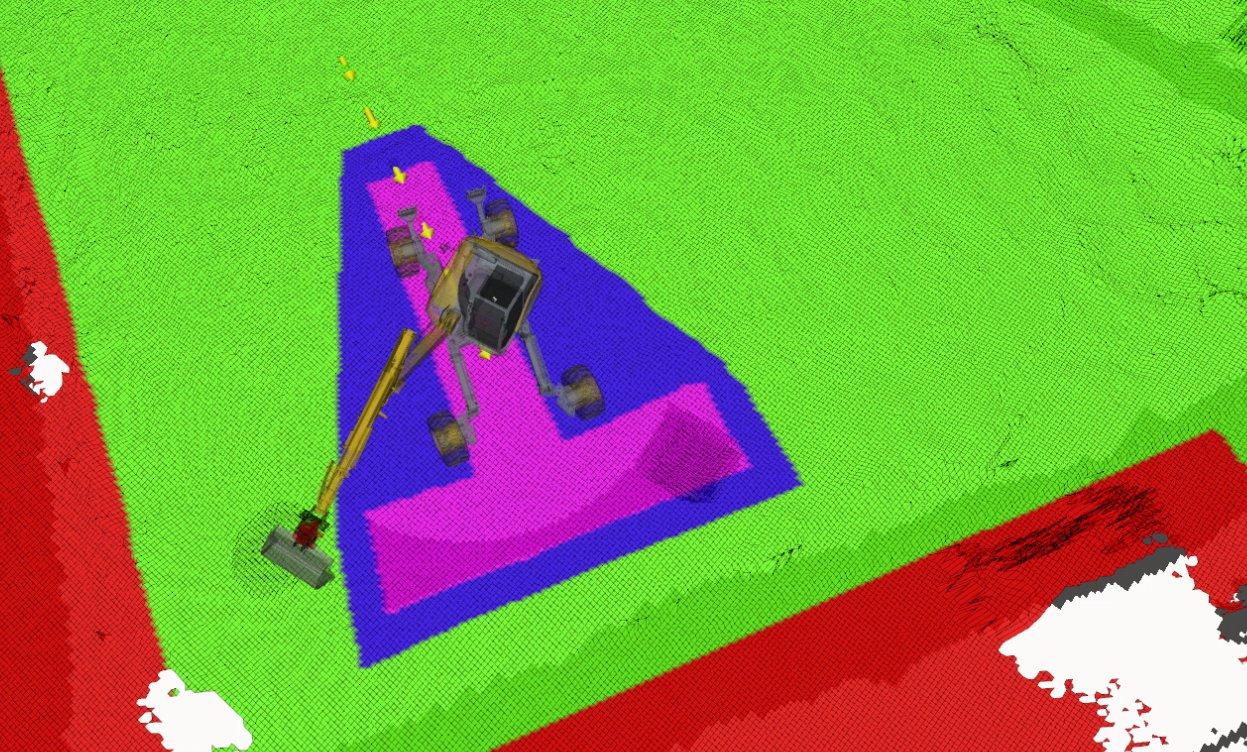}
      \label{fig:t1}
    \end{minipage}
  }
    \subfloat[]{
      \begin{minipage}{0.33\textwidth}
        \centering
        \includegraphics[height=3.3cm,width=\linewidth]{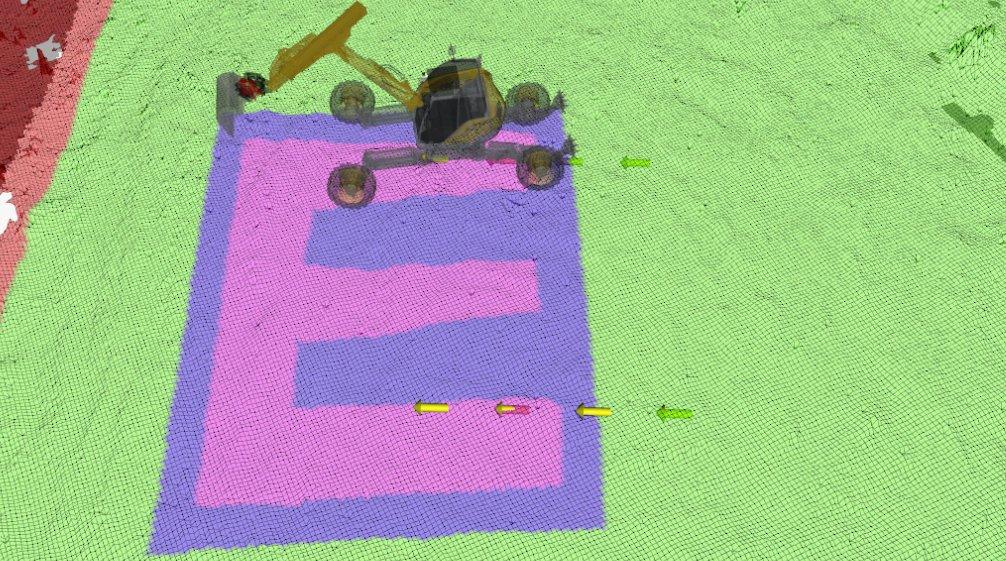}
        \label{fig:e1}
      \end{minipage}
    }
    \hfill
    \subfloat[]{
      \begin{minipage}{0.33\textwidth}
        \includegraphics[height=3.3cm,width=\linewidth]{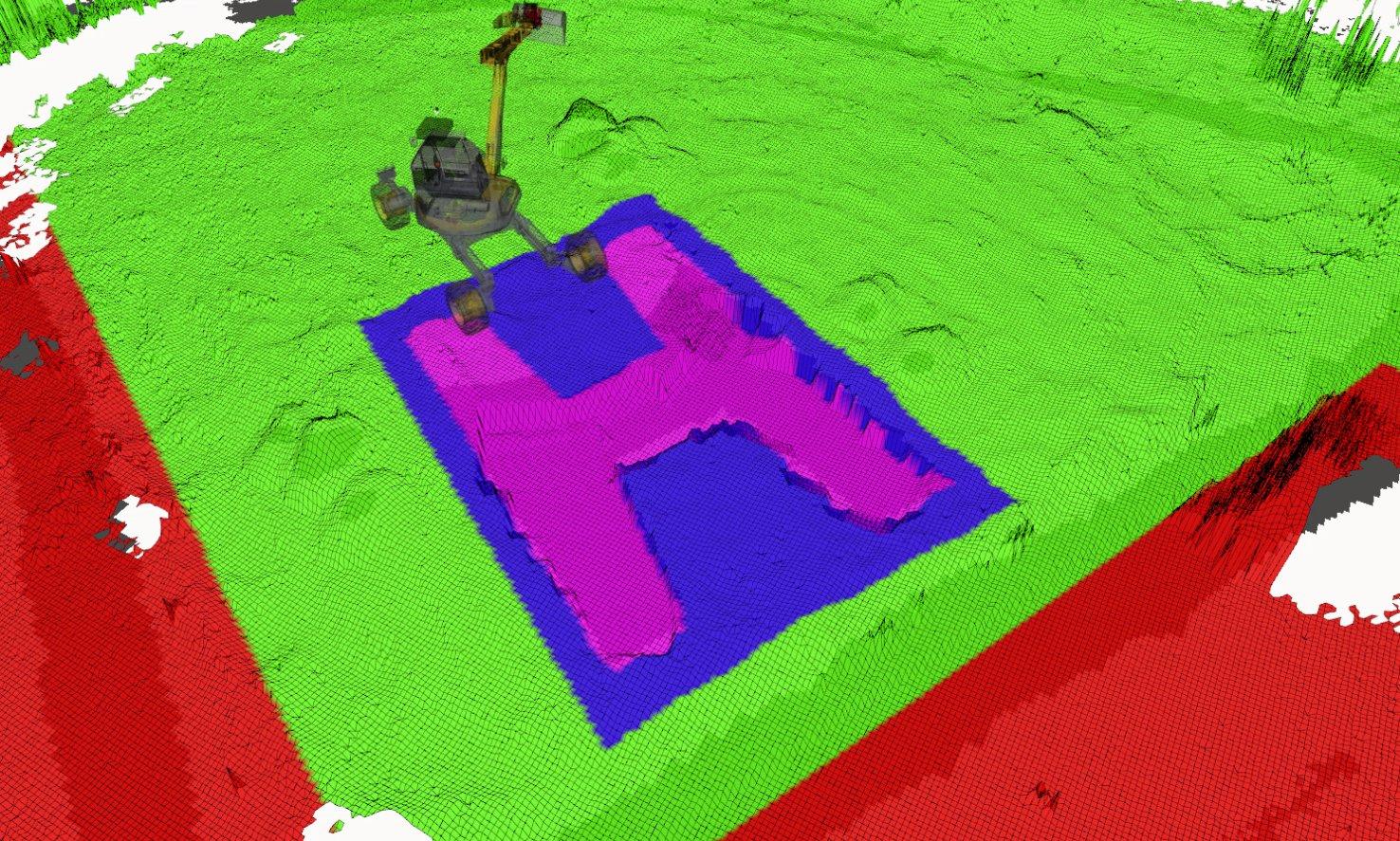}
        \label{fig:h2}
      \end{minipage}
    }
    \subfloat[]{
      \begin{minipage}{0.33\textwidth}
        \includegraphics[height=3.3cm,width=\linewidth]{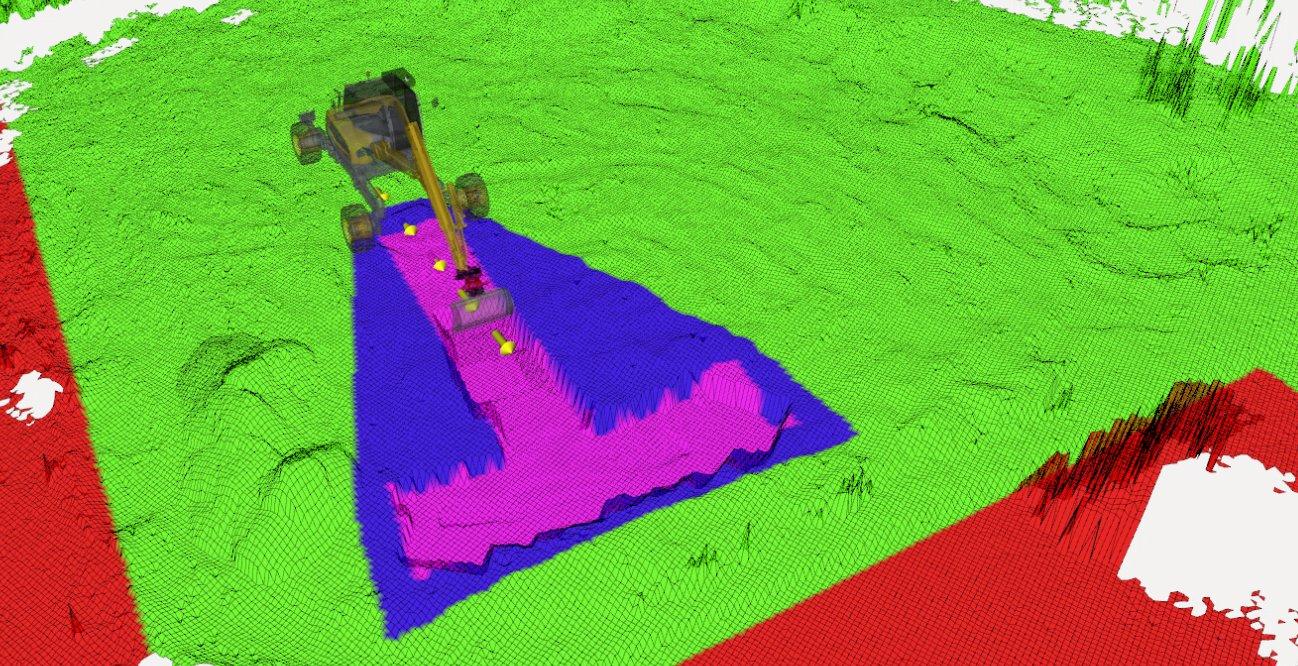}
        \label{fig:t2}
      \end{minipage}
    }
    \subfloat[]{
      \begin{minipage}{0.33\textwidth}
        \includegraphics[height=3.3cm,width=\linewidth]{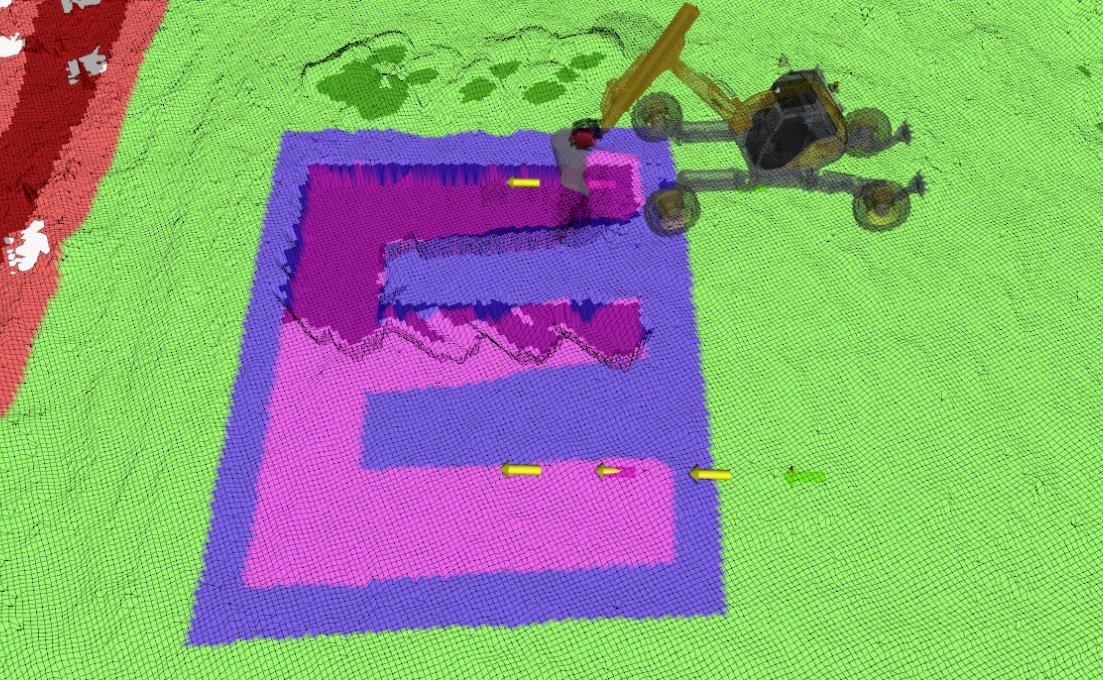}
        \label{fig:e2}
      \end{minipage}
    }
    \hfill
    \subfloat[]{
      \begin{minipage}{0.33\textwidth}
        \includegraphics[height=3.3cm,width=\linewidth]{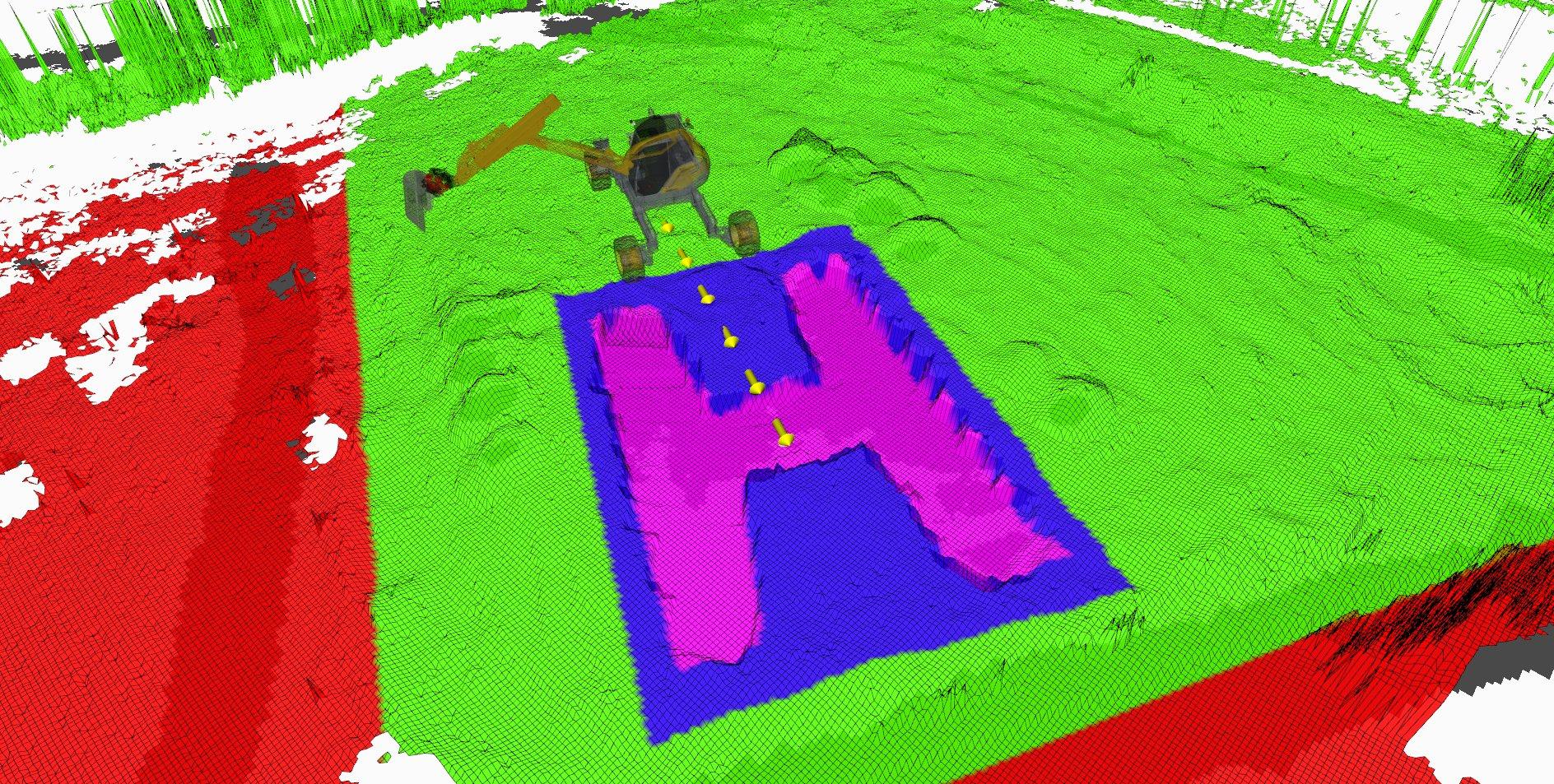}
        \label{fig:h3}
      \end{minipage}
    }
    \subfloat[]{
      \begin{minipage}{0.33\textwidth}
        \includegraphics[height=3.3cm,width=\linewidth]{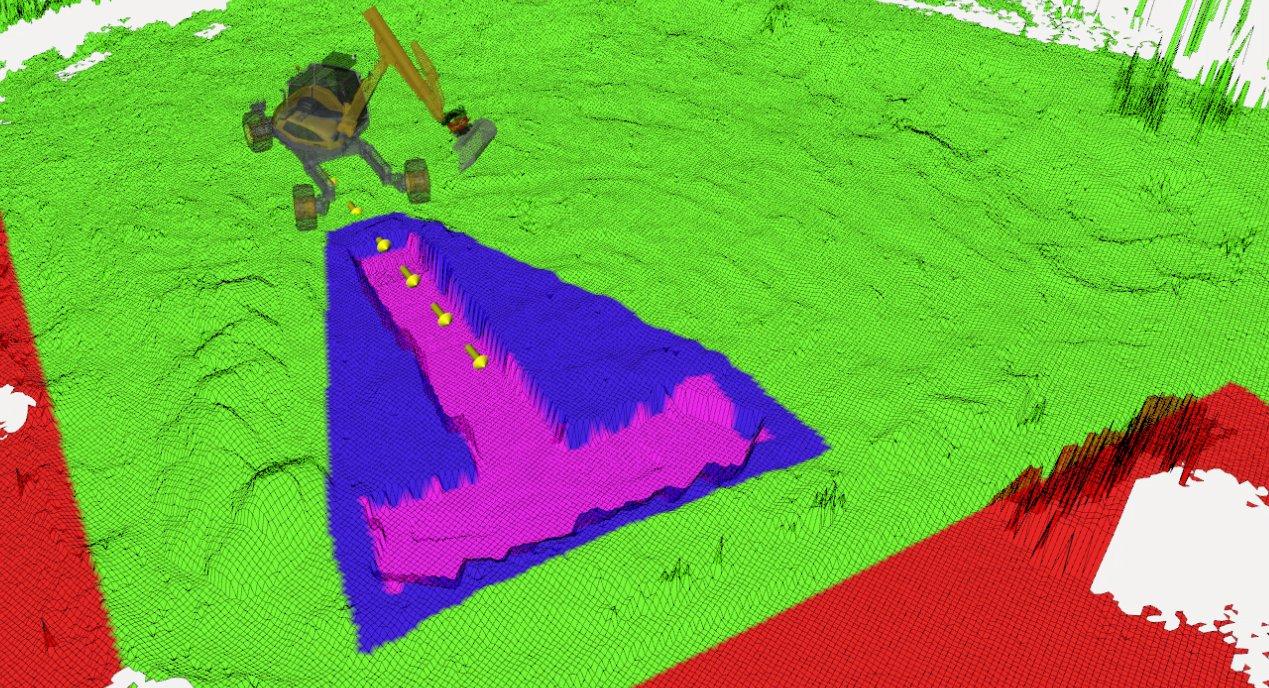}
        \label{fig:t3}
      \end{minipage}
    }
    \subfloat[]{
      \begin{minipage}{0.33\textwidth}
        \includegraphics[width=\linewidth]{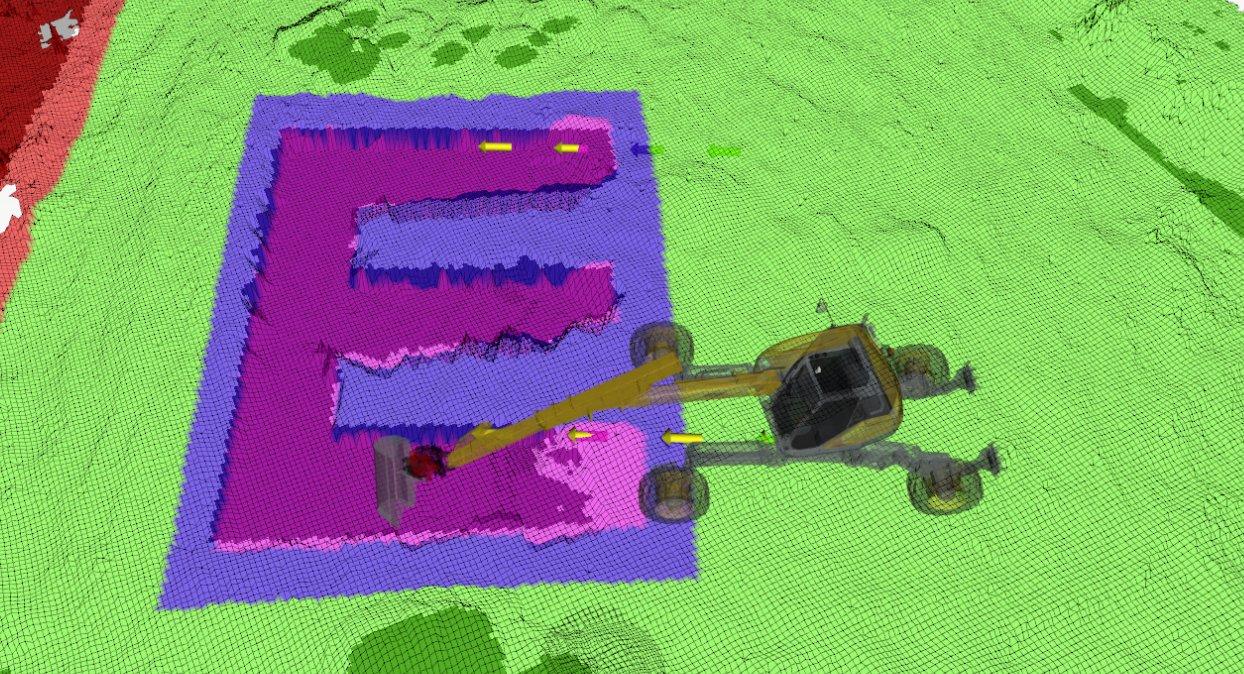}
        \label{fig:e3}
      \end{minipage}
    } 
    \caption{Snapshots of the letters' excavation process in simulation, the target depth is 0.7 meters. Outside of no dumping close to the letters' edges, no further dumping constraint has been added. \textbf{(a)}: H letter. \textbf{(b)}: T letter. \textbf{(c)}: E letter.}
    \label{fig:letters}
  \end{figure}

The system generates plans for various excavation geometries, such as carving the letters H, T, and E on the ground. To protect the edges of the letters, a non-permanent dump area surrounds each letter. Figure \ref{fig:letters} displays snapshots of the excavation process using the same map as in Figure \ref{fig:single_cell}.

\subsection{Excavation Constraints}
\begin{figure}[!htbp]
  \captionsetup[subfloat]{farskip=0pt,captionskip=0pt}
  \centering
  \subfloat[]{
    \begin{minipage}{0.49\textwidth}
      \centering
      \includegraphics[height=4.5cm, width=0.95\linewidth]{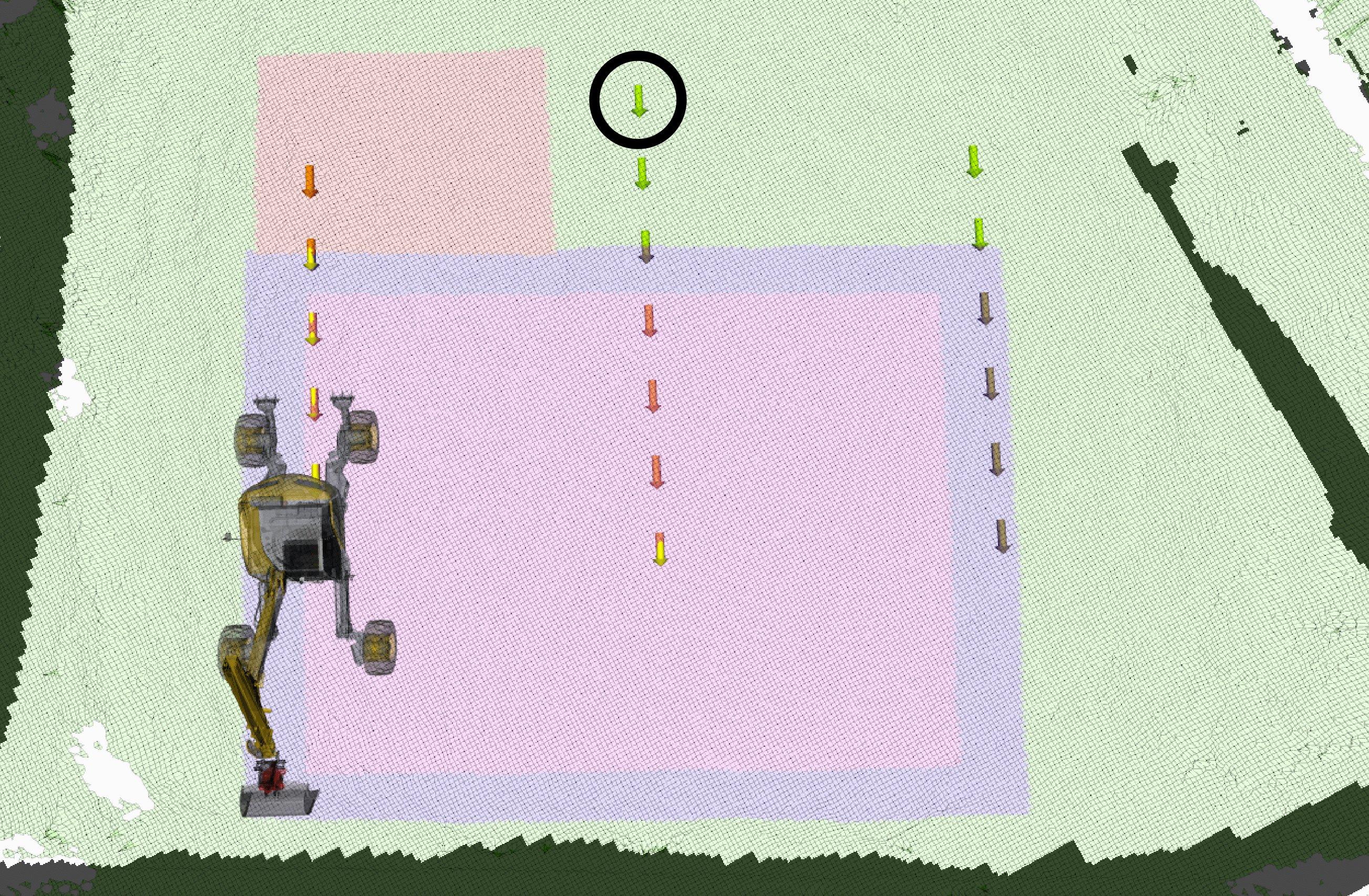}
      \label{fig:22_1_0}
    \end{minipage}
  }
  \subfloat[]{
    \begin{minipage}{0.49\textwidth}
      \centering
      \includegraphics[height=4.5cm, width=0.95\linewidth]{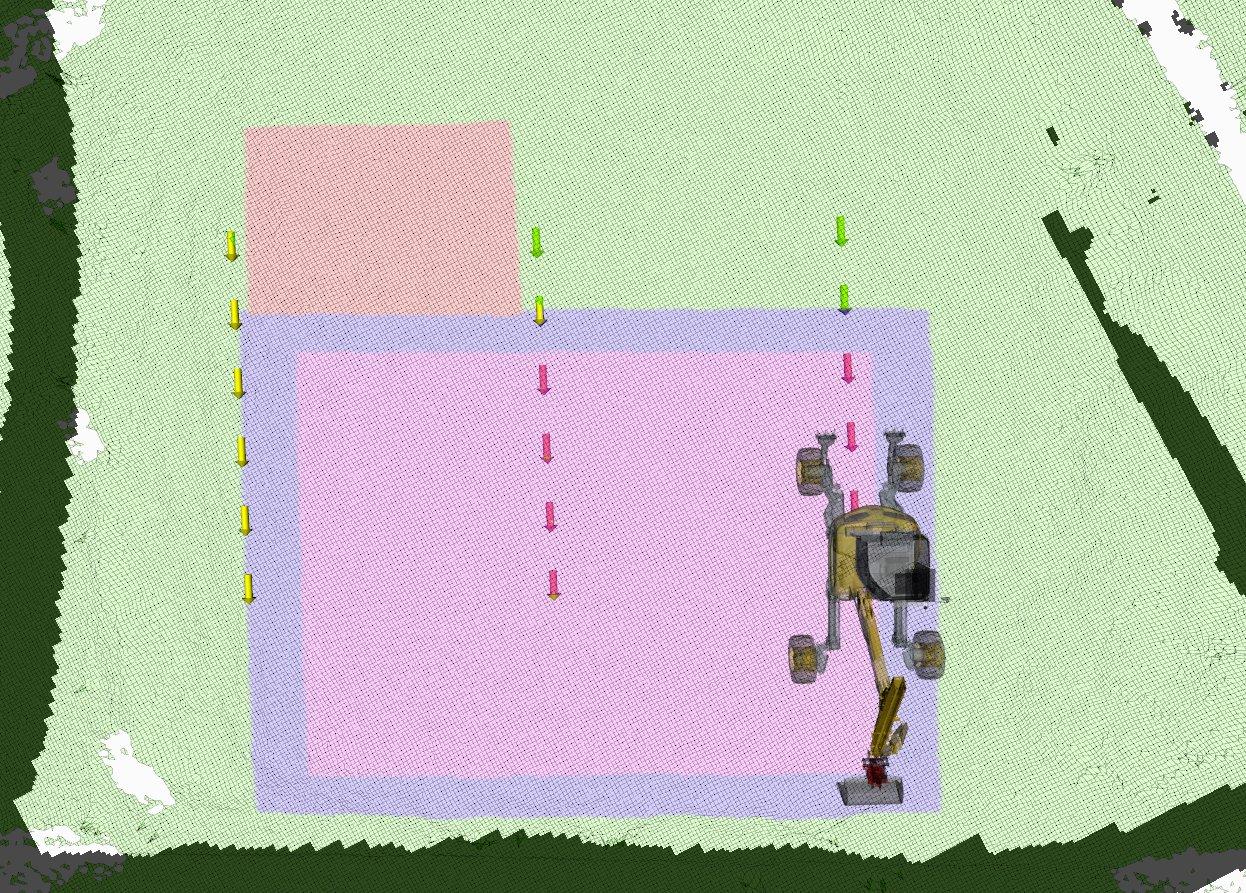}
      \label{fig:22_2_0}
    \end{minipage}
  }
    \hfill
    \subfloat[]{
      \begin{minipage}{0.49\textwidth}
        \centering
        \includegraphics[height=4.5cm, width=0.95\linewidth]{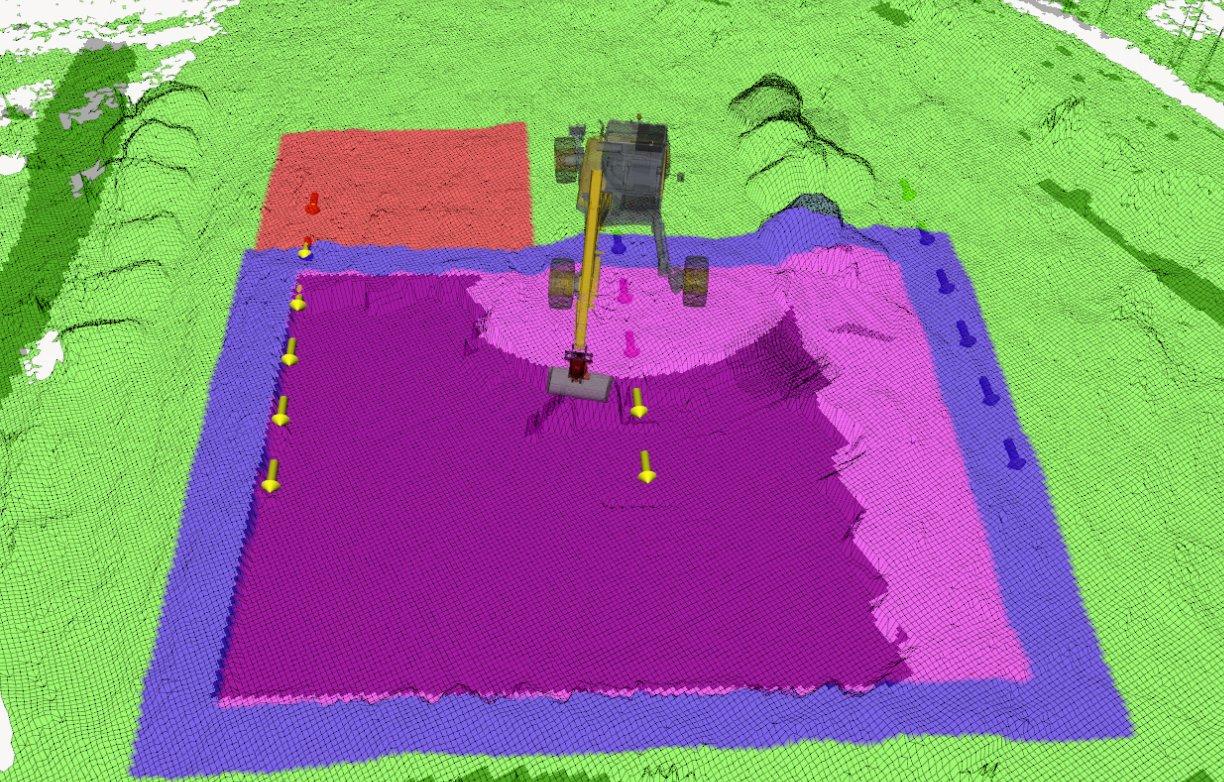}
        \label{fig:22_1_1}
      \end{minipage}
    }
    \subfloat[]{
      \begin{minipage}{0.49\textwidth}
        \centering
        \includegraphics[height=4.5cm, width=0.95\linewidth]{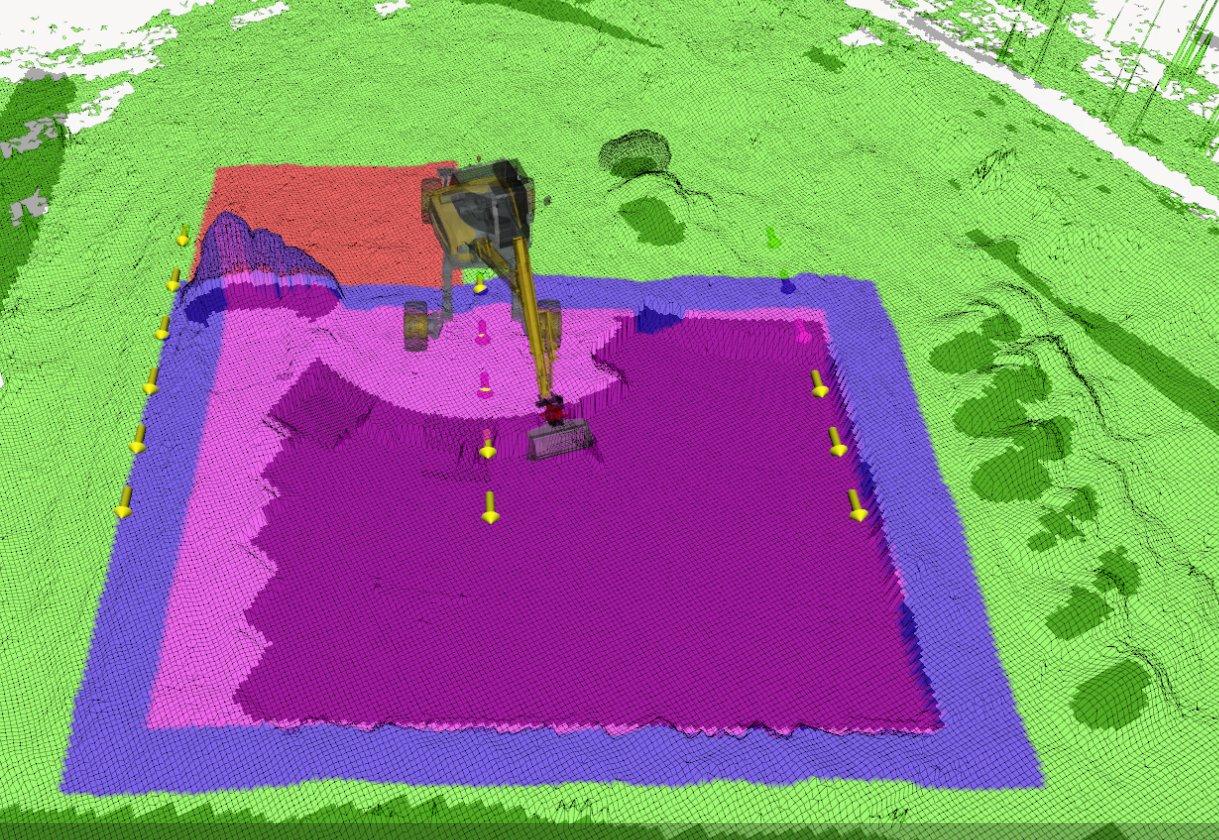}
        \label{fig:22_2_1}
      \end{minipage}
    }
    \hfill
    \subfloat[]{
      \begin{minipage}{0.49\textwidth}
        \centering
        \includegraphics[height=4.5cm, width=0.95\linewidth]{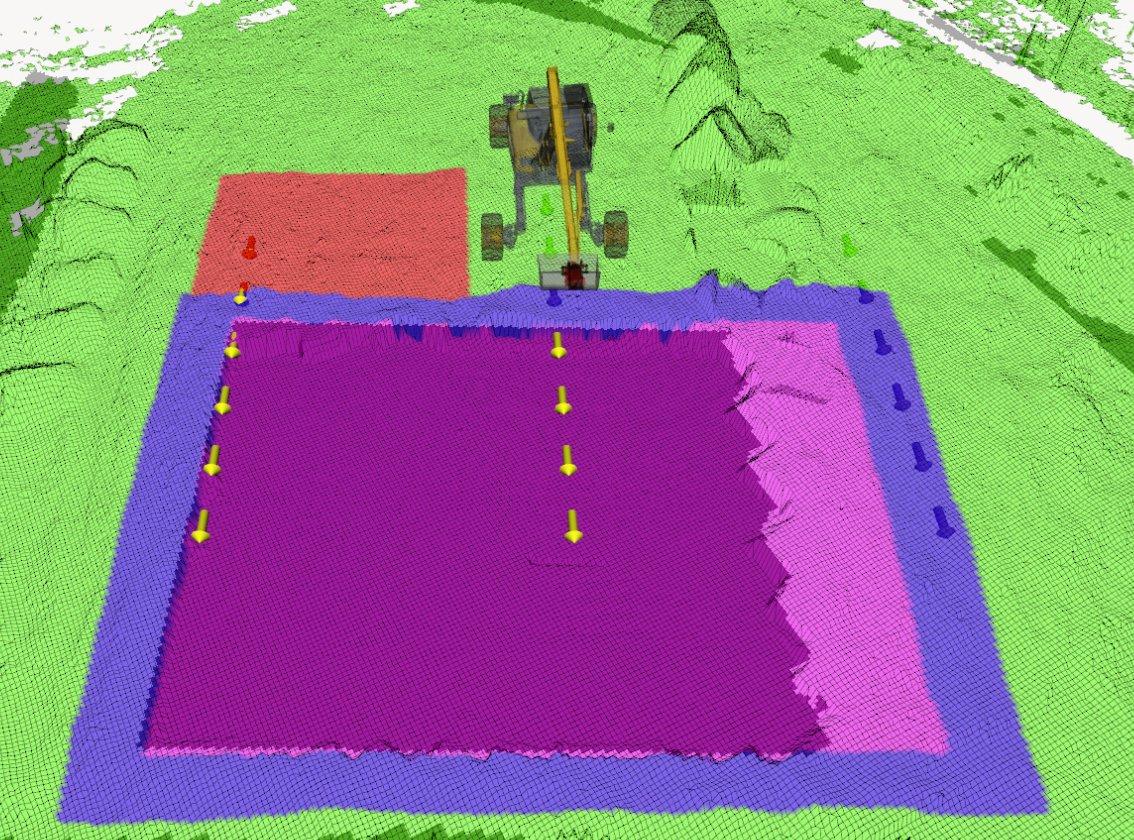}
        \label{fig:22_1_2}
      \end{minipage}
    }
    \subfloat[]{
      \begin{minipage}{0.49\textwidth}
        \centering
        \includegraphics[height=4.5cm, width=0.95\linewidth]{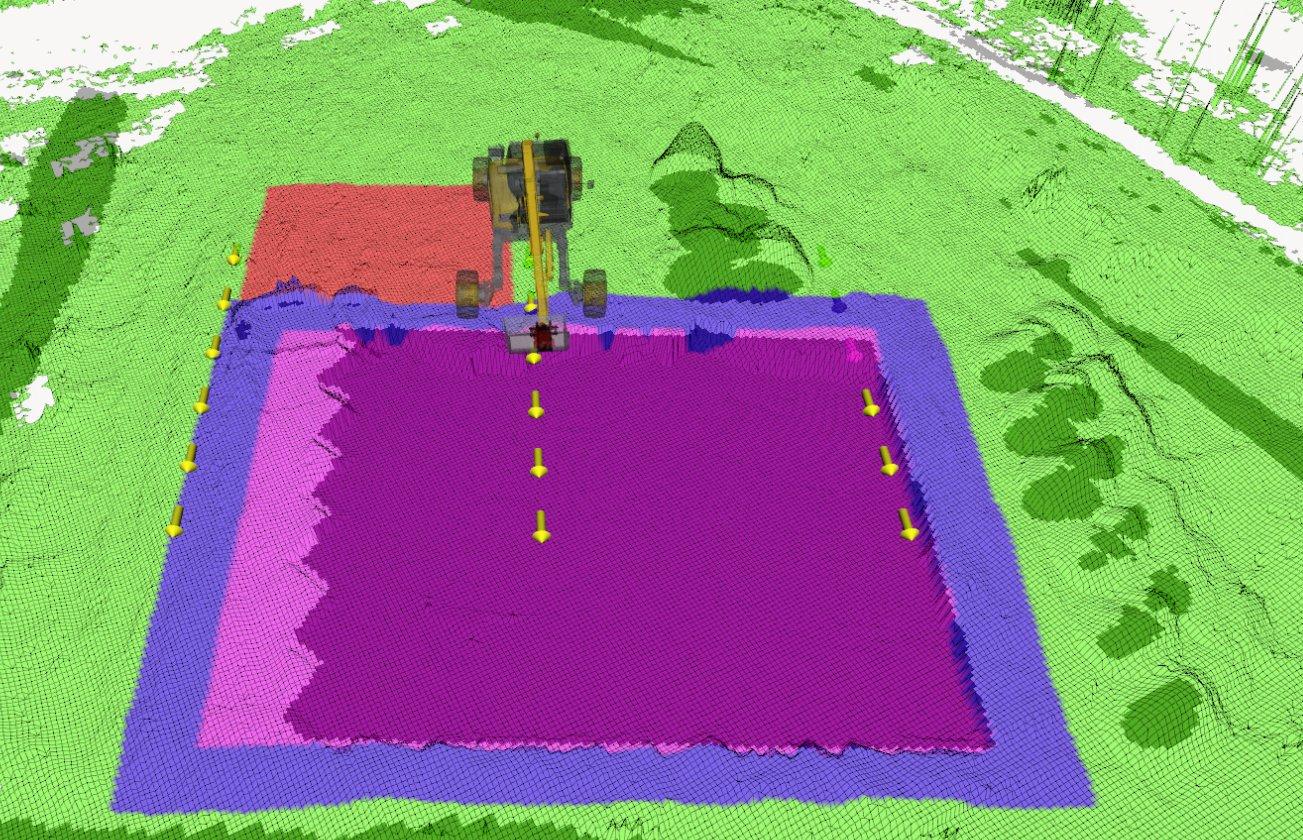}
        \label{fig:22_2_2}
      \end{minipage}
    }
    \hfill
    \subfloat[]{
      \begin{minipage}{0.49\textwidth}
        \centering
        \includegraphics[height=4.5cm, width=0.95\linewidth]{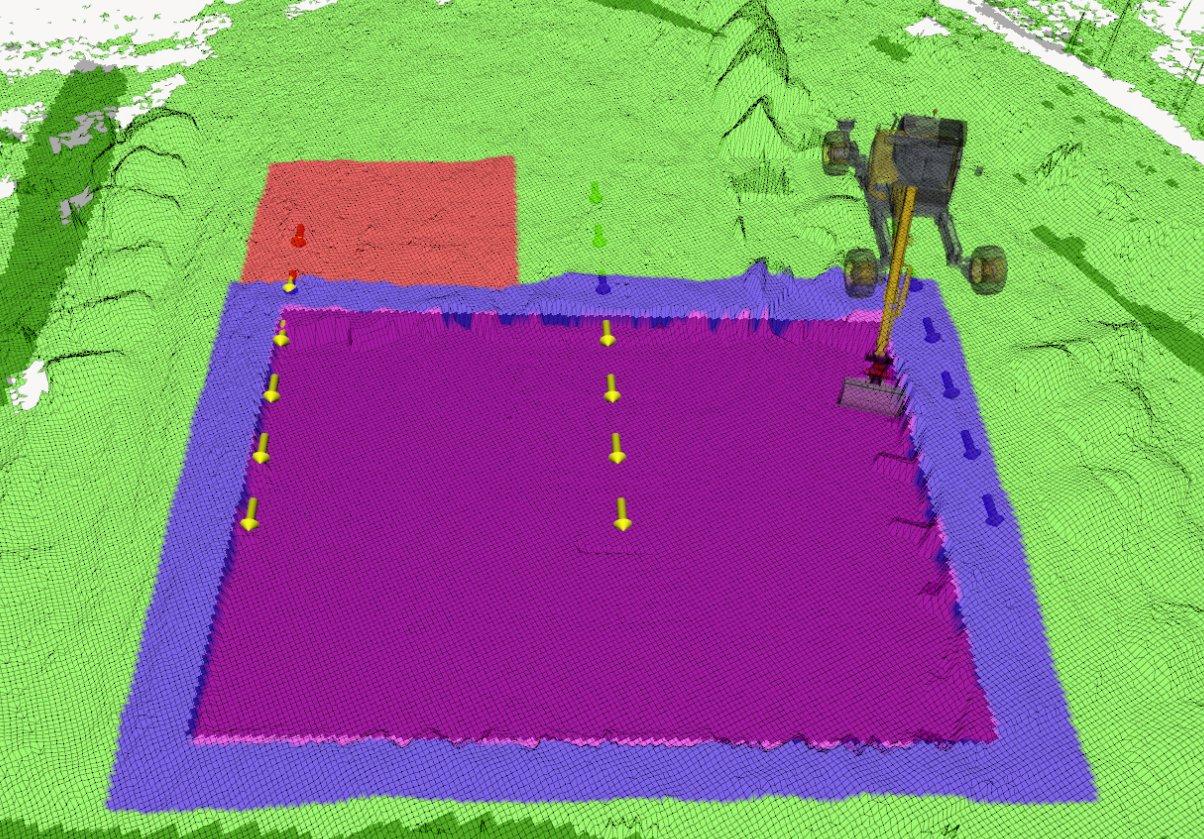}
        \label{fig:22_1_3}
      \end{minipage}
    }
    \subfloat[]{
      \begin{minipage}{0.49\textwidth}
        \centering
        \includegraphics[height=4.5cm, width=0.95\linewidth]{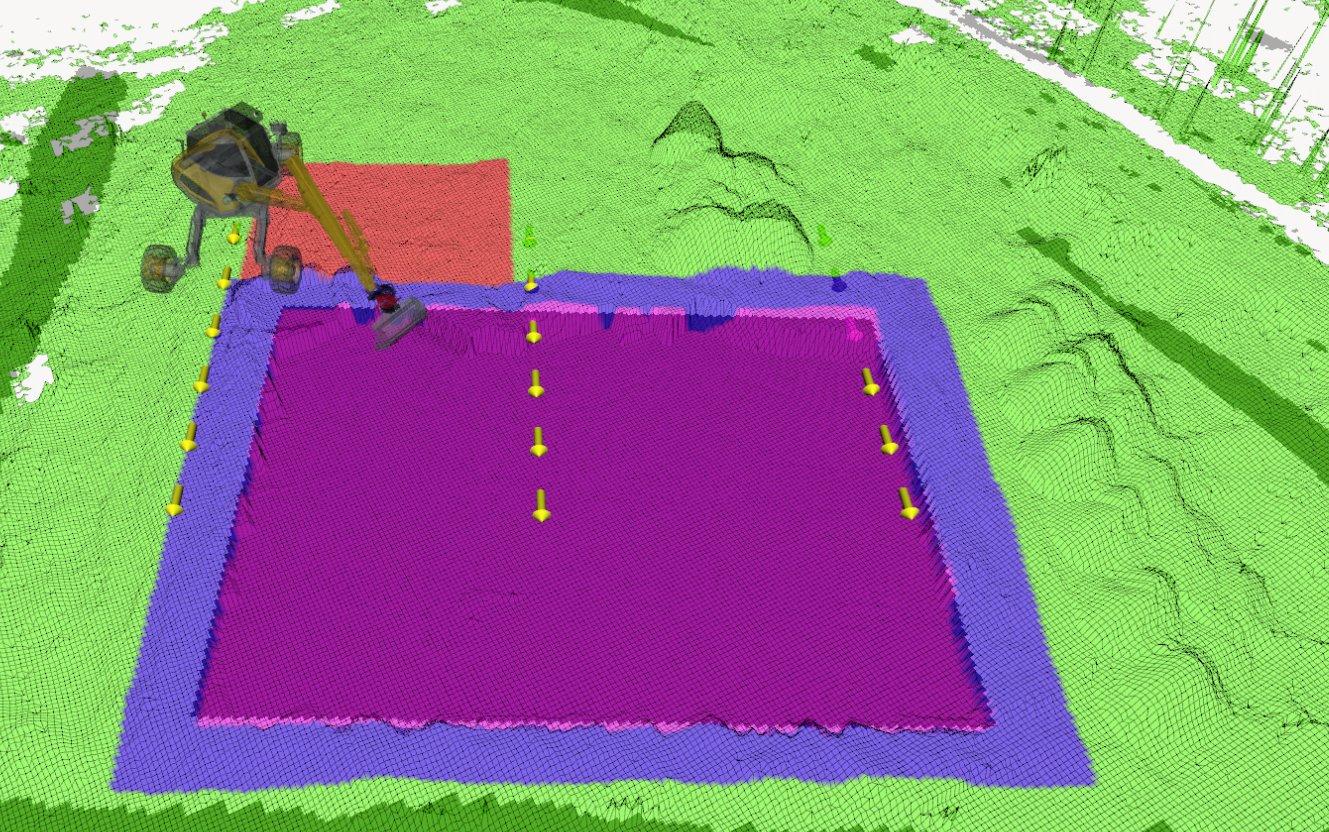}
        \label{fig:22_2_3}
      \end{minipage}
    }
    \caption{The two feasible simulated plans side by side. (a): Plan that starts at corner 1. The extra base pose necessary to complete the workspace is circled. (b): Plan starting at corner 2.}
    \label{fig:v22}
  \end{figure}

\begin{figure}[!htbp]
  \centering
  \subfloat[]{
    \begin{minipage}{0.48\textwidth}
      \centering
      \includegraphics[height=5cm, width=\linewidth]{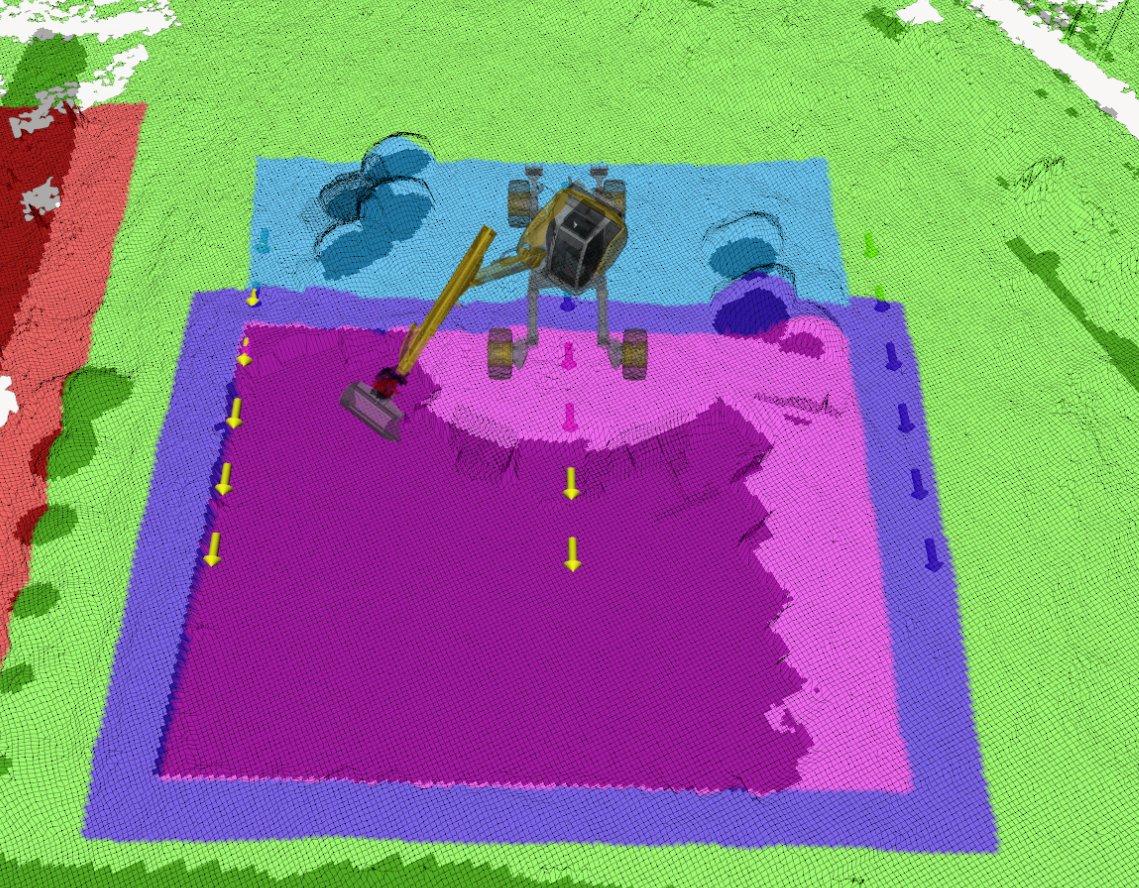}
      \label{fig:24_1}
    \end{minipage}
  }
  \hfill
    \subfloat[]{
    \begin{minipage}{0.48\textwidth}
      \centering
      \includegraphics[height=5cm, width=\linewidth]{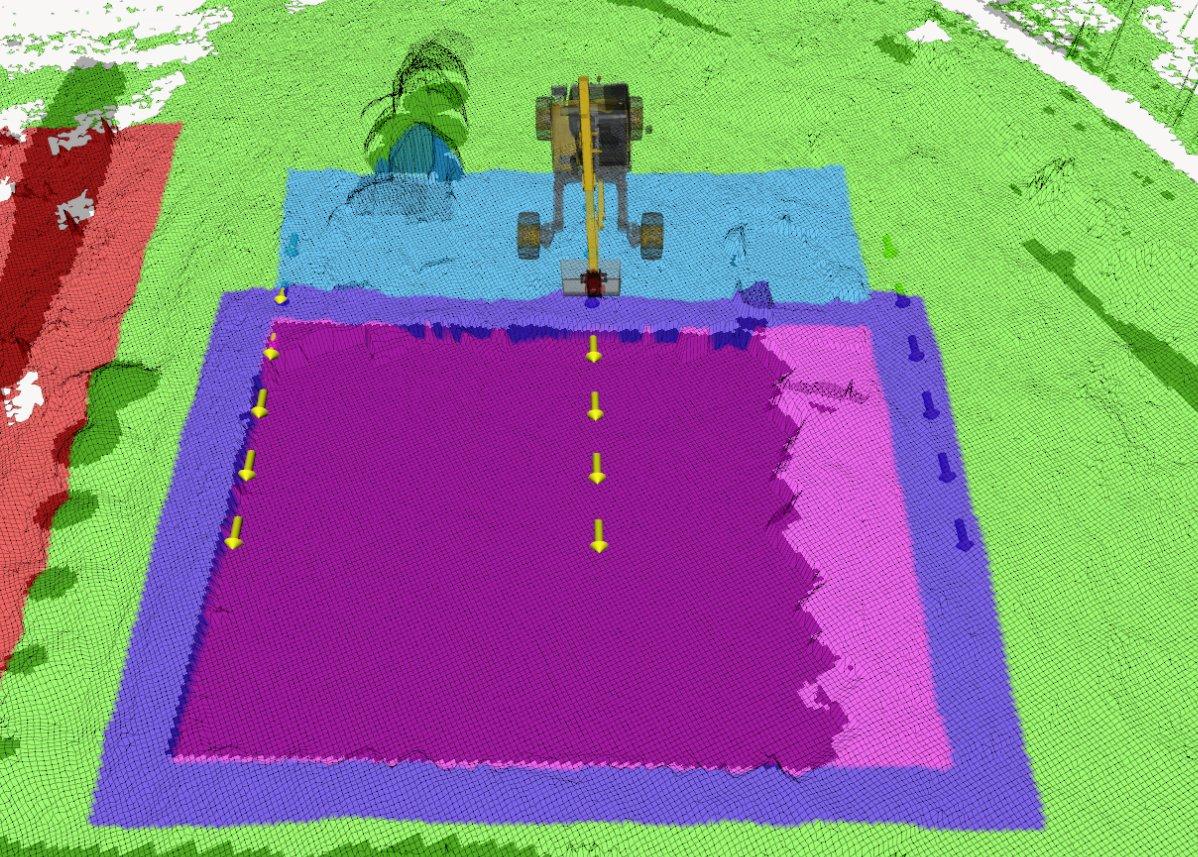}
      \label{fig:24_2}
    \end{minipage}
  }

  \vspace{\floatsep} 
  \subfloat[]{
    \begin{minipage}{0.48\textwidth}
      \centering
      \includegraphics[height=5cm, width=\linewidth]{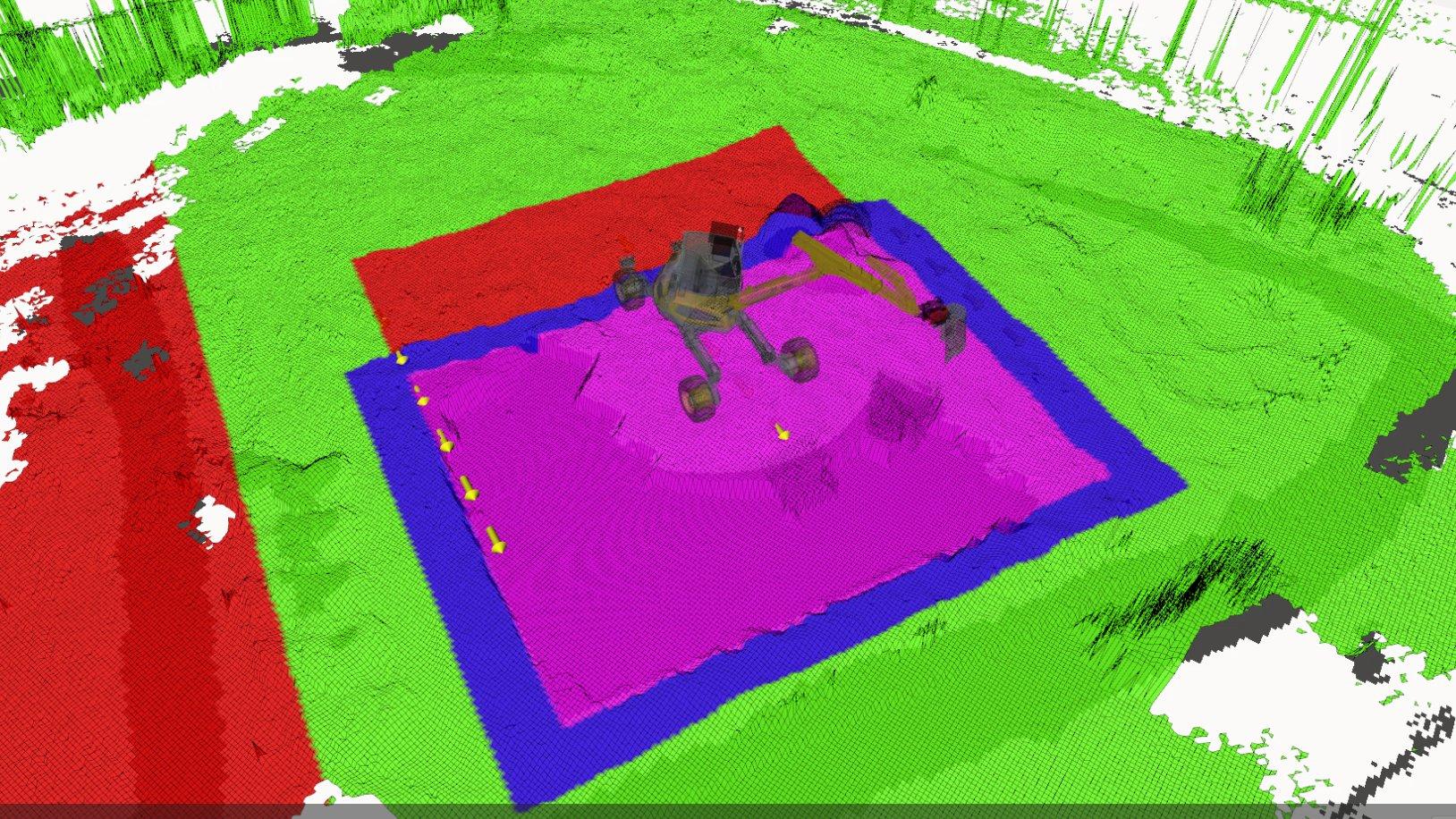}
      \label{fig:v26_1}
    \end{minipage}
}
  \hfill
  \subfloat[]{
    \begin{minipage}{0.48\textwidth}
      \centering
      \includegraphics[height=5cm, width=\linewidth]{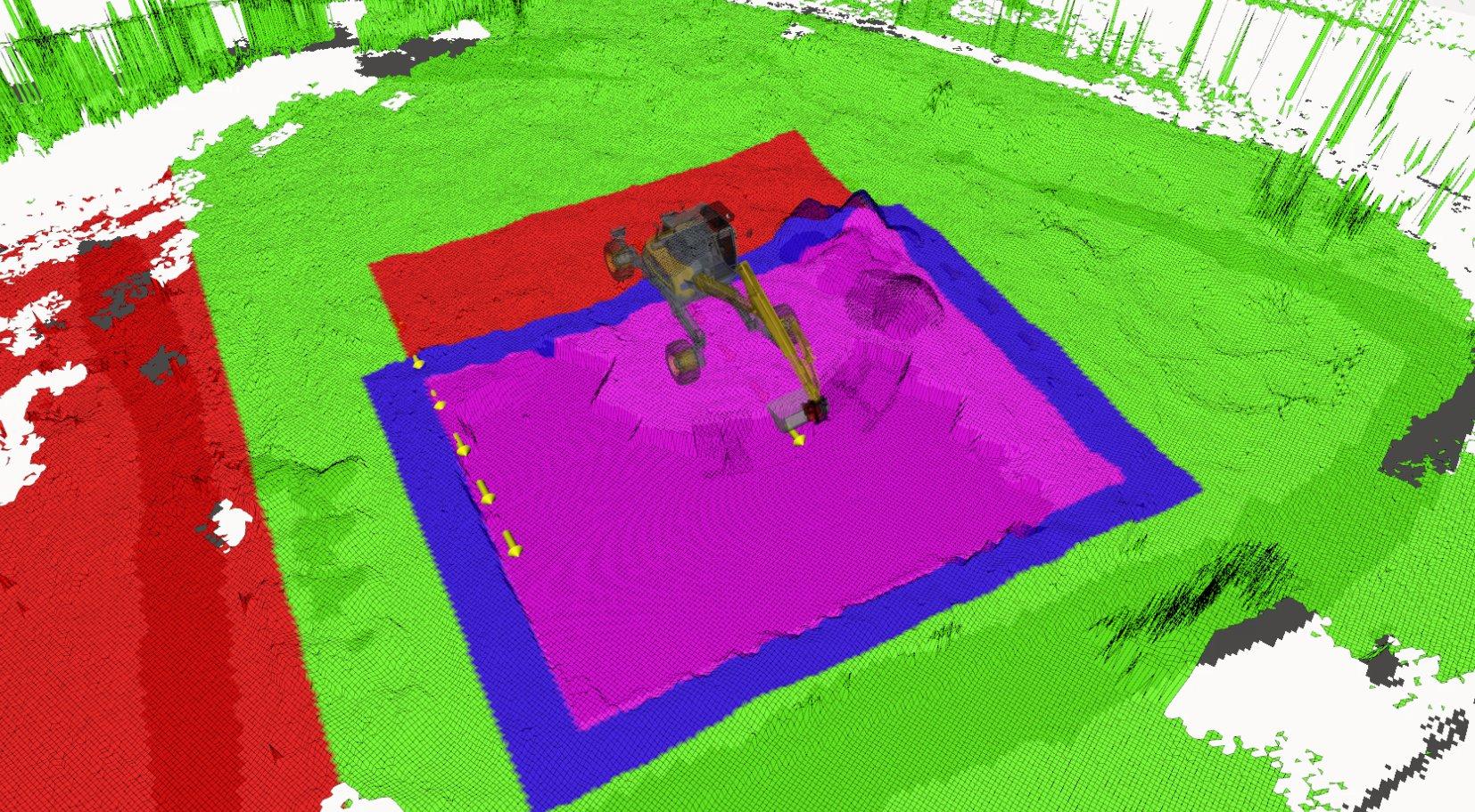}
      \label{fig:v26_2}
    \end{minipage}
  }

  \vspace{\floatsep} 

  \subfloat[]{
    \begin{minipage}{0.48\textwidth}
      \centering
      \includegraphics[height=5cm, width=\linewidth]{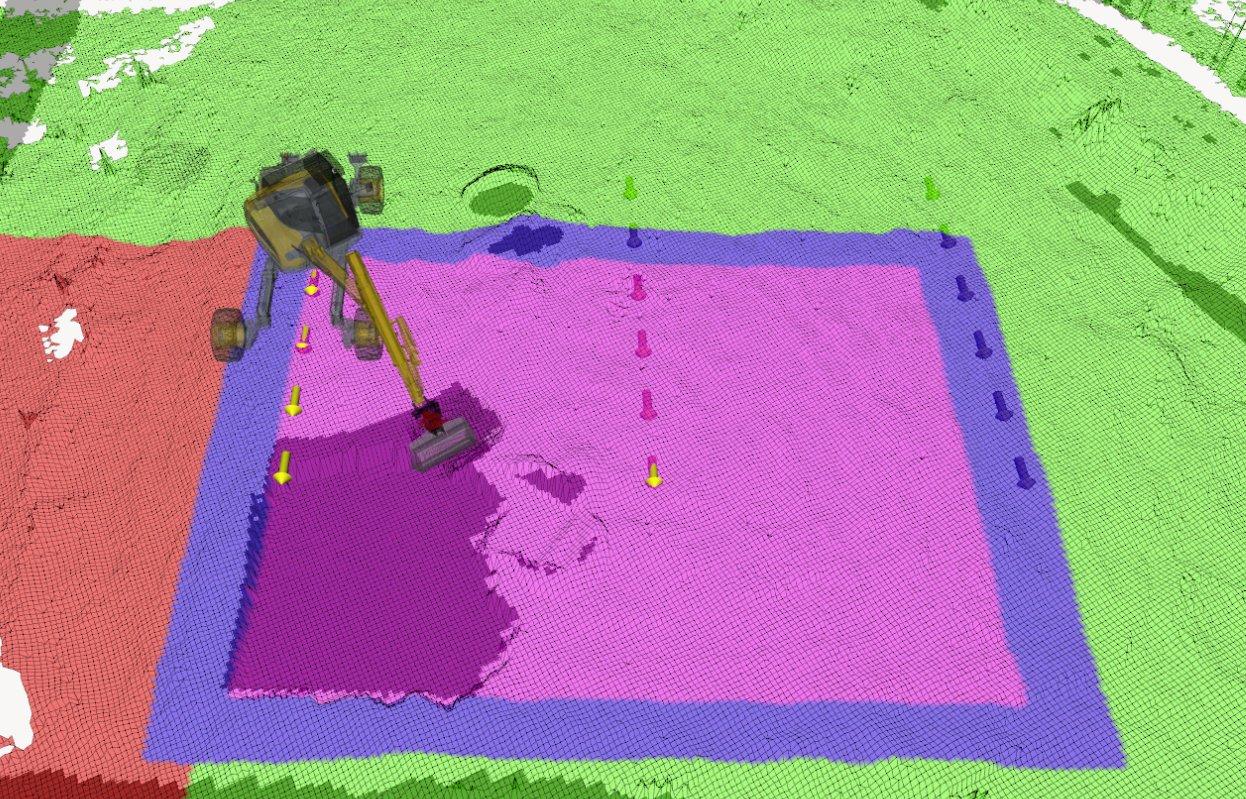}
      \label{fig:v25_2}
    \end{minipage}
  }
  \hfill
  \subfloat[]{
    \begin{minipage}{0.48\textwidth}
      \centering
      \includegraphics[height=5cm, width=\linewidth]{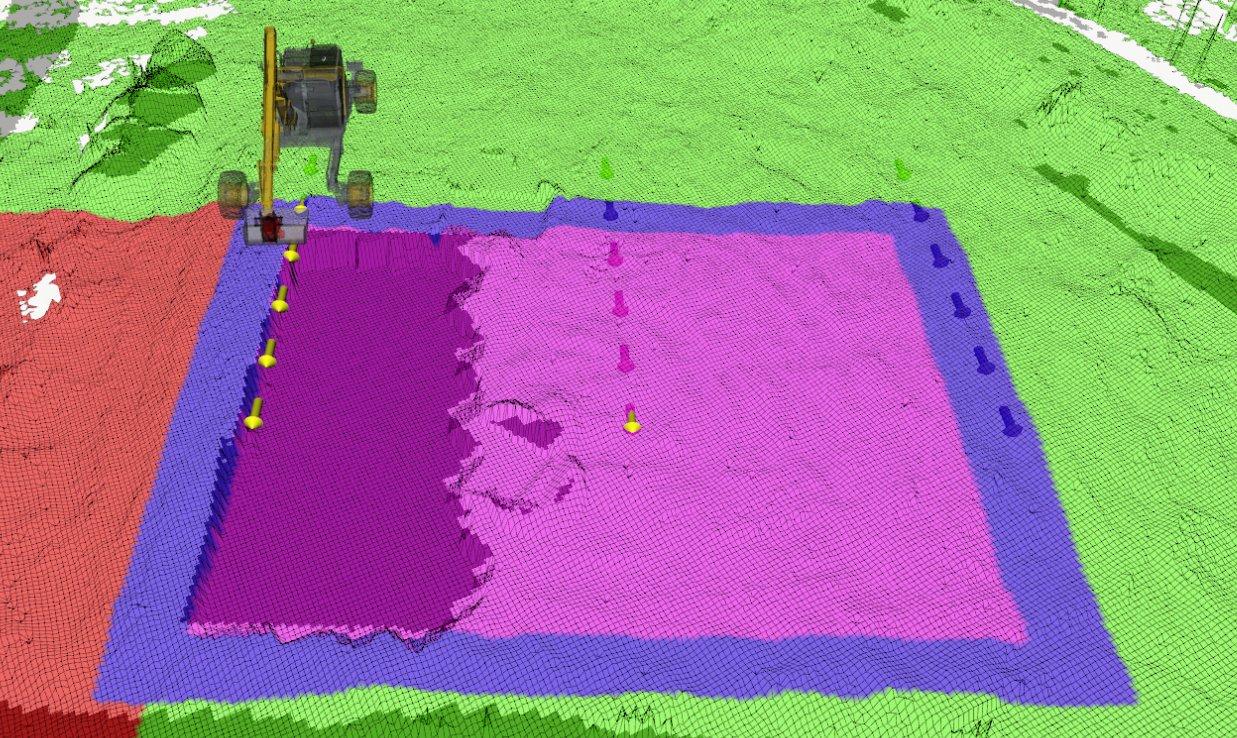}
      \label{fig:v25_1}
    \end{minipage}
  }
  \caption{Excavation scenarios: (a) Soil from the central lane is dumped in the light blue area (temporary dump location). (b) Excavating the middle lane, the soil is starting to accumulate at the pit's edge. (c) Soil has been removed to a permanent dump location. (d) The moment where the local planner failed. There is no space available to dump soil inside the pit, and the dumping area outside the red area is too far away. (e) Simulation of the excavation of a pit with the no-dump area adjacent to the left side of the pit. The soil from the first lane is dumped inside the pit. (f) Soil from the first lane is dumped in a permanent dump area.}
  \label{fig:excavation_scenarios}
\end{figure}

This section provides an overview of different excavation scenarios under varied dumping constraints, shown in the video supplement.
In the first scenario, the back of the first lane is designated as a no-dump area. Two feasible solutions exist, starting at either corner 1 or 2 (see Figure \ref{fig:single_cell}. The solution starting at corner 2 is selected due to its lower cost and efficient excavation pathway (see Figure \ref{fig:v22}).
In the second scenario, permanent dump areas behind all lanes are disallowed, making the excavation paths from both corners longer. Temporary dumping in the back of the first lane precedes the transfer of soil to permanent dump areas (see Figures \ref{fig:24_1} and \ref{fig:24_2}).
In the third scenario, areas behind the lanes are assigned as non-dumping zones. Without any accessible dump areas, the excavator ultimately becomes trapped, illustrating the limitations of the current planning method (see Figures \ref{fig:v26_1} and \ref{fig:v26_2}).
Finally, in the fourth scenario, the side of the first lane is off-limits for dumping. The excavator is forced to temporarily dump the soil inside the pit, making the corner 2 solution infeasible. The corner 1 solution is preferred, and subsequent soil handling follows standard practices (see Figures \ref{fig:v25_1} and \ref{fig:v25_2}).

  

\section{Conclusion and Discussion}
This study presents a fully autonomous system for excavation planning and execution. Using a 12-ton excavator, we demonstrate the ability of the system to dig a pit measuring 15.6 x 11.5 x 1 m in a total of 4 hours and 25 minutes. The system has an average cycle time of about 32 seconds and can move 42.7d m$^3$/h of soil per hour, with a final grade error of 7.2 cm on average.

The global planner tackles the excavation problem by finding a set of base poses that allow the excavator to dig through the excavation area. Using boustrophedon decomposition, it first decomposes the excavation site into cells with simple navigation patterns. It then uses a tree search algorithm to find the optimal order of the cells and minimize the total travel cost, ensuring the feasibility of the excavation. The plan is then completed with dynamic programming to determine the start and end points of the excavator's path for each cell. The local excavation planner determines how to move the soil around the excavator for each base pose. It targets moving the dirt from a reachable digging area into an allowed dumping area. The digging planner uses a Bayesian optimizer to choose the parameters of digging trajectories that maximize the scooped volume in the target workspace. The navigation between base poses is executed by an RRT* sampling-based planner, minimizing travel costs. We utilize Google Earth Pro for target geometries, though professional architecture software is a potential future improvement.


The current global planner and simple zig-zag subroutines restrict the range of possible plans. Future work could integrate a system for navigation and local excavation planning, potentially employing reinforcement learning. Challenges include creating an abstract yet practical environment for sim to real transfer.

One limitation of the current digging planner is that it uses a greedy optimization approach, which can result in trajectories that do not minimize the number of scoops required to complete the local workspace. This limitation can be addressed by using a non-greedy optimization approach. Additionally, simulation speed advances could enable the training of model-free reinforcement learning agents that do excavation planning based on elevation map data  \cite{rudinLearningWalkMinutes2022}.
Instead of going end to end, another possibility is to use a lower-level policy, such as the one described in \cite{egliSoilAdaptiveExcavationUsing2022a}, and train higher-level policy that conditions it to dig on target excavation locations. These approaches could lead to more efficient and effective excavation planning in the future.




Overall, addressing these limitations could lead to a more effective and versatile excavation system in the future.


\bibliographystyle{apalike}

\newpage
\bibliography{bib_static}

\begin{thebibliography}{}

\bibitem[Acar et~al., 2002]{acarMorseDecompositionsCoverage2002b}
Acar, E.~U., Choset, H., Rizzi, A.~A., Atkar, P.~N., and Hull, D. (2002).
\newblock Morse {{Decompositions}} for {{Coverage Tasks}}.
\newblock {\em The International Journal of Robotics Research}, 21(4):331--344.

\bibitem[Bradley and Seward, 1995]{bradleyDevelopingRealtimeAutonomous1995}
Bradley, D. and Seward, D. (1995).
\newblock Developing real-time autonomous excavation-the {{LUCIE}} story.
\newblock In {\em Proceedings of 1995 34th {{IEEE Conference}} on {{Decision}}
  and {{Control}}}, volume~3, pages 3028--3033.

\bibitem[Bureau~of Labor~Statistics, 2022]{LookWorkplaceDeaths}
Bureau~of Labor~Statistics, U. D. o.~L. (2022).
\newblock A look at workplace deaths, injuries, and illnesses on {Workers'
  Memorial Day}: {The Economics Daily}: {U.S. Bureau of Labor Statistics}.
\newblock
  \url{https://www.bls.gov/opub/ted/2022/a-look-at-workplace-deaths-injuries-and-illnesses-on-workers-memorial-day.htm}.

\bibitem[Cannon, 1999]{cannonExtendedEarthmovingAutonomous1999}
Cannon, H. (1999).
\newblock Extended {{Earthmoving}} with an {{Autonomous Excavator}}.
\newblock Master's thesis, Carnegie Mellon University, {Pittsburgh, PA}.

\bibitem[Cao et~al., 2020]{caoHierarchicalCoveragePath2020}
Cao, C., Zhang, J., Travers, M., and Choset, H. (2020).
\newblock Hierarchical {{Coverage Path Planning}} in {{Complex 3D
  Environments}}.
\newblock In {\em 2020 {{IEEE International Conference}} on {{Robotics}} and
  {{Automation}} ({{ICRA}})}, pages 3206--3212, {Paris, France}. {IEEE}.

\bibitem[Choset, 2000]{chosetCoverageKnownSpaces2000}
Choset, H. (2000).
\newblock Coverage of {{Known Spaces}}: {{The Boustrophedon Cellular
  Decomposition}}.
\newblock {\em Autonomous Robots}, 9(3):247--253.

\bibitem[Egli et~al., 2022]{egliSoilAdaptiveExcavationUsing2022a}
Egli, P., Gaschen, D., Kerscher, S., Jud, D., and Hutter, M. (2022).
\newblock Soil-{{Adaptive Excavation Using Reinforcement Learning}}.
\newblock {\em IEEE Robotics and Automation Letters}, 7(4):9778--9785.

\bibitem[Galceran and Carreras, 2013]{galceranSurveyCoveragePath2013}
Galceran, E. and Carreras, M. (2013).
\newblock A survey on coverage path planning for robotics.
\newblock {\em Robotics and Autonomous Systems}, 61(12):1258--1276.

\bibitem[Groll et~al., 2019]{grollAutonomousTrenchingHierarchically2019}
Groll, T., Hemer, S., Ropertz, T., and Berns, K. (2019).
\newblock Autonomous trenching with hierarchically organized primitives.
\newblock {\em Automation in Construction}, 98:214--224.

\bibitem[Hurkxkens, 2020]{hurkxkensRoboticLandscapesTopological2020}
Hurkxkens, I. (2020).
\newblock {\em Robotic {{Landscapes}}: {{Topological Approaches}} to
  {{Terrain}}, {{Design}}, and {{Fabrication}}}.
\newblock Doctoral {{Thesis}}, ETH Zurich.

\bibitem[Hutter et~al., 2017]{hutterForceControlActive2017}
Hutter, M., Leemann, P., Hottiger, G., Figi, R., Tagmann, S., Rey, G., and
  Small, G. (2017).
\newblock Force {{Control}} for {{Active Chassis Balancing}}.
\newblock {\em IEEE/ASME Transactions on Mechatronics}, 22(2):613--622.

\bibitem[Jelavi{\'c} et~al., 2022]{jelavicRoboticPrecisionHarvesting2022}
Jelavi{\'c}, E., Jud, D., Egli, P., and Hutter, M. (2022).
\newblock Robotic {{Precision Harvesting}}: {{Mapping}}, {{Localization}},
  {{Planning}} and {{Control}} for a {{Legged Tree Harvester}}.
\newblock {\em Field Robotics}, 2(1):1386--1431.

\bibitem[Jelavic et~al., 2022]{jelavicOpen3DSLAMPoint2022b}
Jelavic, E., Nubert, J., and Hutter, M. (2022).
\newblock {{Open3D SLAM}}: {{Point Cloud Based Mapping}} and {{Localization}}
  for {{Education}}.
\newblock In {\em Robotic {{Perception}} and {{Mapping}}: {{Emerging
  Techniques}}, {{ICRA}} 2022 {{Workshop}}}, page~24. {ETH Zurich, Robotic
  Systems Lab}.

\bibitem[Johns et~al., 2020]{johnsAutonomousDryStone2020a}
Johns, R.~L., Wermelinger, M., Mascaro, R., Jud, D., Gramazio, F., Kohler, M.,
  Chli, M., and Hutter, M. (2020).
\newblock Autonomous dry stone.
\newblock {\em Construction Robotics}, 4(3):127--140.

\bibitem[Jud et~al., 2017]{judPlanningControlAutonomous2017a}
Jud, D., Hottiger, G., Leemann, P., and Hutter, M. (2017).
\newblock Planning and {{Control}} for {{Autonomous Excavation}}.
\newblock {\em IEEE Robotics and Automation Letters}, 2(4):2151--2158.

\bibitem[Jud et~al., 2021a]{judRoboticEmbankment2021}
Jud, D., Hurkxkens, I., Girot, C., and Hutter, M. (2021a).
\newblock Robotic embankment.
\newblock {\em Construction Robotics}, 5(2):101--113.

\bibitem[Jud et~al., 2021b]{judHEAPAutonomousWalking2021}
Jud, D., Kerscher, S., Wermelinger, M., Jelavic, E., Egli, P., Leemann, P.,
  Hottiger, G., and Hutter, M. (2021b).
\newblock {{HEAP}} - {{The}} autonomous walking excavator.
\newblock {\em Automation in Construction}, 129:103783.

\bibitem[Jud et~al., 2019]{judAutonomousFreeFormTrenching2019}
Jud, D., Leemann, P., Kerscher, S., and Hutter, M. (2019).
\newblock Autonomous {{Free-Form Trenching Using}} a {{Walking Excavator}}.
\newblock {\em IEEE Robotics and Automation Letters}, 4(4):3208--3215.

\bibitem[Khattak et~al., 2020]{khattakComplementaryMultiModal2020}
Khattak, S., Nguyen, H., Mascarich, F., Dang, T., and Alexis, K. (2020).
\newblock Complementary {{Multi}}\textendash{{Modal Sensor Fusion}} for
  {{Resilient Robot Pose Estimation}} in {{Subterranean Environments}}.
\newblock In {\em 2020 {{International Conference}} on {{Unmanned Aircraft
  Systems}} ({{ICUAS}})}, pages 1024--1029.

\bibitem[Kim et~al., 2012]{kimIntelligentNavigationStrategies2012}
Kim, S.-K., Seo, J., and Russell, J.~S. (2012).
\newblock Intelligent navigation strategies for an automated earthwork system.
\newblock {\em Automation in Construction}, 21:132--147.

\bibitem[Kumparak, 2022]{kumparakBuiltRoboticsRaises2022a}
Kumparak, G. (2022).
\newblock Built {{Robotics Raises Another}} \${{64M}} to {{Make Construction
  Equipment Autonomous}}.
\newblock
  https://techcrunch.com/2022/04/01/built-robotics-raises-another-64m-to-make-construction-equipment-autonomous/.

\bibitem[Lee et~al., 2021]{leeRealTimeMotionPlanning2021b}
Lee, D., Jang, I., Byun, J., Seo, H., and Kim, H.~J. (2021).
\newblock Real-{{Time Motion Planning}} of a {{Hydraulic Excavator}} using
  {{Trajectory Optimization}} and {{Model Predictive Control}}.

\bibitem[{Leica Geosystems}, 2023]{LeicaGeosystems}
{Leica Geosystems} (2023).
\newblock {Leica Geosystems - Laser Scanners}.
\newblock \url{https://leica-geosystems.com/products/laser-scanners}.

\bibitem[Litvin and Litvin, 2020]{litvinEvaluationEffectExcavator2020}
Litvin, O. and Litvin, Y. (2020).
\newblock Evaluation of {{Effect}} of the {{Excavator Cycle Duration}} on its
  {{Productivity}}.
\newblock {\em E3S Web of Conferences}, 174:01010.

\bibitem[McKinsey, 2016]{BeatingLowproductivityTrap}
McKinsey (2016).
\newblock Beating the low-productivity trap: {{How}} to transform construction
  operations.
\newblock
  https://www.mckinsey.com/capabilities/operations/our-insights/beating-the-low-productivity-trap-how-to-transform-construction-operations.

\bibitem[McKinsey, 2022]{AddressingUSConstruction}
McKinsey (2022).
\newblock Addressing the {{US}} construction labor shortage.
\newblock
  https://www.mckinsey.com/industries/public-and-social-sector/our-insights/will-a-labor-crunch-derail-plans-to-upgrade-us-infrastructure.

\bibitem[Nubert et~al., 2022]{nubertGraphbasedMultisensorFusion2022}
Nubert, J., Khattak, S., and Hutter, M. (2022).
\newblock Graph-based {{Multi-sensor Fusion}} for {{Consistent Localization}}
  of {{Autonomous Construction Robots}}.
\newblock In {\em 2022 {{International Conference}} on {{Robotics}} and
  {{Automation}} ({{ICRA}})}, pages 10048--10054, {Philadelphia, PA, USA}.
  {IEEE}.

\bibitem[Rudin et~al., 2022]{rudinLearningWalkMinutes2022}
Rudin, N., Hoeller, D., Reist, P., and Hutter, M. (2022).
\newblock Learning to {{Walk}} in {{Minutes Using Massively Parallel Deep
  Reinforcement Learning}}.
\newblock In {\em Proceedings of the 5th {{Conference}} on {{Robot Learning}}},
  pages 91--100. {PMLR}.

\bibitem[Seo et~al., 2011]{seoTaskPlannerDesign2011a}
Seo, J., Lee, S., Kim, J., and Kim, S.-K. (2011).
\newblock Task planner design for an automated excavation system.
\newblock {\em Automation in Construction}, 20(7):954--966.

\bibitem[Silvestri et~al., 2017]{silvestriBranchandcutAlgorithmMinimum2017}
Silvestri, S., Laporte, G., and Cerulli, R. (2017).
\newblock A branch-and-cut algorithm for the minimum branch vertices spanning
  tree problem.
\newblock {\em Computers \& Operations Research}, 81:322--332.

\bibitem[Singh and Cannon, 1998]{singhMultiresolutionPlanningEarthmoving1998a}
Singh, S. and Cannon, H. (1998).
\newblock Multi-resolution planning for earthmoving.
\newblock In {\em Proceedings. 1998 {{IEEE International Conference}} on
  {{Robotics}} and {{Automation}} ({{Cat}}. {{No}}.{{98CH36146}})}, volume~1,
  pages 121--126 vol.1.

\bibitem[Son et~al., 2020]{sonExpertEmulatingExcavationTrajectory2020a}
Son, B., Kim, C., Kim, C., and Lee, D. (2020).
\newblock Expert-{{Emulating Excavation Trajectory Planning}} for {{Autonomous
  Robotic Industrial Excavator}}.
\newblock In {\em 2020 {{IEEE}}/{{RSJ International Conference}} on
  {{Intelligent Robots}} and {{Systems}} ({{IROS}})}, pages 2656--2662, {Las
  Vegas, NV, USA}. {IEEE}.

\bibitem[Stentz et~al., 1998]{stentzRoboticExcavatorAutonomous1998}
Stentz, A., Bares, J., Singh, S., and Rowe, P. (1998).
\newblock A robotic excavator for autonomous truck loading.
\newblock In {\em Proceedings. 1998 {{IEEE}}/{{RSJ International Conference}}
  on {{Intelligent Robots}} and {{Systems}}. {{Innovations}} in {{Theory}},
  {{Practice}} and {{Applications}} ({{Cat}}. {{No}}.{{98CH36190}})}, volume~3,
  pages 1885--1893, {Victoria, BC, Canada}. {IEEE}.

\bibitem[Sucan et~al., 2012]{sucanOpenMotionPlanning2012}
Sucan, I.~A., Moll, M., and Kavraki, L.~E. (2012).
\newblock The {{Open Motion Planning Library}}.
\newblock {\em IEEE Robotics \& Automation Magazine}, 19(4):72--82.

\bibitem[{Wingtra}, 2023]{WingtraMappingDrone}
{Wingtra} (2023).
\newblock {Wingtra - Mapping Drone WingtraOne: All Features}.
\newblock \url{https://wingtra.com/mapping-drone-wingtraone/all-features/}.

\bibitem[Woo et~al., 2018]{wooDeepReinforcementLearning2018a}
Woo, S., Yeon, J., Ji, M., Moon, I.-C., and Park, J. (2018).
\newblock Deep {{Reinforcement Learning}} with {{Fully Convolutional Neural
  Network}} to {{Solve}} an {{Earthwork Scheduling Problem}}.
\newblock In {\em 2018 {{IEEE International Conference}} on {{Systems}},
  {{Man}}, and {{Cybernetics}} ({{SMC}})}, pages 4236--4242.

\bibitem[Xie et~al., 2019]{xieIntegratedTravelingSalesman2019}
Xie, J., Carrillo, L. R.~G., and Jin, L. (2019).
\newblock An {{Integrated Traveling Salesman}} and {{Coverage Path Planning
  Problem}} for {{Unmanned Aircraft Systems}}.
\newblock {\em IEEE Control Systems Letters}, 3(1):67--72.

\bibitem[Zhang et~al., 2021]{zhangAutonomousExcavatorSystem2021a}
Zhang, L., Zhao, J., Long, P., Wang, L., Qian, L., Lu, F., Song, X., and
  Manocha, D. (2021).
\newblock An autonomous excavator system for material loading tasks.
\newblock {\em Science Robotics}, 6(55):eabc3164.

\bibitem[Šopić et~al., 2021]{cycle_time}
Šopić, M., Vukomanović, M., Car-Pušić, D., and Završki, I. (2021).
\newblock Estimation of the excavator actual productivity at the construction
  site using video analysis.
\newblock {\em Organization, Technology and Management in Construction: an
  International Journal}, 13(1):2341--2352.

\end{thebibliography}

\end{document}